%% file: main.tex
\documentclass{article} 
\usepackage{iclr2027_conference,times}
\input{math_commands.tex}

\usepackage[utf8]{inputenc}
\usepackage{url}
\usepackage{graphicx}
\usepackage{booktabs}
\usepackage{multirow}
\usepackage{makecell}
\usepackage{array}
\usepackage{enumitem}
\usepackage{amsmath}
\usepackage{amssymb} 
\usepackage{float} 
\usepackage{wrapfig} 

\usepackage[hidelinks]{hyperref}
\hypersetup{
  colorlinks=true,
  linkcolor=TsinghuaPurple,
  citecolor=TsinghuaPurple,
  urlcolor=TsinghuaPurple
}

\title{
ConfAL-WM: Confidence-Guided Active Learning\\[-0.12em]
for Action-Conditioned World Models
}

\author{%
\begin{minipage}{\textwidth}
\centering
\makebox[\textwidth][c]{%
\authname{Xiang Liu\textsuperscript{1,*}}\hfill
\authname{Kunwei Wu\textsuperscript{2}}\hfill
\authname{Miao Liu\textsuperscript{1}}\hfill
\authname{Sen Cui\textsuperscript{1,\textdaggerdbl}}\hfill
\authname{Changshui Zhang\textsuperscript{1,\textdagger}}}\endgraf
\vspace{1.35em}
{\sffamily\small
\textsuperscript{1}Tsinghua University\qquad
\textsuperscript{2}National University of Singapore\endgraf}
\vspace{1.25em}
{\fontsize{9pt}{10pt}\selectfont\bfseries\color{TsinghuaPurple}
\textsuperscript{*}Core Author\quad
\textsuperscript{\textdaggerdbl}Project Leader \& Corresponding Author\quad
\textsuperscript{\textdagger}Corresponding Author\endgraf}
\vspace{1.15em}
{\fontsize{7.1pt}{8.5pt}\selectfont
\makebox[\textwidth][c]{%
\textsf{\textbf{Contact:}}\enspace
\href{mailto:xiang-liu25@mails.tsinghua.edu.cn}{\texttt{xiang-liu25@mails.tsinghua.edu.cn}}\enspace\textbar\enspace
\href{mailto:cuis@mail.tsinghua.edu.cn}{\texttt{cuis@mail.tsinghua.edu.cn}}\enspace\textbar\enspace
\href{mailto:zcs@mail.tsinghua.edu.cn}{\texttt{zcs@mail.tsinghua.edu.cn}}}\endgraf}
\end{minipage}%
}

\newcommand{\method}{ConfAL-WM}

\newcommand{\cprobe}{C$^3$}

\begin{document}

\maketitle

\begin{abstract}
Action-conditioned world models have become an important foundation for embodied prediction, planning, and synthetic data generation, but their errors under new task and scene distributions are often concentrated in localized spatiotemporal regions such as robot arms, manipulated objects, contact areas, and occluded objects. This paper presents \textbf{\method}, a confidence-guided active learning framework for post-training embodied world models. Building upon EnerVerse-AC (EVAC), we attach a lightweight confidence probe to UNet decoder features and predict dense confidence maps in the latent space. These maps are aggregated into task-, frame-, and patch-level scores, enabling data-budget allocation and localized training enhancement. Our pipeline trains the probe and warms up EVAC on a small target-domain subset. EVAC-v1 then supplies task-level acquisition and optional frame/patch weighting signals; all selected-data models are initialized from the original pretrained EVAC checkpoint for retraining. Experiments on RoboTwin2.0 at the default 40\% data budget show that confidence-guided selection improves post-training quality, while dense frame and patch weighting offers complementary reconstruction and semantic gains compared with scalar reward, progress, and judge-based scoring baselines. A quick visual overview of this work is available at \url{https://ConfAL-WM.github.io}.
\end{abstract}

\begin{figure}[H]
    \centering
    \vspace{-0.45cm}
    \includegraphics[width=1\linewidth]{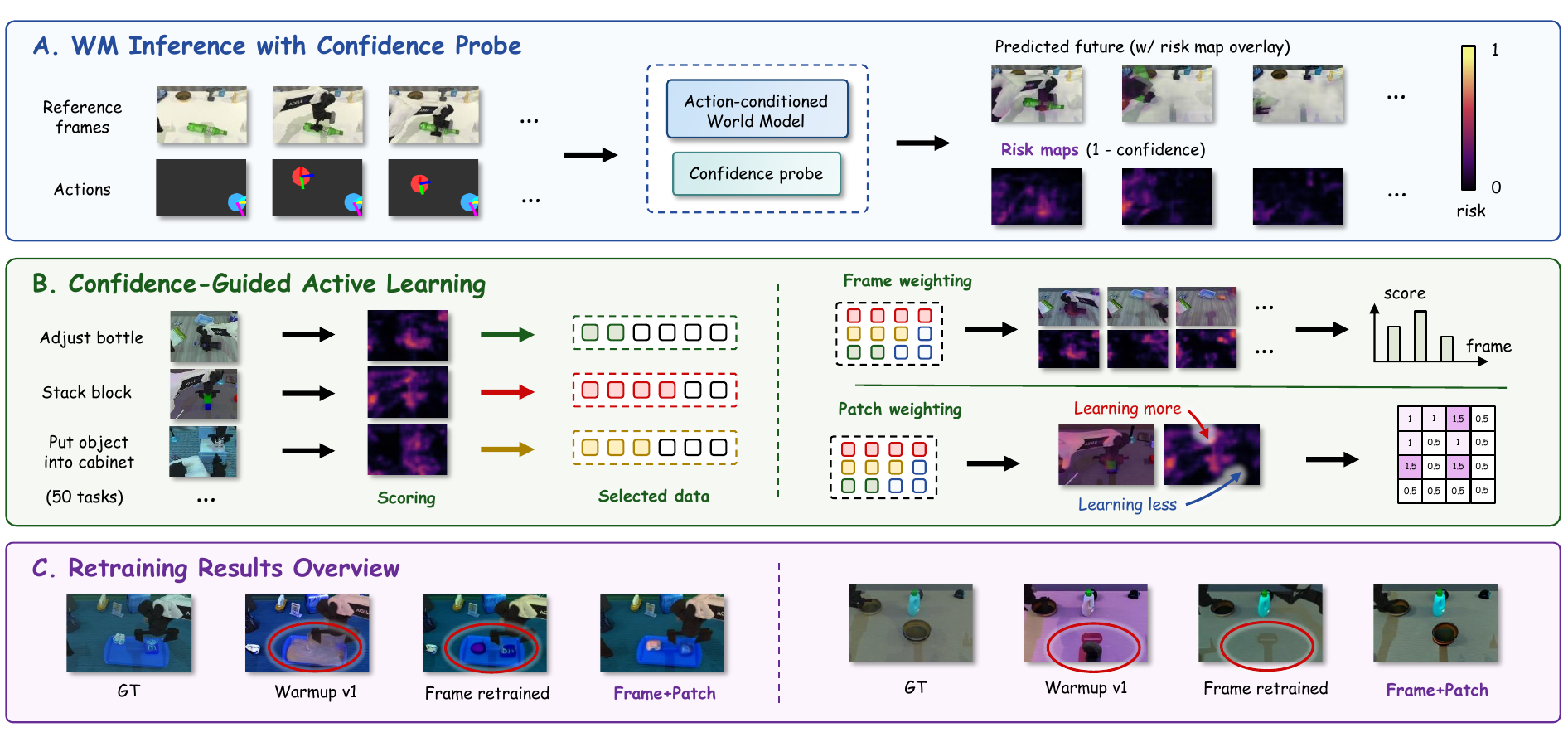}
    \vspace{-0.45cm}
\caption{
\textbf{An introduction of our work.}
(A) Given reference frames and action conditions, we use a UNet-based latent diffusion world model as the backbone and train a confidence probe on decoder features to predict dense future risk maps.
(B) Confidence guides active learning in two ways: selecting more informative post-training data and enhancing supervision on unreliable frames or patches.
(C) Two representative episodes compare the same video frame from the ground truth, warmup model, frame-weighted retraining, and frame-and-patch-weighted retraining.
}
    \vspace{-0.75cm}
    \label{fig:An introduction of our work}
\end{figure}

\newpage
\input{sections/1_introduction}

\input{sections/2_related_work}

\input{sections/3_method}

\input{sections/4_experiments}

\input{sections/5_conclusion}

\newpage
\input{sections/6_Checklist}

\newpage
\bibliography{iclr2027_conference}
\bibliographystyle{iclr2027_conference}

\newpage

\appendix

\input{sections/7_appendix_A}

\input{sections/7_appendix_B}

\newpage
\input{sections/7_appendix_B2}
\newpage
\input{sections/7_appendix_B_Qualitative}

\end{document}

%% file: math_commands.tex
\usepackage{amsmath,amsfonts,bm}

\def\eqref#1{equation~\ref{#1}}

\def\1{\bm{1}}

\DeclareMathAlphabet{\mathsfit}{\encodingdefault}{\sfdefault}{m}{sl}
\SetMathAlphabet{\mathsfit}{bold}{\encodingdefault}{\sfdefault}{bx}{n}



%% file: sections/1_introduction.tex
\section{Introduction}

Action-conditioned embodied world models aim to predict future visual observations under given robot actions, enabling offline policy evaluation, synthetic data generation, and planning without direct environment interaction. EnerVerse-AC (EVAC) is a representative action-conditioned world model that generates future multi-view observations conditioned on robot actions~\citep{jiang2025enerverseac}. However, whether under zero-shot transfer to new datasets or during post-training on existing domains, prediction errors are rarely uniformly distributed over the whole video. They often concentrate around moving robot arms, manipulated objects, contact regions, occlusions, and long-horizon interaction events. Recent studies on confidence-aware video generation and physics-reinforced world simulation also suggest that unreliable or physically inconsistent regions are usually spatially and temporally localized~\citep{mei2025c3,physisforcing2026}. This motivates a post-training pipeline that allocates its available data budget and strengthens local supervision according to the tasks, frames, and regions where the world model is likely to fail.

In this work, we study confidence-guided active learning for action-conditioned world models, and refer to the framework as \textbf{ConfAL-WM}. We build upon EVAC as the backbone and attach a lightweight confidence probe to its UNet decoder features. Following \cprobe{}~\citep{mei2025c3}, the probe learns dense confidence from local latent errors with random threshold conditioning. Decoder features provide spatially resolved representations that combine local detail with contextual information from the latent diffusion backbone. The probe predicts dense confidence maps in the latent space, which are aggregated into patch-, frame-, and task-level scores for selection and retraining. During probe training, an exponential moving average (EMA) adjusts the threshold sampling interval to track the distribution of local prediction errors. Ablations show that decoder features improve confidence prediction, while EMA matches the accuracy of a fixed calibration range.


Our active learning pipeline uses confidence in a staged and budget-aware manner. EVAC is originally pretrained on AgiBot World, a large-scale real-world manipulation dataset~\citep{agibot2025world}, and we use RoboTwin2.0 as the post-training dataset to evaluate the effectiveness of active learning under a new task and scene distribution~\citep{chen2025robotwin}. First, a small subset of data is used to retrain the confidence probe and warm up EVAC, producing a domain-adapted EVAC-v1. Second, we perform fast task-level prescreening with the confidence probe: harder tasks are assigned larger sampling budgets, while easier tasks receive fewer selected scenes. Third, after within-task random sampling, all selected-data models are initialized from the original pretrained EVAC checkpoint. EVAC-v1 supplies the acquisition and weighting signals. EVAC-v2 uses either the selected data directly or frame- and patch-level scores from additional EVAC-v1 inference and probe scoring. This extra scoring stage enables confidence-guided data enhancement during retraining.

This design gives confidence two roles. At the selection level, it allocates sampling budgets across tasks before random within-task selection. At the training level, it localizes which frames and patches should receive stronger supervision. This differs from scalar reward, progress, or judge-based scoring models, which can rank trajectories or frames but usually cannot provide patch-level prediction risk for world-model retraining~\citep{lee2026roboreward,liang2026robometer,ma2024gvl,ji2026prmjudge}. In our comparison, these scoring models are used as alternative selection signals, and some of them are further equipped with frame-weighted retraining. Confidence is used as our main signal, and we additionally study frame-and-patch-weighted retraining as a complementary dense enhancement strategy. Following the EWMBench-style evaluation protocol~\citep{yue2025ewmbench}, our results show that confidence-guided selection and confidence-guided frame/patch weighting can improve prediction quality and embodied trajectory consistency over scalar scoring baselines.

Our contributions are summarized as follows:
\begin{enumerate}[leftmargin=2.6em]
\item We propose ConfAL-WM, a staged confidence-guided active learning pipeline for data-budget allocation when post-training embodied action-conditioned world models.
\item We adapt dense confidence estimation to a UNet-based action-conditioned world model using a decoder-feature confidence probe with EMA-based adaptive threshold supervision.
\item We introduce confidence-guided data enhancement, especially frame-and-patch-weighted retraining, to focus supervision on unreliable spatiotemporal regions.
\item We compare confidence with scalar scoring and global top-error selection, demonstrating gains in data selection, low-budget post-training, and localized weighted retraining.
\end{enumerate}

%% file: sections/2_related_work.tex
\section{Related Work}

\paragraph{Action-conditioned world models and evaluation.} 
Action-conditioned world models predict future observations under robot actions. EnerVerse-AC (EVAC), pretrained on AgiBot World, is the UNet-based backbone of our framework~\citep{jiang2025enerverseac,agibot2025world}; we post-train it on RoboTwin2.0 under new tasks, scenes, and robot embodiments~\citep{chen2025robotwin}. Recent alternatives improve action controllability or latent prediction, including IRASim and DINO-WM~\citep{zhu2025irasim,zhou2025dinowm}. Most closely related, \citet{mei2025c3} introduce \cprobe{} for dense confidence estimation in DiT-based video models.

EWMBench evaluates embodied world models through reconstruction, scene, motion, and semantic metrics, and provides our main evaluation protocol~\citep{yue2025ewmbench}. Other benchmarks connect generated futures with policy performance or assess instruction following and physical plausibility~\citep{jang2025dreamgen,li2025worldmodelbench,nvidia2025pbench}.

\paragraph{Dense confidence, reward, progress, and judge signals.}
Dense uncertainty is the closest signal to our method. S$^3$ analyzes uncertainty in generative video models, while C$^3$ introduces dense calibrated confidence maps for controllable video generation~\citep{mei2025squbed,mei2025c3}. PRM-as-a-Judge provides process-level robotic auditing with macro- and micro-level signals~\citep{ji2026prmjudge}. Our confidence additionally serves as an acquisition score and a local training weight.

We compare against several scalar scoring methods. GVL estimates temporal progress through in-context value learning~\citep{ma2024gvl}; RoboReward learns general-purpose vision-language rewards for robotics~\citep{lee2026roboreward}; Robometer combines frame-level progress with trajectory preferences~\citep{liang2026robometer}; and LRMs generate process and completion rewards online~\citep{wu2026lrm}. These methods score trajectories or frames, whereas we also localize patch errors.

Localized supervision has recently been explored from complementary perspectives. CD-LAM reduces action-irrelevant latent bias through embodiment-focused reconstruction and action-aware objectives~\citep{wei2026cdlam}. PhysisForcing strengthens physical consistency by emphasizing physics-informative interaction regions~\citep{physisforcing2026}. Our method identifies such regions through world-model confidence and uses them for active selection and patch-weighted retraining.

\paragraph{Active learning, uncertainty-aware robot learning, and auxiliary world-model objectives.}
Classical active learning selects samples using uncertainty, disagreement, or coverage~\citep{settles2009active,seung1992query,sener2018coreset}. Robotics-specific methods extend these principles to view selection, rollout control, and demonstration curation~\citep{eren2024muse1,dasgupta2024actnerf,li2025rwm,dass2025datamil}. Most closely related, \citet{romer2026uncertainty} derive velocity-field disagreement (VFD) to quantify epistemic uncertainty in flow-based VLAs, and further introduce SAVE, which uses this uncertainty to prioritize tasks and initial scenes for active multitask fine-tuning. This shares our goal of using model uncertainty to reduce adaptation cost and focus supervision on informative data, but operates on action-policy uncertainty, whereas our method estimates dense video-prediction confidence and uses it for both data selection and localized world-model retraining. Future-aware auxiliary objectives can also improve embodied representations~\citep{yuan2026fastwam,lv2026viva,luo2026beingh07}.

%% file: sections/3_method.tex
\section{Method}
\label{sec:method}

In this section, we first introduce a dense confidence probe for a UNet-based action-conditioned world model, which estimates patch-level prediction reliability from intermediate decoder features. We then describe how the resulting dense risk maps are aggregated into task-level acquisition scores and further converted into weights for confidence-guided EVAC retraining.

\subsection{Dense Confidence Probe for UNet World Models}
\label{subsec:confidence_probe}

\begin{figure}[t]
    \centering
    \vspace{-0.3cm}
    \includegraphics[width=1\linewidth]{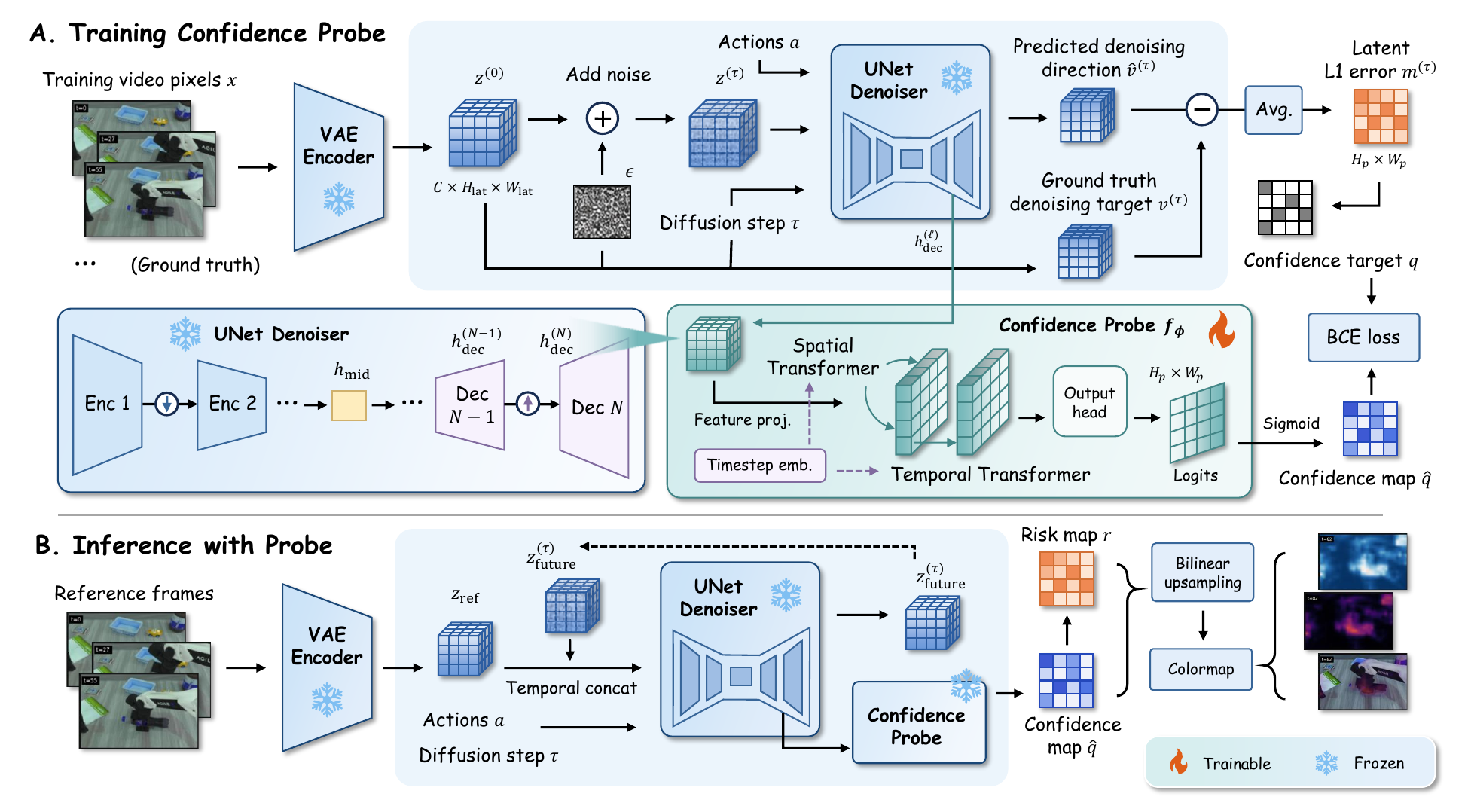}
    \vspace{-0.8cm}
    \caption{\textbf{Training and inference of the confidence probe in the UNet latent diffusion world model.} Video pixels are encoded into latent variables, diffused into $z^{(\tau)}$, processed by the UNet denoiser, and supervised by the denoising target $v^{(\tau)}$. The confidence probe is attached to selectable decoder features $h_{\mathrm{dec}}^{(\ell)}$ and predicts dense confidence maps.}
    \vspace{-0.2cm}
    \label{fig:confidence_probe_arch}
\end{figure}

\paragraph{Confidence definition.}
Given an action-conditioned world model, we define confidence as the predicted probability that a local prediction error is below a threshold. In EVAC, the diffusion model predicts a denoising direction $\hat{v}^{(\tau)}$ for the ground-truth target $v^{(\tau)}$. For a predicted future frame $t$ and latent patch $(i,j)$, we compute a \textbf{local mean absolute error} as
\begin{equation}
    m_{t,(i,j)}^{(\tau)}
    =
    \frac{\sqrt{1-\bar{\alpha}_\tau}}{|\mathcal{P}_{i,j}|}
    \sum_{p\in\mathcal{P}_{i,j}}
    \left|\hat{v}^{(\tau)}_{t,p}-v^{(\tau)}_{t,p}\right|
    =
    \frac{1}{|\mathcal{P}_{i,j}|}
    \sum_{p\in\mathcal{P}_{i,j}}
    \left|\hat{z}_{t,p}^{(\tau)}-z^{(0)}_{t,p}\right|.
\end{equation}
Here, $p=(c,x,y)$ indexes one channel-spatial entry in the latent tensor: $c$ is the latent feature channel and $(x,y)$ is a spatial location in the latent grid. The set $\mathcal{P}_{i,j}$ denotes the latent region corresponding to probe patch $(i,j)$. $z^{(0)}_{t,p}$ is the clean video latent encoded from the ground-truth video, and $\hat{z}^{(\tau)}_{t,p}$ is its closed-form prediction recovered from $\hat{v}^{(\tau)}_{t,p}$ at diffusion timestep $\tau$. This equivalent form avoids executing a complete reverse diffusion trajectory when constructing the confidence target. $\bar{\alpha}_\tau$ is the cumulative noise-schedule coefficient at timestep $\tau$.

The \textbf{binary confidence target} at frame $t$ is defined as
\begin{equation}
    q_t=\{q_{t,(i,j)}\}_{i,j}
    \in\{0,1\}^{ H_p\times W_p},\qquad q_{t,(i,j)}
    =
    \mathbb{I}\{m_{t,(i,j)}^{(\tau)}<\theta_t\},
\end{equation}
where $\theta_t$ is a stochastic threshold sampled from an adaptive interval. Further architectural details of the latent-space error construction are provided in Appendix~\ref{app:patch_error}.

\paragraph{Decoder-feature confidence probe.}

\begin{wrapfigure}{r}{0.35\textwidth}
	\vspace{-.5cm}
    \centering
    \includegraphics[width=0.35\textwidth]{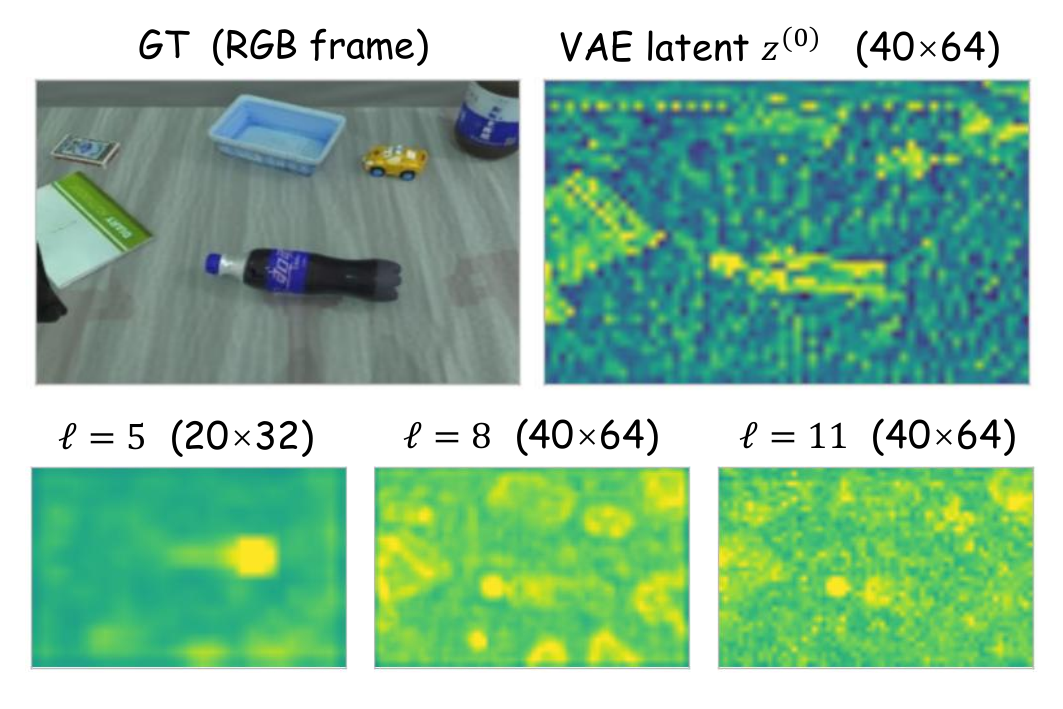}
	\vspace{-0.7cm}
	\caption{\textbf{Direct comparison for a single frame.} All operations are performed in the VAE latent space. The lower panels visualize the channel-averaged $\mathbb{E}_c\left[\left|h_{\mathrm{dec}}^{(\ell)}\right|\right]$.}
    \label{fig:decoder index mean}
	\vspace{-0.6cm}
\end{wrapfigure}

Let $h_{\mathrm{dec}}^{(\ell)}$ denote the feature map extracted from the $\ell$-th UNet decoder block. For each predicted future frame, the \textbf{confidence probe prediction} is
\begin{equation}
    \hat{q}_t=\sigma\!\left(
    f_{\phi}\!\left(
    h_{\mathrm{dec}}^{(\ell)},
    e_{\tau},
    e_{\theta}
    \right)\right)\in[0,1]^{H_p\times W_p},
\end{equation}
where $e_{\tau}$ denotes the diffusion-timestep embedding, $e_{\theta}$ denotes the sampled error threshold embedding, $f_{\phi}$ is the trainable confidence probe parameterized by $\phi$, and $\sigma(\cdot)$ is the element-wise sigmoid function. Internally, the probe applies channel projection, spatial Transformer layers, temporal Transformer layers, and a patch-wise output head. The decoder index $\ell$ trades coarse semantic context for finer spatial resolution, as shown in Figure~\ref{fig:decoder index mean}; we use $\ell=5$, yielding a $20\times32$ spatial grid for dense confidence prediction.

\paragraph{Adaptive threshold sampling.}
At training step $s$, we compute the lower and upper percentiles of the batch-wise local error distribution, and we update the threshold bounds as
\begin{equation}
    p_l^{(s)}=\operatorname{Quantile}_{0.1}\!\left(m^{(s)}\right),\qquad
    p_h^{(s)}=\operatorname{Quantile}_{0.9}\!\left(m^{(s)}\right),
\end{equation}
\begin{equation}
    l^{(s)}=(1-\gamma_s)l^{(s-1)}+\gamma_s p_l^{(s)},\qquad
    h^{(s)}=(1-\gamma_s)h^{(s-1)}+\gamma_s p_h^{(s)}.
\end{equation}
The update rate $\gamma_s$ is larger during warmup to follow the initial change in error scale, then decreases to smooth the batch-wise quantile estimates.

A separate threshold is sampled for each predicted future frame $t$ as $\theta_t\sim\mathcal{U}\!\left(l^{(s)},h^{(s)}\right)$. The sampled threshold $\theta_t$ is shared by all spatial patches $(i,j)$ within the same frame. Further implementation details and the conditional interpretation of $\theta_{t}$ are provided in Appendix~\ref{app:ema_threshold}.

\paragraph{Probe training objective.}

The probe is trained with binary cross-entropy over all supervised patches in the predicted future frames:
\begin{equation}
    \mathcal{L}_{\mathrm{conf}}
    =
    -\mathbb{E}_{t,(i,j)}
    \Big[
    q_{t,(i,j)}\log\hat{q}_{t,(i,j)}
    +
    (1-q_{t,(i,j)})
    \log(1-\hat{q}_{t,(i,j)})
    \Big].
\end{equation}
During probe training, the EVAC backbone remains frozen and gradients are propagated only through $f_{\phi}$. At inference time, the frame-wise dense risk map is defined as $r_{t,(i,j)}=1-\hat{q}_{t,(i,j)}$.

\subsection{Confidence-Guided Active Learning}
\label{subsec:confidence_al}

\paragraph{Full pipeline.} Given a candidate data pool, our active learning pipeline uses confidence for both sample acquisition and localized retraining. First, we train the confidence probe and warm up EVAC on a small subset of the data pool. The resulting EVAC-v1 is used for task-level prescreening, where representative episodes estimate task difficulty and determine the per-task candidate quota, according to which episodes are randomly sampled within each task. 

All selected-data models are initialized from the original pretrained EVAC checkpoint; EVAC-v1 supplies the acquisition and weighting signals. EVAC-v2 is trained on the selected subset, optionally using confidence maps from additional EVAC-v1 inference to weight its retraining objective.

\paragraph{Risk-based selection criteria.}
We consider three confidence-based selection criteria: \textbf{1. mean risk}, \textbf{2. tail risk}, and \textbf{3. persistent risk}.
Mean risk measures the overall prediction difficulty by averaging the dense risk map over all spatial locations and predicted future frames, whereas tail risk emphasizes severe localized failures. For tail risk, we flatten all patch risks $\{r_{t,(i,j)}\}$ into a vector of length $K=TH_pW_p$ and sort its entries in ascending order as $r_{[1]}\leq r_{[2]}\leq\cdots\leq r_{[K]}$. Let $\eta$ denote the selected high-risk tail ratio. The two scores are defined as
\begin{equation}
    S_{\mathrm{mean}}
    =\mathbb{E}_{t,(i,j)}\left[r_{t,(i,j)}\right],
    \quad
    S_{\mathrm{tail}}
    =
    \frac{1}{\left\lceil\eta K\right\rceil}
    \sum_{n=K-\left\lceil\eta K\right\rceil+1}^{K}
    r_{[n]}.
\end{equation}
\vspace{-0.3cm}

Persistent risk captures high-risk local failures that recur across multiple frames. Let $K_p=H_pW_p$. For frame $t$, we sort its patch risks as
$r_{t,[1]}\leq\cdots\leq r_{t,[K_p]}$ and average the largest fraction of them to obtain a frame-level score $u_t$. We then sort these frame scores as
$u_{[1]}\leq\cdots\leq u_{[T]}$ and average the highest-scoring frames:
\vspace{-0.1cm}
\begin{equation}
    u_t
    =
    \frac{1}{\left\lceil\eta_pK_p\right\rceil}
    \sum_{n=K_p-\left\lceil\eta_pK_p\right\rceil+1}^{K_p}
    r_{t,[n]},
    \quad
    S_{\mathrm{persistent}}
    =
    \frac{1}{\left\lceil\eta_tT\right\rceil}
    \sum_{n=T-\left\lceil\eta_tT\right\rceil+1}^{T}
    u_{[n]}.
\end{equation}
Here, $\eta_p$ denotes the fraction of high-risk patches retained within each frame, and $\eta_t$ denotes the fraction of high-risk frames retained across the predicted video.

\paragraph{Confidence-guided frame and patch weighting.}
For weighted EVAC-v2 retraining, we perform an additional inference and confidence-scoring stage on the selected episodes. Frame weighting assigns the spatially averaged risk uniformly to all patches within frame $t$, whereas patch weighting retains the original dense risk $r_{t,(i,j)}$. We unify the two forms as
\begin{equation}
    r_{t,(i,j)}^{(\alpha)}
    =\mathbb{E}_{(i',j')}
    \left[
    r_{t,(i',j')}
    \right]
    +
    \alpha
    \left(
    r_{t,(i,j)}
    -
    \mathbb{E}_{(i',j')}
    \left[
    r_{t,(i',j')}
    \right]
    \right),
    \qquad
    \alpha\in[0,1].
\end{equation}
When $\alpha=0$, all patches within a frame receive the same frame-level risk; when $\alpha=1$, the original patch-level risk map is recovered. Intermediate values preserve the frame-level baseline while introducing local residual modulation.

After quantile normalization and clipping, we denote the resulting risk by
$\widetilde{r}_{t,(i,j)}^{(\alpha)}\in[0,1]$. It is converted into a local loss multiplier and applied to the world-model objective as
\begin{equation}
    w_{t,(i,j)}
    =
    1
    +
    \lambda_{\mathrm{eff}}(s)
    \widetilde{r}_{t,(i,j)}^{(\alpha)},
    \qquad
    \mathcal{L}_{\mathrm{WM}}
    =
    \frac{
    \mathbb{E}_{t,(i,j)}
    \left[
    w_{t,(i,j)}\ell_{t,(i,j)}
    \right]
    }{
    \mathbb{E}_{t,(i,j)}
    \left[
    w_{t,(i,j)}
    \right]
    }.
\end{equation}
Here, $\ell_{t,(i,j)}$ is the local world-model training loss, and
$\lambda_{\mathrm{eff}}(s)
=
\lambda_{\mathrm{conf}}
\min(1,s/s_{\mathrm{warm}})$
linearly increases the weighting strength during the first
$s_{\mathrm{warm}}$ EVAC-v2 retraining steps. Thus, higher-risk frames and patches receive larger optimization weights. The confidence maps are detached during EVAC retraining, so gradients are propagated only through the world model.

%% file: sections/4_experiments.tex
\section{Experiments}
\label{sec:experiments}

Our experiments address two questions. \textbf{1. Why Confidence?} Does predicted risk track world-model errors and support selection relative to direct error? \textbf{2. Why Active Learning?} How does confidence-guided allocation affect post-training quality across fixed data and optimization budgets?

\subsection{Experimental Setup}
\label{subsec:exp_setup}

\paragraph{Data setting.}
In our setting, EVAC is pretrained on AgiBot World~\citep{agibot2025world}, while active learning and post-training are conducted on a subset of RoboTwin2.0~\citep{chen2025robotwin}. This creates a transfer setting involving new tasks, scenes, and robot embodiments for evaluating adaptation to the target manipulation domain. We use data from the Aloha-AgileX dual-arm robot, covering 50 manipulation tasks. Each task contains approximately 500 randomized scenes, resulting in 24,992 videos in total. The video length ranges from 98 to 578 frames.

\paragraph{Implementation details.}
During confidence-probe training, the entire EVAC backbone is frozen and only the confidence probe $f_{\phi}$ is optimized. During EVAC-v1 warmup and EVAC-v2 retraining, the UNet denoiser and two projection modules are trainable, while the VAE, CLIP embedder, and action-conditioning resampler remain frozen. The full EVAC model has about 2.33B parameters, and the confidence probe has about 8.2M parameters.

We split the data into 6,248 episodes for probe training and EVAC-v1 warmup, 18,244 candidates, and 250 each for validation and testing. The nominal selection budget is 7,298 episodes. EVAC-v2 uses AdamW, learning rate $5\times10^{-5}$, fp16, and 4,000 updates on two A800 GPUs. Scoring averages three probe timesteps $\tau\in[50,200]$. Main tables retain the three reported runs and their averages. Detailed per-seed results and bootstrap statistics are provided in Appendix~\ref{app:al_full_results}. Appendix~\ref{app:oracle_error} reports the direct-error comparison, and Appendix~\ref{app:budget_sensitivity} evaluates shorter, fixed-update budget sweeps. Appendix~\ref{app:selection_protocol} specifies quotas, weighting, and the seeds.


\paragraph{Baselines.} Probe ablations evaluate bottleneck and decoder features on EVAC using \cprobe{}-style fixed-range supervision~\citep{mei2025c3} and EMA supervision (Appendix~\ref{app:c3_controlled}). For data selection, we compare the following methods (superscripts match Figure~\ref{fig:main_comparison}):
RoboReward$^{1}$, a vision-language reward model for robotics~\citep{lee2026roboreward}; GVL$^{2}$, which uses VLMs as in-context value learners~\citep{ma2024gvl}; Robometer-Prog and Robometer-Pref$^{3}$, trajectory progress and preference models~\citep{liang2026robometer}; PRM-as-Judge$^{4}$, a dense process-level robotic auditing method~\citep{ji2026prmjudge}; and LRMs$^{5}$, large reward models for online robot reward generation~\citep{wu2026lrm}.

\paragraph{World-model evaluation.}
We follow the EWMBench-style evaluation pipeline~\citep{yue2025ewmbench}. PSNR and SSIM measure low-level reconstruction quality. Scene consistency measures layout and object preservation. Logics evaluates higher-level physical and interaction plausibility. Sem.-CLIP and Sem.-BLEU measure visual-semantic and textual-semantic agreement. Traj-HSD, Traj-Dyn, and Traj-nDTW evaluate robot trajectories; higher values indicate better trajectory consistency.


Beyond quantitative evaluation, Appendix~\ref{app:training_evolution} and Appendix~\ref{app: more results of qualitative confidence visualization} provide qualitative examples of retraining evolution and confidence-map visualization, respectively.

\newpage
\subsection{Why Confidence?}
\label{subsec:why_confidence}

We evaluate EVAC-v1 confidence as an estimate of world-model prediction error under two protocols. Probe ablations use 250 held-out episodes spanning 43 tasks. Multi-scale risk diagnostics use one episode from each of the 50 prescreened tasks. 

\begin{figure*}[t]
    \centering
    \vspace{-0.2cm}
    \includegraphics[width=0.95\textwidth]{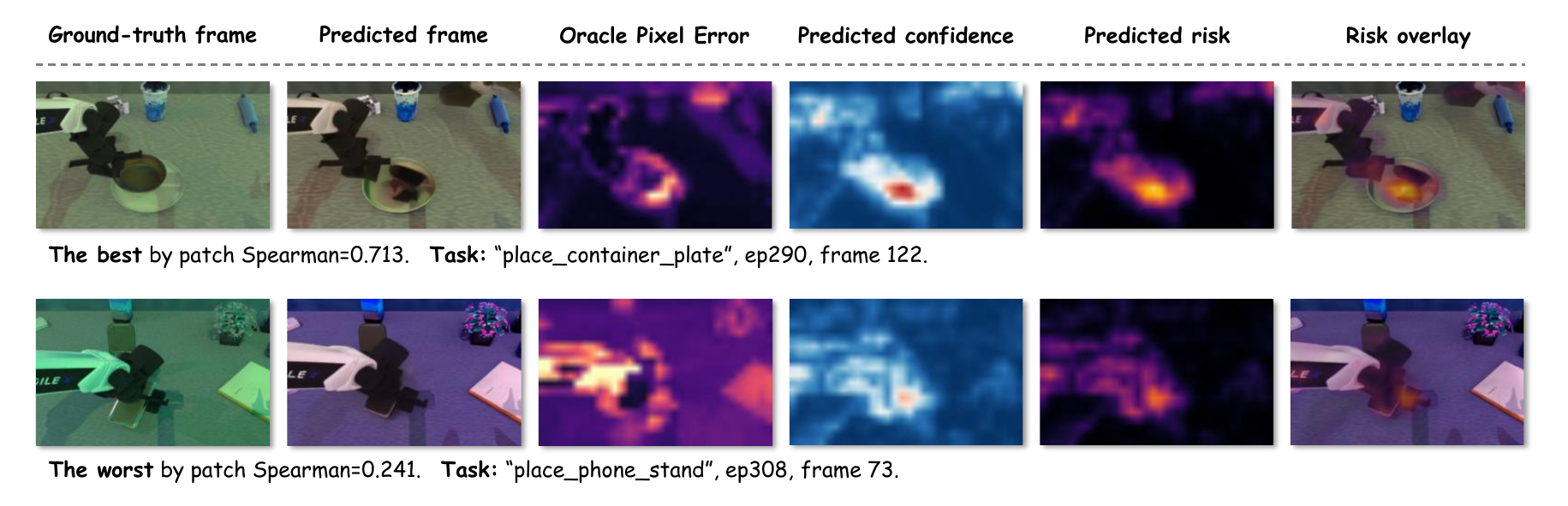}
    \vspace{-0.45cm}
    \caption{
    \textbf{Qualitative confidence visualization.}
    High- and low-agreement episodes from different manipulation tasks are shown. High-risk regions generally coincide with errors around robot arms, manipulated objects, contacts, and occlusions. More examples: see Appendix~\ref{app: more results of qualitative confidence visualization}.
    }
    \label{fig:confidence_qualitative}
    \vspace{-0.2cm}
\end{figure*}

\begin{figure}[t]
    \centering
    \includegraphics[width=0.95\linewidth]{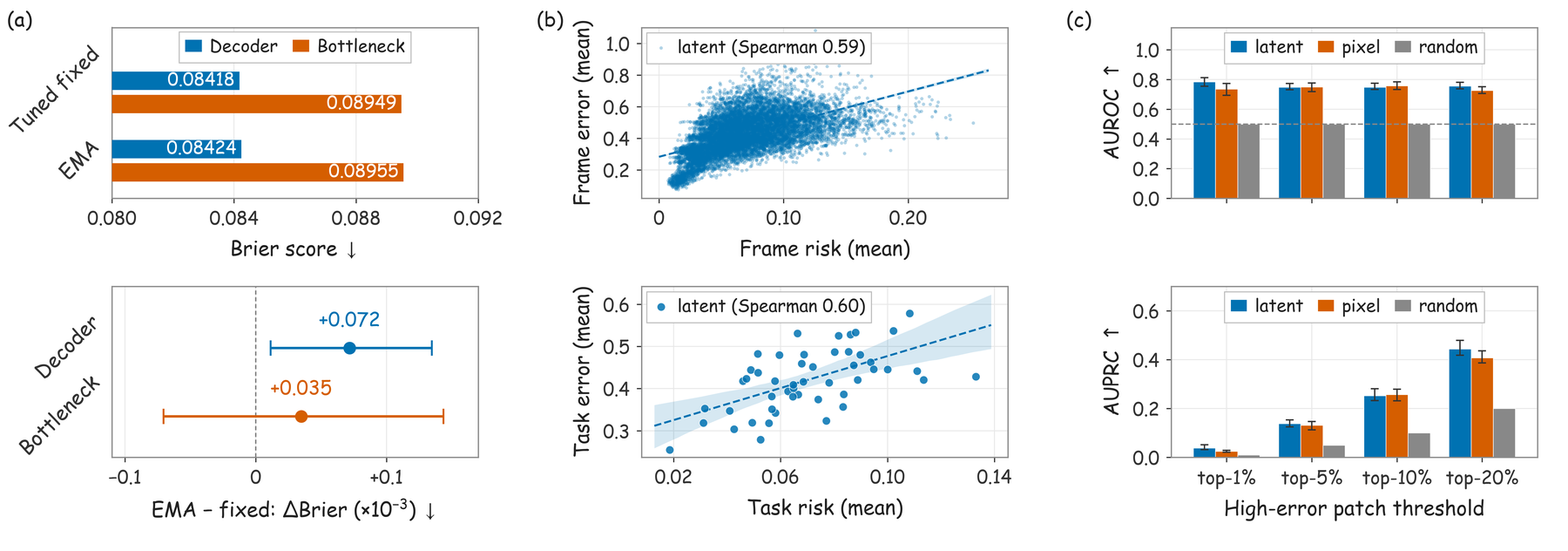}
    \vspace{-0.5cm}
\caption{\textbf{Confidence-probe ablations and multi-scale risk validity.} (a) Feature and threshold-supervision ablations on EVAC-v1, showing episode-mean Brier (top) and paired task-macro EMA-minus-tuned-fixed Brier (bottom). Tuned fixed uses the calibrated $[Q_{.05},Q_{.95}]$ range; error bars denote 95\% task-bootstrap intervals, with positive differences favoring fixed supervision. (b) Mean risk versus latent RMS rollout error at frame (top) and task (bottom) levels. (c) AUROC (top) and AUPRC (bottom) for detecting the top-$Q\%$ highest-error patches in latent and pixel spaces.}

    \label{fig:confidence_signal_validity}
    \vspace{-0.3cm}
\end{figure}

\paragraph{Qualitative localization.}
Figure~\ref{fig:confidence_qualitative} shows that predicted risk maps identify spatially localized failures across the illustrated manipulation episodes and their future video frames. High-risk regions commonly appear around moving manipulators, object interactions, contacts, and temporarily occluded objects. Their agreement with pixel prediction-error maps indicates that the probe captures local failures during the execution of diverse manipulation tasks.

\paragraph{Confidence as a multi-scale ranking signal.}
We quantitatively compare risk with oracle error across patch, frame, and task levels. Figure~\ref{fig:confidence_signal_validity}(a) summarizes the controlled probe ablations: decoder features reduce Brier by 5.93\% under both supervision rules, while EMA and tuned fixed supervision yield closely matched scores. Figure~\ref{fig:confidence_signal_validity}(b--c) evaluates the resulting risk signal across scales. In latent space, risk achieves Spearman correlations of $0.540$, $0.590$, and $0.595$ at the patch, frame, and task levels, respectively. For detecting the top-5\% highest-error patches, it obtains an AUROC of $0.761$ and an AUPRC of $0.146$, compared with random baselines of $0.5$ and $0.05$. These results support aggregating dense risk into frame- and task-level scores for active data selection.

\paragraph{Spatial and temporal behavior.}
Table~\ref{tab:confidence_validity} further shows that risk maps are temporally stable: adjacent frames obtain a top-region IoU of $0.740$, while the flicker score is only $0.005$. Risk and latent error are also temporally synchronized, with a peak correlation of $0.602$ occurring near zero lag. Nevertheless, the top-$5\%$ spatial IoU is $0.130$, suggesting that confidence partially identifies error-prone regions but does not precisely reproduce their boundaries.

\begin{wraptable}{r}{0.55\textwidth}
    \centering
    \vspace{-0.2cm}
    \caption{\textbf{Summary of latent-space confidence validity across multiple scales.}
Frame- and task-level statistics use mean aggregation.}
    \vspace{0.2cm}
    \label{tab:confidence_validity}
    \scriptsize
    \setlength{\tabcolsep}{3pt}
    \renewcommand{\arraystretch}{1.05}
    \resizebox{0.55\textwidth}{!}{
    \begin{tabular}{l|l|c}
        \toprule
        \rowcolor{gray!12}
        Property & Metric & Result \\
        \midrule
        Multi-scale ranking
        & Patch / Frame / Episode Spearman ($\uparrow$)
        & $0.540/0.590/0.595$ \\
        High-error detection
        & AUROC / AUPRC@top-$5\%$ ($\uparrow$)
        & $0.761/0.146$ \\
        Spatial agreement
        & Top-$5\%$ IoU ($\uparrow$)
        & $0.130$ \\
        Temporal stability
        & Adjacent-frame IoU ($\uparrow$) / Flicker ($\downarrow$)
        & $0.740/0.005$ \\
        Temporal alignment
        & Peak correlation ($\uparrow$) / $|\mathrm{lag}|$ ($\downarrow$)
        & $0.602/\approx 0$ \\
        \bottomrule
    \end{tabular}
    }
    \vspace{-1em}
\end{wraptable}


Overall, the probe provides an \emph{ordinal} risk signal for ranking patches, frames, and tasks. On frozen EVAC-v1, decoder features reduce Brier from $0.08949$ to $0.08418$ under fixed-range supervision, while EMA achieves a similar Brier of $0.08424$. Appendix~\ref{app:c3_controlled} presents feature and threshold ablations; further confidence diagnostics appear in Appendix~\ref{app:Additional Confidence Evaluation Details}.

\subsection{Why Active Learning?}
\label{subsec:why_active_learning}

We evaluate whether confidence-guided selection and retraining improve the action-conditioned world model under a fixed data budget. We report selection-only retraining and selection with additional weighting, since the latter requires an extra inference and scoring stage after selection.

For compact comparison, all component metrics are normalized to $[0,1]$ and aggregated into four dimensions: Reconstruction $=(\mathrm{PSNR}+\mathrm{SSIM})/2$, Scene $=\mathrm{Scene\ Consistency}$, Semantics $=(\mathrm{Logics}+\mathrm{Sem.\mbox{-}CLIP}+\mathrm{Sem.\mbox{-}BLEU})/3$, and Motion $=\mathrm{Traj\mbox{-}HSD}+\mathrm{Traj\mbox{-}Dyn}+\mathrm{Traj\mbox{-}nDTW}$.

\begin{figure}[h]
    \centering
    \vspace{-0.2cm}
    \includegraphics[width=\linewidth]{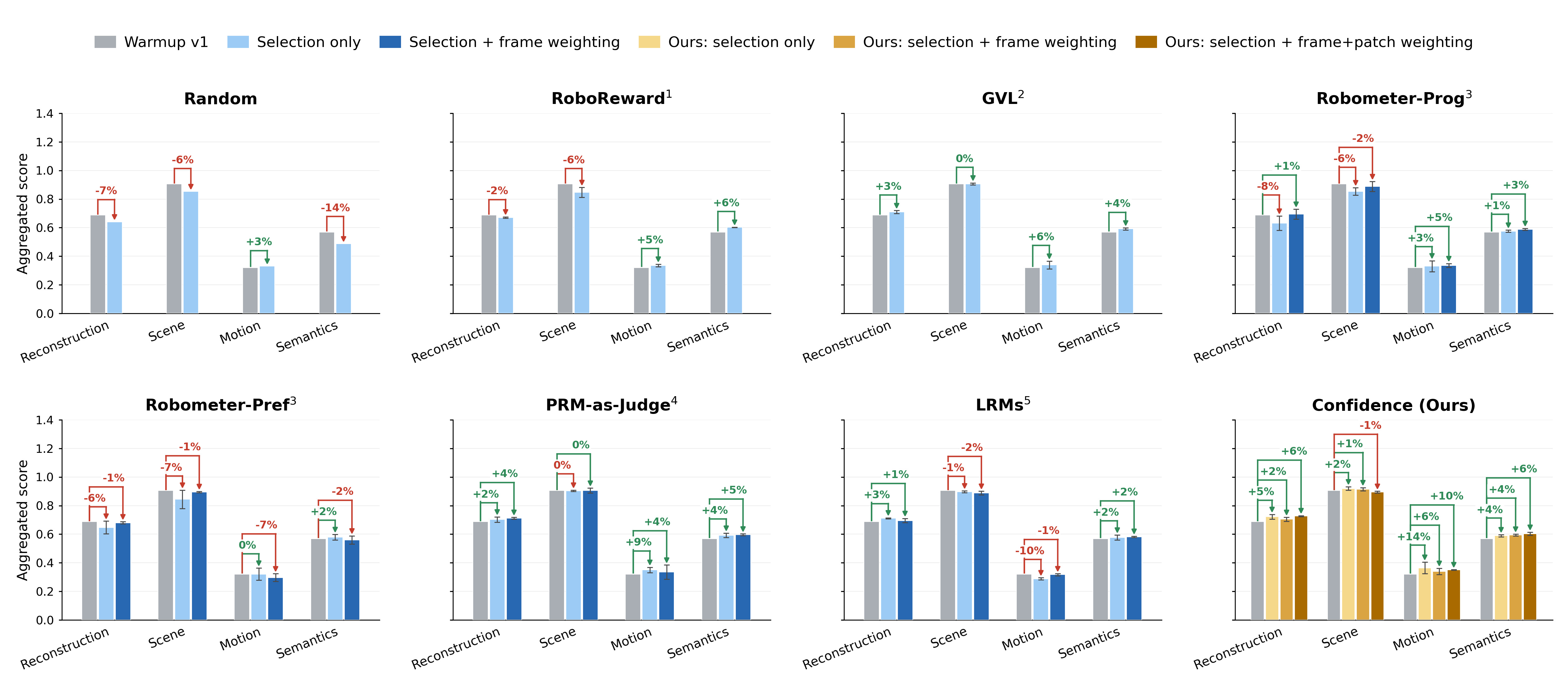}
    \vspace{-0.8cm}
    \caption{
    \textbf{Main comparison of active-learning scoring and retraining strategies.}
    Each panel compares EVAC-v1 with selection-only retraining and additional frame weighting. Confidence enables frame+patch weighting. Learned-method bars report means over three seeds, and arrows report relative changes from EVAC-v1. Error bars indicate paired-bootstrap 95\% confidence intervals.
    }
    \vspace{-0.2 cm}
    \label{fig:main_comparison}
\end{figure}

\paragraph{Main comparison.}
Figure~\ref{fig:main_comparison} and Table~\ref{tab:main_results} show that confidence-guided mean-risk selection leads seven of nine metrics among selection-only methods at the 40\% budget. Confidence-guided task allocation exceeds global top-error selection on eight of nine metrics in the individual-run comparison, improving reconstruction, scene consistency, and motion (Appendix~\ref{app:oracle_error}).


Additional weighting provides complementary gains. Among weighted methods, confidence with frame weighting leads Scene Consistency; frame-and-patch weighting leads PSNR, SSIM, Logics (tied), Sem.-CLIP, Sem.-BLEU, and all three trajectory metrics. These results motivate two uses of confidence: \textbf{1. Budget-aware selection:} mean risk allocates available training data across tasks; \textbf{2. Localized enhancement:} dense maps emphasize unreliable frames and patches during retraining.

These gains involve trade-offs: compared with selection-only confidence retraining, frame-and-patch weighting improves Reconstruction and Semantics but reduces Scene Consistency and Motion. It improves Motion relative to frame-only weighting. This pattern is compatible with metric trade-offs discussed in prior work~\citep{liao2024devil,dou2026sgc,ye2026shift}, but does not establish their cause. It also highlights the limitations of these proxy metrics.

We also ablate active learning across different candidate-pool budget fractions. With 750 updates and a separate 64-episode evaluation subset, confidence improves all four aggregates over random at 10\% budget; larger budgets show metric-dependent trade-offs (Appendix~\ref{app:budget_sensitivity}).

\begin{table*}[t]
\vspace{-0.4cm}
\caption{
\textbf{Detailed normalized results under the same active-learning setting.}
Learned-method results are averaged over three seeds
(42, 3407, and 123), and oracle uses optimizer seed 42. 
Methods are separated into selection-only retraining and selection with
additional weighting.
Green, yellow, and red indicate the best, second-best, and third-best results
within each block, respectively.
}
\label{tab:main_results}
\centering
\scriptsize
\setlength{\tabcolsep}{2.8pt}
\renewcommand{\arraystretch}{1.04}
\resizebox{\textwidth}{!}{
\begin{tabular}{l|l||cc|c|ccc|ccc}
\toprule
\rowcolor{gray!12}
\multirow{2}{*}{Scoring} & \multirow{2}{*}{Weighting} & \multicolumn{2}{c}{Reconstruction ($\uparrow$)} & \multicolumn{1}{c}{Scene ($\uparrow$)} & \multicolumn{3}{c}{Semantics ($\uparrow$)} & \multicolumn{3}{c}{Motion ($\uparrow$)} \\
\cmidrule{3-11}
\rowcolor{gray!12}
& & PSNR ($\uparrow$) & SSIM ($\uparrow$) & Scene Cons. ($\uparrow$) & Logics ($\uparrow$) & Sem.-CLIP ($\uparrow$) & Sem.-BLEU ($\uparrow$) & Traj-HSD ($\uparrow$) & Traj-Dyn ($\uparrow$) & Traj-nDTW ($\uparrow$) \\
\midrule

\multicolumn{2}{l||}{Base EVAC} & 0.5532 & 0.5778 & 0.8757 & 0.4298 & 0.8523 & 0.1799 & 0.0007 & 0.0001 & 0.0006 \\

\multicolumn{2}{l||}{Base EVAC (Warmup v1)} & 0.6446 & 0.7309 & 0.9047 & 0.5537 & 0.8824 & 0.2708 & 0.1045 & 0.0721 & 0.1439 \\

\midrule
\rowcolor{gray!6}
\multicolumn{11}{c}{\textbf{Selection-only retraining: no additional weighting after selection}} \\
\midrule

Random & None & 0.5968 & 0.6849 & 0.8524 & 0.3554 & 0.8689 & 0.2372 & 0.1067 & \cellcolor{red!12}0.0725 & 0.1518 \\

Oracle error & None & 0.6458 & 0.7542 & 0.8988 & 0.5826 & 0.8875 & \cellcolor{green!15}\textbf{0.2914} & 0.0985 & 0.0587 & 0.1396 \\

GVL$^{2}$ & None & \cellcolor{yellow!20}0.6630 & 0.7546 & \cellcolor{yellow!20}0.9057 & \cellcolor{red!12}0.5992 & \cellcolor{yellow!20}0.8898 & 0.2849 & \cellcolor{red!12}0.1114 & 0.0688 & \cellcolor{red!12}0.1592 \\

RoboReward$^{1}$ & None & 0.6266 & 0.7158 & 0.8470 & \cellcolor{green!15}\textbf{0.6391} & 0.8861 & 0.2815 & 0.1113 & 0.0693 & 0.1554 \\

Robometer-Prog$^{3}$ & None & 0.5809 & 0.6825 & 0.8532 & 0.5840 & 0.8810 & 0.2627 & 0.1090 & 0.0690 & 0.1526 \\

Robometer-Pref$^{3}$ & None & 0.5919 & 0.7029 & 0.8443 & 0.5813 & 0.8859 & 0.2695 & 0.1046 & 0.0638 & 0.1524 \\

PRM-as-Judge$^{4}$ & None & 0.6467 & \cellcolor{red!12}0.7580 & \cellcolor{red!12}0.9033 & \cellcolor{yellow!20}0.6033 & \cellcolor{red!12}0.8880 & \cellcolor{red!12}0.2852 & \cellcolor{yellow!20}0.1154 & \cellcolor{yellow!20}0.0746 & \cellcolor{yellow!20}0.1606 \\

LRMs$^{5}$ & None & \cellcolor{red!12}0.6617 & \cellcolor{yellow!20}0.7603 & 0.8966 & 0.5744 & 0.8827 & 0.2756 & 0.0946 & 0.0587 & 0.1352 \\

\textbf{Confidence (Ours)} & \textbf{None} & \cellcolor{green!15}\textbf{0.6746} & \cellcolor{green!15}\textbf{0.7692} & \cellcolor{green!15}\textbf{0.9196} & 0.5923 & \cellcolor{green!15}\textbf{0.8903} & \cellcolor{yellow!20}0.2865 & \cellcolor{green!15}\textbf{0.1181} & \cellcolor{green!15}\textbf{0.0812} & \cellcolor{green!15}\textbf{0.1650} \\

\midrule
\rowcolor{gray!6}
\multicolumn{11}{c}{\textbf{Selection + additional weighting: extra scoring after selection}} \\
\midrule

Robometer-Prog$^{3}$ & Frame & 0.6439 & 0.7449 & 0.8882 & 0.5882 & \cellcolor{yellow!20}0.8893 & \cellcolor{yellow!20}0.2878 & \cellcolor{red!12}0.1097 & \cellcolor{red!12}0.0697 & \cellcolor{red!12}0.1563 \\

Robometer-Pref$^{3}$ & Frame & 0.6317 & 0.7289 & \cellcolor{red!12}0.8945 & 0.5289 & 0.8849 & 0.2665 & 0.0974 & 0.0654 & 0.1342 \\

PRM-as-Judge$^{4}$ & Frame & \cellcolor{red!12}0.6584 & \cellcolor{yellow!20}0.7661 & \cellcolor{yellow!20}0.9053 & \cellcolor{green!15}\textbf{0.6226} & 0.8868 & \cellcolor{red!12}0.2795 & 0.1082 & \cellcolor{yellow!20}0.0728 & 0.1536 \\

LRMs$^{5}$ & Frame & 0.6463 & 0.7448 & 0.8882 & 0.5826 & 0.8869 & 0.2750 & 0.1053 & 0.0635 & 0.1483 \\

\textbf{Confidence (Ours)} & \textbf{Frame} & \cellcolor{yellow!20}0.6595 & \cellcolor{red!12}0.7500 & \cellcolor{green!15}\textbf{0.9143} & \cellcolor{red!12}0.6171 & \cellcolor{red!12}0.8874 & 0.2789 & \cellcolor{yellow!20}0.1106 & 0.0684 & \cellcolor{yellow!20}0.1608 \\

\textbf{Confidence (Ours)} & \textbf{Fr.+Patch} & \cellcolor{green!15}\textbf{0.6772} & \cellcolor{green!15}\textbf{0.7758} & 0.8942 & \cellcolor{green!15}\textbf{0.6226} & \cellcolor{green!15}\textbf{0.8925} & \cellcolor{green!15}\textbf{0.2952} & \cellcolor{green!15}\textbf{0.1118} & \cellcolor{green!15}\textbf{0.0759} & \cellcolor{green!15}\textbf{0.1633} \\

\bottomrule
\end{tabular}}
\end{table*}

\paragraph{Ablation studies.}
Table~\ref{tab:selection_ablation} isolates the effect of the confidence-risk aggregation used for data selection, without additional frame or patch weighting. Mean risk gives the strongest overall result, achieving the best performance on seven of the nine component metrics, including both reconstruction metrics, Scene Consistency, Logics, and all three trajectory metrics. Tail risk obtains the highest Sem.-BLEU, while persistent risk obtains the highest Sem.-CLIP. 
We therefore use mean risk as the default acquisition criterion in our main experiments.

\begin{table*}[t]
\vspace{-0.3cm}
\caption{
\textbf{Ablation of selection criteria.}
All experiments use the default seed 42 and no additional weighting.
Green, yellow, and red indicate the best, second-best, and third-best results.
}
\label{tab:selection_ablation}
\centering
\scriptsize
\setlength{\tabcolsep}{3.0pt}
\renewcommand{\arraystretch}{1.04}
\resizebox{\textwidth}{!}{
\begin{tabular}{l||cc|c|ccc|ccc}
\toprule
\rowcolor{gray!12}
\multirow{2}{*}{Variant}
& \multicolumn{2}{c}{Reconstruction ($\uparrow$)}
& \multicolumn{1}{c}{Scene ($\uparrow$)}
& \multicolumn{3}{c}{Semantics ($\uparrow$)}
& \multicolumn{3}{c}{Motion ($\uparrow$)} \\
\cmidrule{2-10}
\rowcolor{gray!12}
& PSNR ($\uparrow$)
& SSIM ($\uparrow$)
& Scene Cons. ($\uparrow$)
& Logics ($\uparrow$)
& Sem.-CLIP ($\uparrow$)
& Sem.-BLEU ($\uparrow$)
& Traj-HSD ($\uparrow$)
& Traj-Dyn ($\uparrow$)
& Traj-nDTW ($\uparrow$) \\
\midrule

Random
& 0.5968
& 0.6849
& 0.8524
& 0.3554
& 0.8689
& 0.2372
& \cellcolor{red!12}0.1067
& \cellcolor{yellow!20}0.0725
& \cellcolor{red!12}0.1518 \\

\textbf{Mean Risk}
& \cellcolor{green!15}\textbf{0.6838}
& \cellcolor{green!15}\textbf{0.7797}
& \cellcolor{green!15}\textbf{0.9322}
& \cellcolor{green!15}\textbf{0.6198}
& \cellcolor{yellow!20}0.8894
& \cellcolor{red!12}0.2754
& \cellcolor{green!15}\textbf{0.1313}
& \cellcolor{green!15}\textbf{0.0921}
& \cellcolor{green!15}\textbf{0.1780} \\

Tail Risk
& \cellcolor{red!12}0.6317
& \cellcolor{red!12}0.7197
& \cellcolor{red!12}0.9218
& \cellcolor{yellow!20}0.5579
& \cellcolor{red!12}0.8890
& \cellcolor{green!15}\textbf{0.3047}
& \cellcolor{yellow!20}0.1207
& \cellcolor{red!12}0.0711
& \cellcolor{yellow!20}0.1630 \\

Persistent Risk
& \cellcolor{yellow!20}0.6529
& \cellcolor{yellow!20}0.7470
& \cellcolor{yellow!20}0.9230
& \cellcolor{red!12}0.5496
& \cellcolor{green!15}\textbf{0.8928}
& \cellcolor{yellow!20}0.2858
& 0.0926
& 0.0608
& 0.1342 \\

\bottomrule
\end{tabular}}
\vspace{-0.2cm}
\end{table*}

%% file: sections/5_conclusion.tex
\section{Conclusion}

We presented ConfAL-WM, a confidence-guided active learning framework for post-training action-conditioned world models. By attaching a lightweight confidence probe to UNet decoder features, our method supports task-level data selection as well as frame- and patch-level weighted retraining. Experiments on RoboTwin2.0 show that confidence provides an effective risk signal and improves reconstruction, scene consistency, and semantic quality over scalar scoring baselines.

Our current approach still has several limitations. First, the confidence probe is designed around the internal decoder features of a UNet diffusion backbone and is trained only with EVAC on RoboTwin2.0, making it difficult to transfer directly across world-model architectures and domains. Future work could train a general-purpose world confidence model using multiple backbone world models and diverse embodied datasets. Such a model would also require a more universal input interface. 
Second, the probe-design ablations use cached single-step features. Evaluating confidence along free-running rollouts and jointly varying probe design, acquisition, and loss weighting would clarify how local error estimates affect post-training performance.
Third, current world-model evaluation remains incomplete, particularly because visual fidelity and motion accuracy may conflict. Future protocols could separately evaluate robot-arm motion, manipulated-object dynamics, interaction regions, and static backgrounds, providing a more precise account of both local reconstruction quality and action-conditioned physical evolution.

%% file: sections/6_Checklist.tex

\section*{AI Use Statement}

Generative AI tools were used throughout this work as research assistants in three main aspects.
First, they were used to aid and polish manuscript writing, including language editing,
reorganization, and improvements to presentation and readability.
Second, they assisted with retrieval and discovery of relevant literature and baseline methods;
all cited papers, bibliographic information, and related-work claims were subsequently checked
and verified by the authors.
Third, they assisted with research ideation and execution, including feedback on experimental
design, preparation and debugging of experiment and evaluation scripts and commands,
organization and interpretation of experimental results, and preparation of the released code,
datasets, checkpoints, and associated documentation.

All AI-assisted experimental procedures were reviewed and executed by the authors.
Reported numerical results were obtained from the actual experimental runs and were checked
against the corresponding logs and evaluation artifacts.
AI-assisted code and release materials were manually inspected and verified before use or
publication.
The authors take full responsibility for the final content of this work, including its text,
scientific claims, experimental results, code, and released artifacts.

\section*{Reproducibility Statement}

We provide the implementation, model checkpoints, data and evaluation artifacts necessary to
reproduce the main results of this work.
The project page\footnote{\url{https://confal-wm.github.io/}}
provides an overview of the method, experimental results, confidence visualizations, and links
to all released resources.
The source code and experimental pipeline are available in the code
repository\footnote{\url{https://github.com/ConfAL-WM/ConfAL-WM}}.
We additionally release the EVAC-v1 and EVAC-v2 checkpoints, confidence probes, and
trajectory detector through the model
repository\footnote{\url{https://huggingface.co/anonymous89793/ConfAL-WM}},
together with the corresponding data and evaluation
artifacts\footnote{\url{https://huggingface.co/datasets/anonymous89793/ConfAL-WM-Dataset}}.

The released artifacts include task-level prescreening outputs, EVAC-v1 inference results,
dense confidence maps and metadata used for confidence-guided retraining, baseline scoring
artifacts, trajectory annotations, and the evaluation tables and bootstrap statistics reported
in the paper.
Detailed model architecture, confidence-target construction, active-learning procedure,
training objectives, implementation settings, random seeds, parameter sensitivity analyses,
and seed-wise numerical results are further documented in the main paper and appendices.

\section*{Ethics Statement}

This work aims to improve the data efficiency and reliability of embodied world-model
post-training by focusing computation and supervision on informative data and unreliable
spatiotemporal regions.
The experiments do not involve human subjects or personally identifiable information.
More accurate action-conditioned world models may benefit robotic simulation, planning,
policy evaluation, and synthetic-data generation.
However, predictions produced by such models are not guaranteed to correspond to physically
valid outcomes, particularly under distribution shifts or in previously unseen environments.
They should therefore not be treated as substitutes for real-world validation in safety-critical
robotic applications.
The released code, checkpoints, and derived artifacts are intended for research use and should
be used in accordance with the licenses and usage conditions of the corresponding upstream
models and datasets.

%% file: sections/7_appendix_A.tex
\section{More Details on Methods}
\label{app:confidence_details}

\subsection{Patch-level Error Construction}
\label{app:patch_error}

\begin{figure}[h]
    \centering
    \includegraphics[width=1\linewidth]{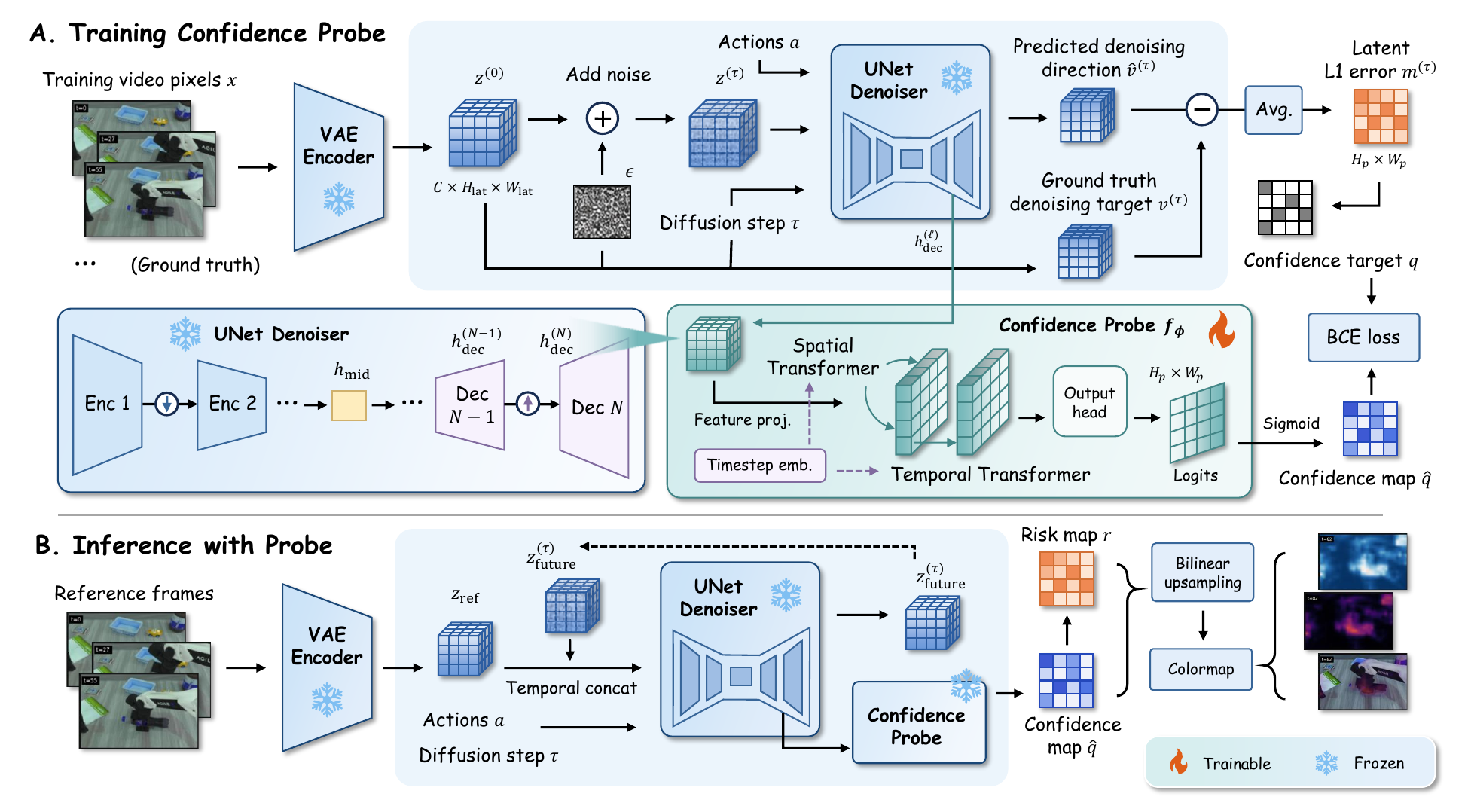}
    \vspace{-0.7cm}
    \caption{Training and inference of the confidence probe in the UNet latent diffusion world model. The figure follows the same training and inference pipeline as Figure~\ref{fig:confidence_probe_arch}.}
    \label{fig:app:confidence_probe_arch}
\end{figure}

Figure~\ref{fig:app:confidence_probe_arch} illustrates the overall architecture used to construct the patch-level confidence target.

\paragraph{Latent diffusion and velocity target.}
Given a ground-truth training video $x$, the frozen VAE encoder first maps it into a clean latent representation
\begin{equation}
    z^{(0)}=\mathrm{Enc}_{\mathrm{VAE}}(x),
    \qquad
    z^{(0)}\in\mathbb{R}^{C\times H_{\mathrm{lat}}\times W_{\mathrm{lat}}},
\end{equation}
where $C$ denotes the latent channel dimension and $(H_{\mathrm{lat}},W_{\mathrm{lat}})$ denotes the spatial resolution of the latent grid. For simplicity, the batch and video-time dimensions are omitted in this subsection.

At diffusion timestep $\tau$, Gaussian noise $\varepsilon\sim\mathcal{N}(0,I)$ is sampled and added according to the forward diffusion schedule:
\begin{equation}
    z^{(\tau)}
    =
    \sqrt{\bar{\alpha}_\tau}\,z^{(0)}
    +
    \sqrt{1-\bar{\alpha}_\tau}\,\varepsilon,
\label{z^(tau)}
\end{equation}
where $\bar{\alpha}_\tau$ is the cumulative noise-schedule coefficient at timestep $\tau$. The action-conditioned UNet predicts the corresponding velocity parameterization as
\begin{equation}
    \hat{v}^{(\tau)}
    =
    \mathrm{UNet}_{\Theta}
    \left(
    z^{(\tau)},a,e_\tau
    \right),
    \qquad
    \hat{v}^{(\tau)}
    \in
    \mathbb{R}^{C\times H_{\mathrm{lat}}\times W_{\mathrm{lat}}},
\end{equation}
where $a$ is the robot-action condition and $e_\tau$ is the diffusion-timestep embedding. Under the velocity-prediction formulation, the ground-truth denoising target is analytically constructed as
\begin{equation}
    v^{(\tau)}
    =
    \sqrt{\bar{\alpha}_\tau}\,\varepsilon
    -
    \sqrt{1-\bar{\alpha}_\tau}\,z^{(0)},
    \qquad
    v^{(\tau)}
    \in
    \mathbb{R}^{C\times H_{\mathrm{lat}}\times W_{\mathrm{lat}}}.
\label{v^(tau)}
\end{equation}
Therefore, $\hat{v}^{(\tau)}$ and $v^{(\tau)}$ have the same tensor shape and lie in the same velocity-prediction latent space. The action $a$ and timestep embedding $e_\tau$ condition the prediction $\hat{v}^{(\tau)}$, whereas the target $v^{(\tau)}$ is directly determined by $z^{(0)}$, $\varepsilon$, and the diffusion schedule.

\paragraph{Equivalent clean-latent reconstruction.}
Rather than executing the complete reverse diffusion trajectory, we can directly recover the predicted clean latent from the current noisy latent and the velocity prediction:
\begin{equation}
\hat{z}^{(\tau)}
=
\sqrt{\bar{\alpha}_{\tau}}\,z^{(\tau)}
-
\sqrt{1-\bar{\alpha}_{\tau}}\,\hat{v}^{(\tau)}.
\label{hat z^(tau)}
\end{equation}
This provides an equivalent way to measure the local prediction error in the clean-latent space. Specifically, substituting~\eqref{z^(tau)} and~\eqref{hat z^(tau)} into
$\left|\hat{z}^{(\tau)}-z^{(0)}\right|$ gives
\begin{equation}
\begin{split}
\left|
\hat{z}^{(\tau)}-z^{(0)}
\right|
=&
\left|
\sqrt{\bar{\alpha}_{\tau}}\,z^{(\tau)}
-
\sqrt{1-\bar{\alpha}_{\tau}}\,\hat{v}^{(\tau)}
-
z^{(0)}
\right|
\\
=&
\left|
\left(\bar{\alpha}_{\tau}-1\right)z^{(0)}
+
\sqrt{\bar{\alpha}_{\tau}\left(1-\bar{\alpha}_{\tau}\right)}\,\varepsilon
-
\sqrt{1-\bar{\alpha}_{\tau}}\,\hat{v}^{(\tau)}
\right|
\\
=&
\sqrt{1-\bar{\alpha}_{\tau}}
\left|
\sqrt{1-\bar{\alpha}_{\tau}}\,z^{(0)}
-
\sqrt{\bar{\alpha}_{\tau}}\,\varepsilon
+
\hat{v}^{(\tau)}
\right|.
\end{split}
\end{equation}
Using the velocity target in~\eqref{v^(tau)}, we therefore obtain
\begin{equation}
\left|
\hat{z}^{(\tau)}-z^{(0)}
\right|
=
\sqrt{1-\bar{\alpha}_{\tau}}
\left|
\hat{v}^{(\tau)}-v^{(\tau)}
\right|.
\label{eq:latent_velocity_error_equivalence}
\end{equation}
Hence, at a fixed diffusion timestep $\tau$, the clean-latent reconstruction error and the velocity-prediction error differ only by the scalar factor
$\sqrt{1-\bar{\alpha}_{\tau}}$. This allows confidence supervision to use velocity-prediction error directly, without running the full reverse diffusion process.

\begin{wrapfigure}{r}{0.45\textwidth}
    \vspace{-0.6cm}
    \centering
    \includegraphics[width=\linewidth]{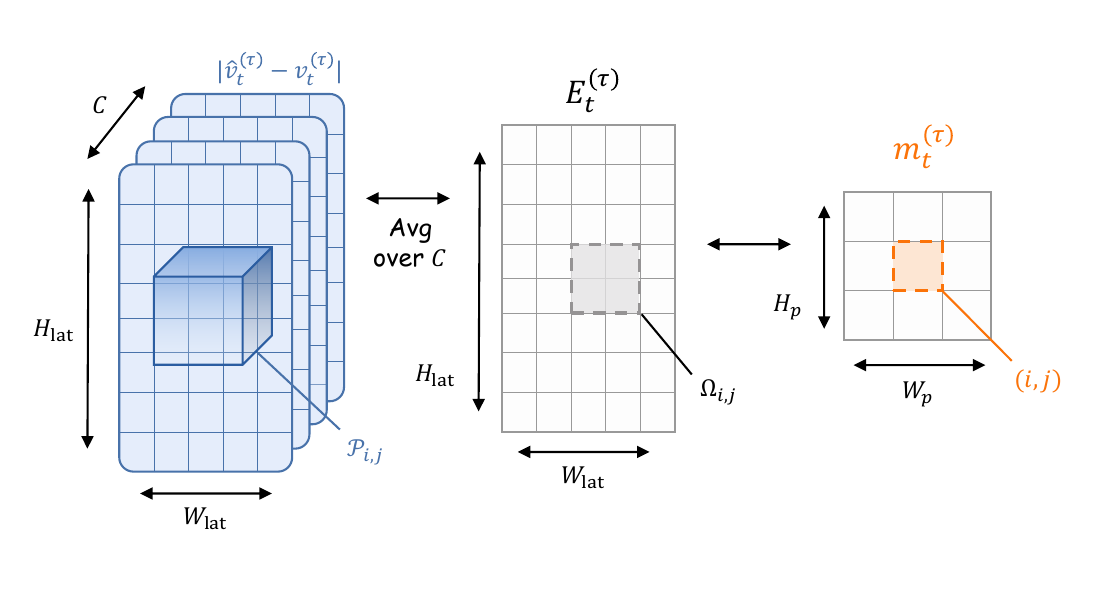}
    \vspace{-0.8cm}
\caption{
\textbf{Patch-level latent error construction.}
Channel-wise latent prediction errors are first averaged into a spatial error map and then pooled over the region aligned with each confidence-probe patch.
}
    \label{fig:app_tensor_feature_map}
    \vspace{-2cm}
\end{wrapfigure}

\paragraph{Patch-level latent prediction error.}
We now restore the frame index $t$. For each latent spatial location
$(x,y)$, we first average the absolute clean-latent reconstruction error over
the $C$ latent channels:
\begin{equation}
\begin{split}
E_{t,(x,y)}^{(\tau)}
&=
\frac{\sqrt{1-\bar{\alpha}_{\tau}}}{C}
\sum_{c=1}^{C}
\left|
\hat{v}_{t,p}^{(\tau)}
-
v_{t,p}^{(\tau)}
\right|\\
&=
\frac{1}{C}
\sum_{c=1}^{C}
\left|
\hat{z}_{t,p}^{(\tau)}
-
z_{t,p}^{(0)}
\right|.
\end{split}
\label{eq:spatial_latent_error}
\end{equation}
Here $p=(c,x,y)$ denotes a latent entry; $c$ indexes the latent channel and $(x,y)$ denotes the spatial location
on the latent grid. Thus,
$E_t^{(\tau)}\in\mathbb{R}^{H_{\mathrm{lat}}\times W_{\mathrm{lat}}}$
forms a spatial error map for frame $t$.

As illustrated in Figure~\ref{fig:app_tensor_feature_map}, let
$\Omega_{i,j}$ denote the latent spatial region aligned with confidence-probe
output location $(i,j)$. The corresponding patch-level error is obtained by
spatially averaging $E_t^{(\tau)}$ over this region. Equivalently, defining
\begin{equation}
\mathcal{P}_{i,j}
=
\{1,\ldots,C\}\times\Omega_{i,j},
\label{eq:latent_patch_set}
\end{equation}
the same quantity can be written directly over all latent entries within the
patch:
\begin{equation}
\begin{split}
m_{t,(i,j)}^{(\tau)}
&=
\frac{1}{|\Omega_{i,j}|}
\sum_{(x,y)\in\Omega_{i,j}}
E_{t,(x,y)}^{(\tau)}
\\
&=
\frac{1}{C\cdot|\Omega_{i,j}|}
\sum_{(x,y)\in\Omega_{i,j}}
\sum_{c=1}^{C}
\left|
\hat{z}_{t,p}^{(\tau)}
-
z_{t,p}^{(0)}
\right|
\\
&=
\frac{1}{|\mathcal{P}_{i,j}|}
\sum_{p\in\mathcal{P}_{i,j}}
\left|
\hat{z}_{t,p}^{(\tau)}
-
z_{t,p}^{(0)}
\right|.
\end{split}
\label{eq:patch_latent_error}
\end{equation}
Hence,
$m_t^{(\tau)}
=
\{m_{t,(i,j)}^{(\tau)}\}_{i,j}
\in\mathbb{R}^{H_p\times W_p}$
has the same spatial indexing as the confidence output.

\paragraph{Binary confidence target.}
The binary supervision target is obtained by comparing each local prediction
error with the adaptive threshold $\theta_t$:
\begin{equation}
q_{t,(i,j)}
=
\mathbb{I}
\left\{
m_{t,(i,j)}^{(\tau)} < \theta_t
\right\},
\qquad
q_t
=
\left\{
q_{t,(i,j)}
\right\}_{i,j}
\in
\{0,1\}^{H_p\times W_p}.
\label{eq:binary_confidence_target}
\end{equation}
A value $q_{t,(i,j)}=1$ indicates that the corresponding latent patch is
treated as reliable, while $q_{t,(i,j)}=0$ denotes a high-error prediction.
The adaptive construction of $\theta_t$ is described in the following
subsection.

\paragraph{UNet feature tapping and confidence prediction.}
The denoising UNet contains multiple encoder blocks, a bottleneck, and multiple
decoder blocks:
\begin{equation}
z^{(\tau)}
\rightarrow
h_{\mathrm{enc}}^{(1)}
\rightarrow \cdots \rightarrow
h_{\mathrm{enc}}^{(K)}
\rightarrow
h_{\mathrm{mid}}
\rightarrow
h_{\mathrm{dec}}^{(1)}
\rightarrow \cdots \rightarrow
h_{\mathrm{dec}}^{(L)}
\rightarrow
\hat{v}^{(\tau)}.
\end{equation}
Our confidence probe taps a selectable decoder feature
$h_{\mathrm{dec}}^{(\ell)}$, where the layer index $\ell$ controls the balance
between global context and spatial locality. Earlier decoder layers provide
coarser, more global representations, whereas later layers retain finer spatial
information.

For future frame $t$, the dense confidence prediction is
\begin{equation}
\hat{q}_t
=
\sigma
\left(
f_{\phi}
\left(
h_{\mathrm{dec}}^{(\ell)},
e_{\tau}, e_\theta
\right)
\right),
\qquad
\hat{q}_t
\in
[0,1]^{H_p\times W_p},
\label{eq:dense_confidence_prediction}
\end{equation}
where $e_\tau$ denotes the diffusion-timestep embedding, $e_\theta$ denotes the sampled error threshold embedding, $f_{\phi}$ is the trainable confidence probe and $\sigma(\cdot)$ denotes
the element-wise sigmoid function. Each output location $(i,j)$ is spatially
aligned with $\Omega_{i,j}$, and therefore predicts the confidence associated
with the local error $m_{t,(i,j)}^{(\tau)}$.

\subsection{EMA-Calibrated Random Threshold Supervision}
\label{app:ema_threshold}

The numerical scale of the local prediction error can vary across datasets and training regimes, making a fixed confidence threshold difficult to transfer. We therefore estimate the threshold interval online from the batch-wise error distribution and stabilize it using exponential moving averages (EMA). For notational simplicity, the derivation below considers a fixed training sample and omits the sample index; the percentile statistics $p_l^{(s)}$ and $p_h^{(s)}$ are still computed from all local errors in the current training batch.

\paragraph{Batch-wise percentile estimation.}
At training step $s$, let $m^{(s)}$ denote the collection of patch-level latent errors in the current batch, where each element follows the definition of $m_{t,(i,j)}^{(\tau)}$ in Equation~\ref{eq:patch_latent_error}. We estimate the lower and upper error statistics as
\begin{equation}
p_l^{(s)}
=
\operatorname{Quantile}_{0.1}\!\left(m^{(s)}\right),
\qquad
p_h^{(s)}
=
\operatorname{Quantile}_{0.9}\!\left(m^{(s)}\right).
\label{eq:batch_error_quantiles}
\end{equation}
These percentiles adapt the supervision range to the current error scale without manually specifying dataset-dependent MAE thresholds.

\paragraph{Two-stage EMA update.}
The lower and upper threshold bounds are updated as
\begin{equation}
l^{(s)}
=
(1-\gamma_s)l^{(s-1)}
+
\gamma_s p_l^{(s)},
\qquad
h^{(s)}
=
(1-\gamma_s)h^{(s-1)}
+
\gamma_s p_h^{(s)}.
\label{eq:ema_bounds_update}
\end{equation}
We use a two-stage update schedule,
\begin{equation}
\gamma_s
=
\begin{cases}
0.20, & s<S_{\mathrm{warm}},\\
0.002, & s\ge S_{\mathrm{warm}},
\end{cases}
\qquad
\beta_s=1-\gamma_s
=
\begin{cases}
0.80, & s<S_{\mathrm{warm}},\\
0.998, & s\ge S_{\mathrm{warm}},
\end{cases}
\label{eq:two_stage_ema}
\end{equation}
where $\gamma_s$ is the injection rate of the current batch statistics and $\beta_s$ is the equivalent EMA momentum. Thus, the conventional EMA form is
$\mu^{(s)}=\beta_s\mu^{(s-1)}+(1-\beta_s)p^{(s)}$.
The larger update rate during warmup rapidly adapts the threshold range to the target-domain error scale, while the smaller rate afterwards tracks slower distributional changes with stronger smoothing.

\paragraph{Random threshold sampling and interpretation.}
After updating the bounds, a separate threshold is sampled for each predicted future frame $t$:
\begin{equation}
\theta_t^{(s)}
\sim
\mathcal{U}\!\left(l^{(s)},h^{(s)}\right),
\qquad
q_{t,(i,j)}^{(s)}
=
\mathbb{I}
\left\{
m_{t,(i,j)}^{(\tau)}
<
\theta_t^{(s)}
\right\}.
\label{eq:random_threshold_target}
\end{equation}
The same $\theta_t^{(s)}$ is shared by all spatial patches $(i,j)$ within frame $t$, while different future frames receive independently sampled thresholds.

For a fixed local error $m$, the binary target becomes random only through
$\theta_t^{(s)}$. Since
$\theta_t^{(s)}\sim\mathcal{U}(l^{(s)},h^{(s)})$, its density is
$1/(h^{(s)}-l^{(s)})$ over the current threshold interval. Marginalizing over the sampled threshold gives
\begin{equation}
\begin{split}
\mathbb{E}_{\theta}[q\mid m]
&=
\mathbb{P}_{\theta}(m<\theta)
=
\int_{l^{(s)}}^{h^{(s)}}
\mathbb{I}\{m<\theta\}
\frac{1}{h^{(s)}-l^{(s)}}\,\mathrm{d}\theta
\\
&=
\frac{
\left[
h^{(s)}-\max\{m,l^{(s)}\}
\right]_+
}{
h^{(s)}-l^{(s)}
}
=
\operatorname{clamp}
\left(
\frac{h^{(s)}-m}
{h^{(s)}-l^{(s)}},
0,1
\right)
\\
&=
\begin{cases}
1,
& m\le l^{(s)},\\[1mm]
\dfrac{h^{(s)}-m}{h^{(s)}-l^{(s)}},
& l^{(s)}<m<h^{(s)},\\[3mm]
0,
& m\ge h^{(s)}.
\end{cases}
\end{split}
\label{eq:expected_random_threshold_target}
\end{equation}
Hence, although each training target is binary, averaging over random thresholds induces a continuous target that decreases monotonically with prediction error.

In the threshold-marginalized case, the optimal prediction under binary cross-entropy is this conditional probability (using $0\log 0=0$):
\begin{equation}
\begin{split}
\hat q^{*}(m)
&=
\arg\min_{\hat q\in[0,1]}
\mathbb{E}_{q\mid m}
\left[
-q\log\hat q-(1-q)\log(1-\hat q)
\right]
\\
&=
\mathbb{E}[q\mid m]
=
\operatorname{clamp}
\left(
\frac{h^{(s)}-m}{h^{(s)}-l^{(s)}},
0,1
\right).
\end{split}
\label{eq:bce_optimal_confidence}
\end{equation}
In our implementation, $\theta_t^{(s)}$ is additionally embedded into the confidence probe as described in the main method, so the probe learns threshold-conditioned reliability rather than only this marginalized form. Equation~\ref{eq:expected_random_threshold_target} describes the threshold-marginalized target and does not determine the threshold-conditioned probe output.

\begin{figure*}[h]
\centering
\begin{minipage}[t]{0.47\textwidth}
\centering
\includegraphics[width=\linewidth]{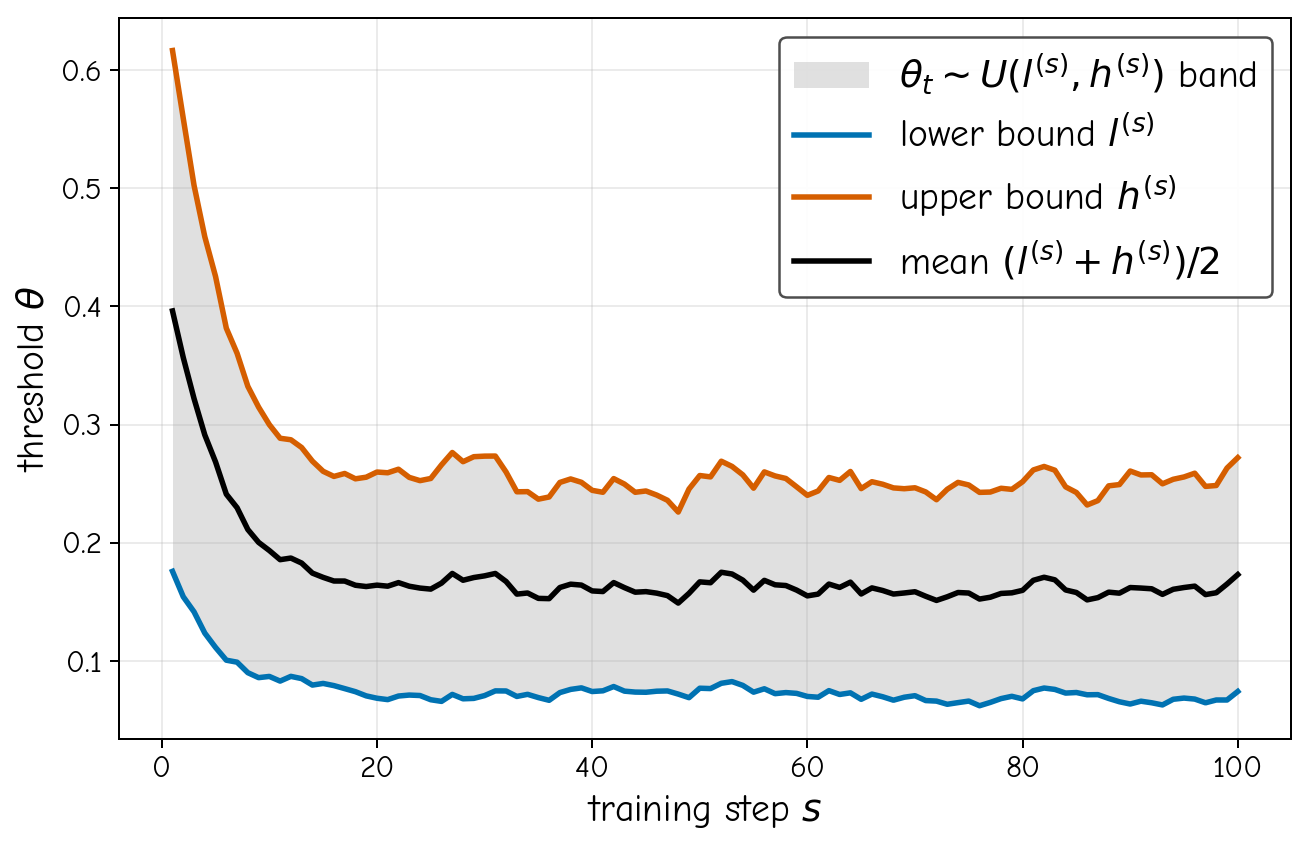}
\centerline{\small (a) Zoom-in of the first 100 training steps}
\end{minipage}
\hfill
\begin{minipage}[t]{0.47\textwidth}
\centering
\includegraphics[width=\linewidth]{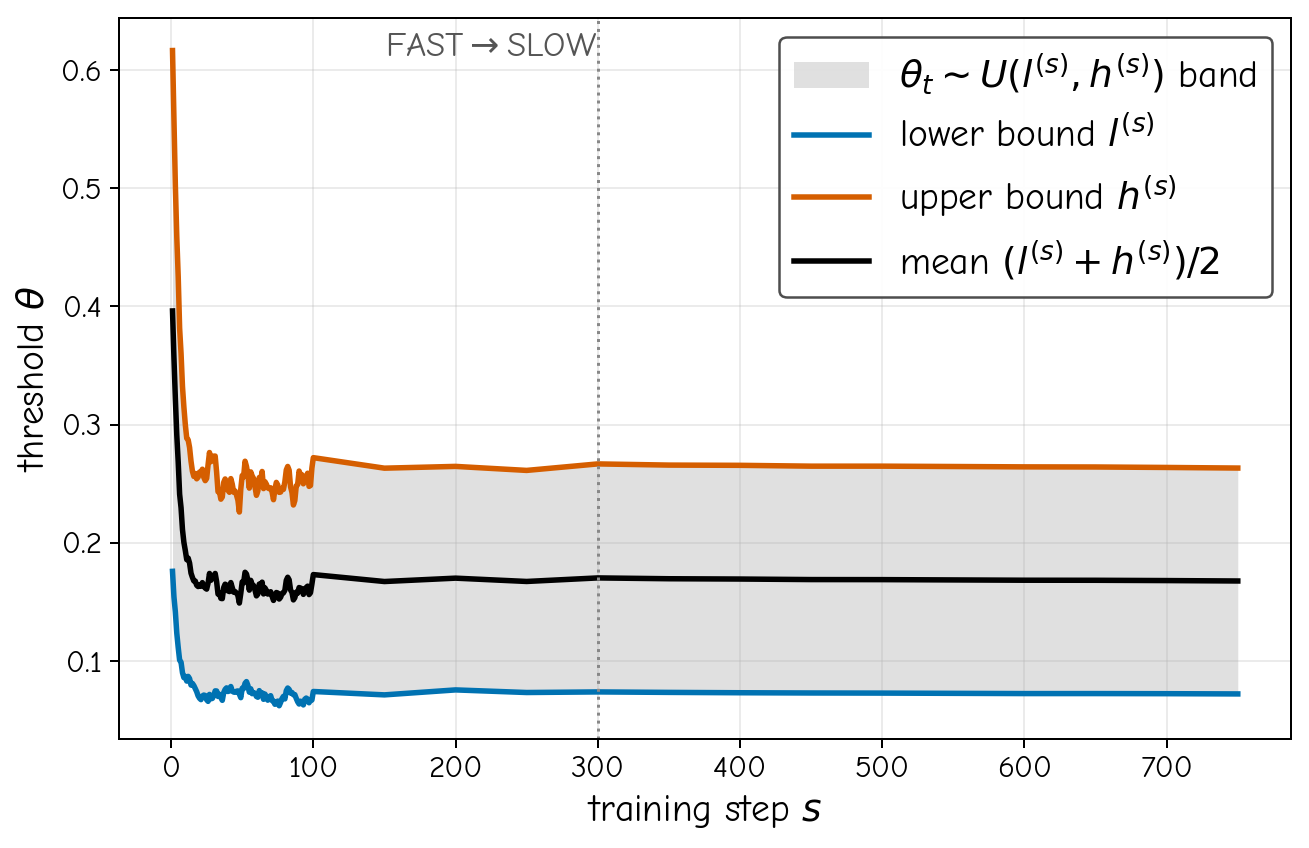}
\centerline{\small (b) Extended training-step evolution}
\end{minipage}
\caption{
\textbf{Dynamics of the EMA-based adaptive threshold band.}
The probe uses a stochastic threshold $\theta \sim \mathcal{U}(l^{(s)}, h^{(s)})$, where $l^{(s)}$ and $h^{(s)}$ are the EMA-tracked lower and upper bounds at training step $s$, and the black curve shows their midpoint $(l^{(s)}+h^{(s)})/2$.
(a) During the first 100 steps, both bounds decrease rapidly and quickly approach a stable range, reflecting fast adaptation to the current error scale.
(b) Over a longer set of sampled training steps, the threshold band shows a clear transition from a fast-changing warmup regime to a slower and more stable regime.
}
\label{fig:ema_threshold_dynamics}
\vspace{-0.2cm}
\end{figure*}

\paragraph{Implementation and EMA dynamics.}
For RoboTwin2.0, we initialize $l^{(0)}=0.20$ and $h^{(0)}=0.70$, use $S_{\mathrm{warm}}=300$, and continue updating the EMA bounds throughout the $6000$-step probe training. The bounds eventually stabilize around $l\approx0.0705$ and $h\approx0.2610$, showing how adaptive supervision automatically tracks the local error scale as training proceeds on this dataset and backbone.

Figure~\ref{fig:ema_threshold_dynamics} visualizes the adaptation process. During the initial stage, the threshold band rapidly contracts toward the observed target-domain error scale. Over a longer range of sampled training steps, the evolution becomes substantially slower and more stable, showing the intended transition from fast initialization to long-term tracking.

%% file: sections/7_appendix_B.tex
\section{More Experimental Results}
\label{app:more_experiments}

\subsection{Additional Confidence Evaluation Details}
\label{app:Additional Confidence Evaluation Details}

This subsection supplements the confidence evaluation in
Section~\ref{subsec:why_confidence}. We report additional results for alternative
risk aggregations, spatial localization, operational calibration, and parameter
sensitivity. Latent-space errors remain our primary oracle, while pixel-space
results are included as an external visual-domain reference.

\begin{figure*}[h]
    \centering
    \vspace{-0.2cm}
    \includegraphics[width=\textwidth]
    {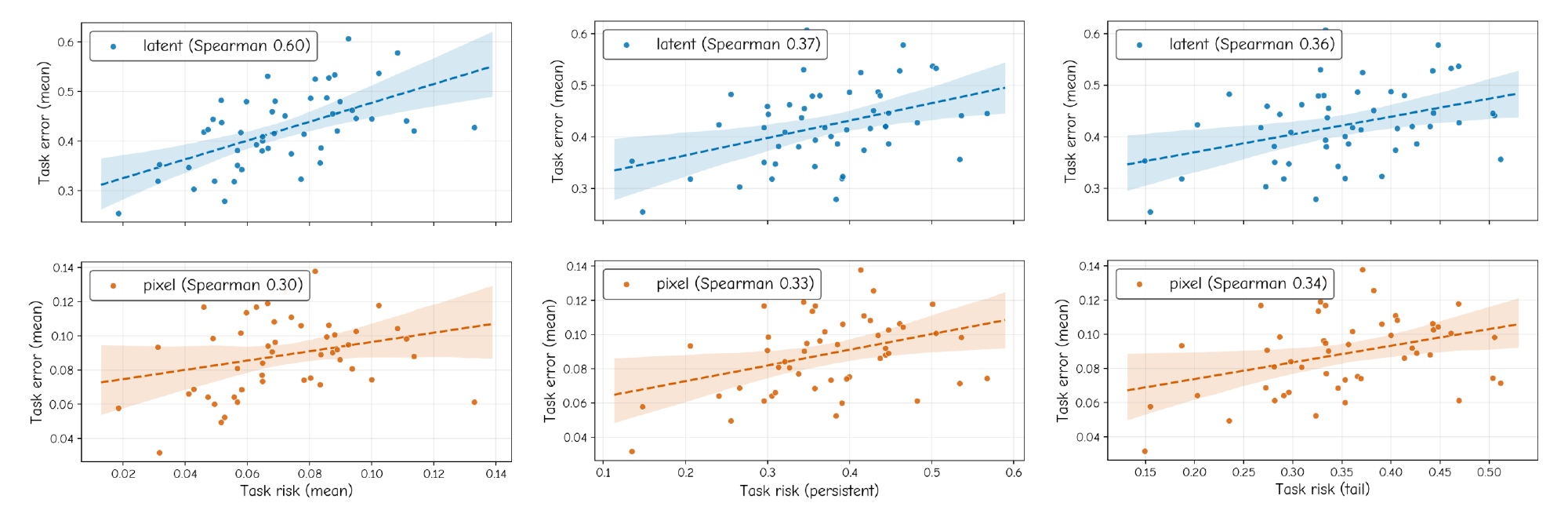}
    \vspace{-0.6cm}
    \caption{
    \textbf{Task-level risk aggregation in latent and pixel spaces.}
    Mean, tail, and persistent risk scores are compared with the mean
    oracle-error statistics over the 50 prescreened tasks. Dashed lines show
    linear fits, shaded regions denote confidence intervals, and each panel
    reports the Spearman correlation. Mean aggregation gives the strongest
    latent-space correlation, whereas tail and persistent aggregation emphasize
    different failure patterns.
    }
    \label{fig:app_confidence_trajectory}
    \vspace{-0.4cm}
\end{figure*}

\paragraph{Alternative task-level risk aggregations.}
Figure~\ref{fig:app_confidence_trajectory} compares mean, tail, and persistent
risk aggregation in latent and pixel spaces. Mean risk achieves the strongest
latent-space Spearman correlation because averaging suppresses local noise and measures
the overall prediction difficulty of an episode. Tail risk instead emphasizes
sparse severe errors, while persistent risk targets failures that recur across
multiple frames; consequently, they need not correlate as strongly with the
episode-wide mean error. These correlations therefore verify that each score
contains a meaningful error signal, but do not by themselves determine which
acquisition rule yields the best post-training model. Their downstream
differences are evaluated separately in Table~\ref{tab:selection_ablation}.

\begin{figure*}[h]
    \centering
    \begin{minipage}[h]{0.4\textwidth}
        \centering
        \includegraphics[width=\linewidth]
        {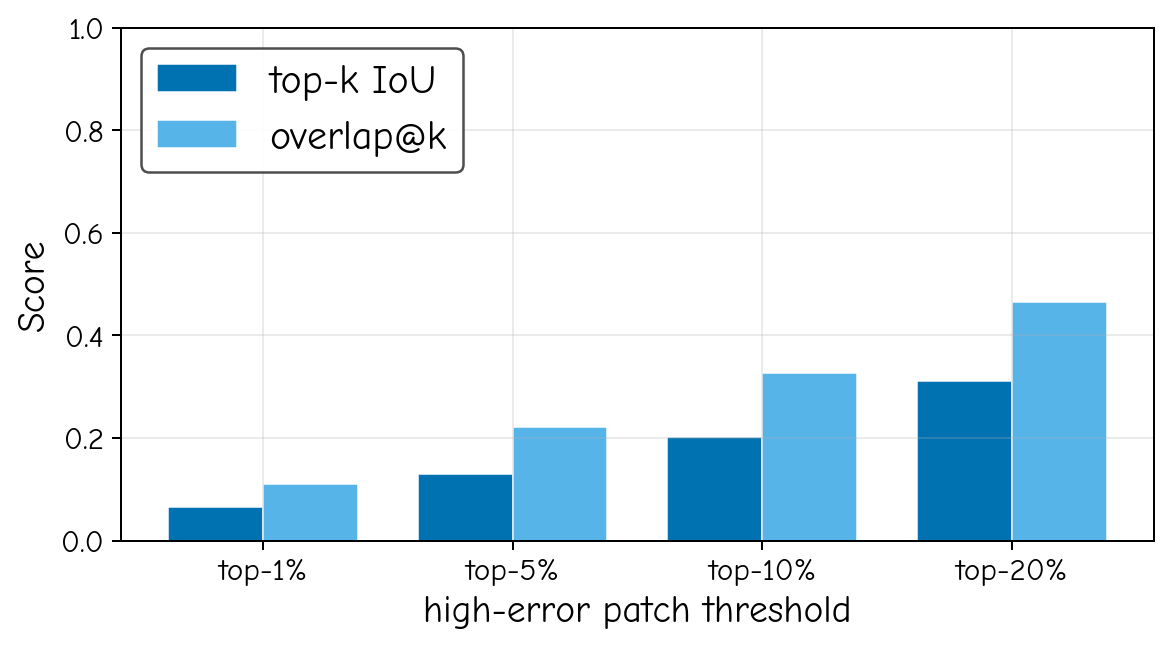}
        \vspace{-0.2cm}
        \centerline{\small (a) Latent-space localization}
    \end{minipage}
    \qquad
    \begin{minipage}[h]{0.4\textwidth}
        \centering
        \includegraphics[width=\linewidth]
        {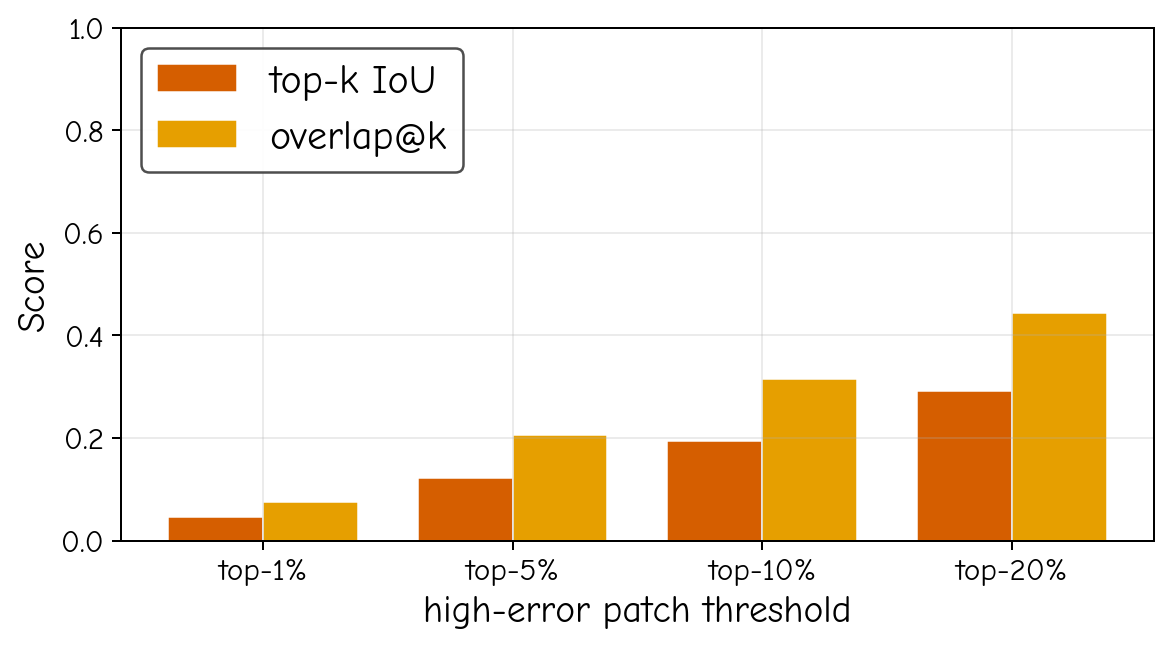}
        \vspace{-0.2cm}
        \centerline{\small (b) Pixel-space localization}
    \end{minipage}
    \vspace{0.15cm}
    \caption{
    \textbf{Spatial agreement between high-risk and high-error patches.}
    Top-$k$ IoU and overlap@$k$ are reported for the top-$1\%$, $5\%$, $10\%$,
    and $20\%$ patches in (a) latent and (b) pixel spaces. The increasing scores
    at larger ratios reflect coarser spatial coverage and should not be
    interpreted as improved localization resolution.
    }
    \label{fig:app_confidence_localization}
    \vspace{-0.4cm}
\end{figure*}

\paragraph{Spatial localization agreement.}
We further compare the spatial locations selected by risk and oracle error.
For each ratio $k$, top-$k$ IoU measures the intersection-over-union between
the highest-risk and highest-error patch sets, while overlap@$k$ measures the
fraction of high-error patches recovered by the high-risk set.
Figure~\ref{fig:app_confidence_localization} shows that both metrics increase
as the selected region becomes larger. At top-$5\%$, latent-space localization
reaches approximately $0.13$ IoU and $0.22$ overlap, with slightly lower values
in pixel space. The limited spatial overlap and moderate IoU indicate that
risk identifies some relevant error regions without exactly reproducing their
boundaries. This supports soft patch weighting rather than hard spatial
selection.

\newpage
\begin{figure*}[h]
    \centering
    \vspace{-0.1cm}
    \begin{minipage}[h]{0.4\textwidth}
        \centering
        \includegraphics[width=\linewidth]{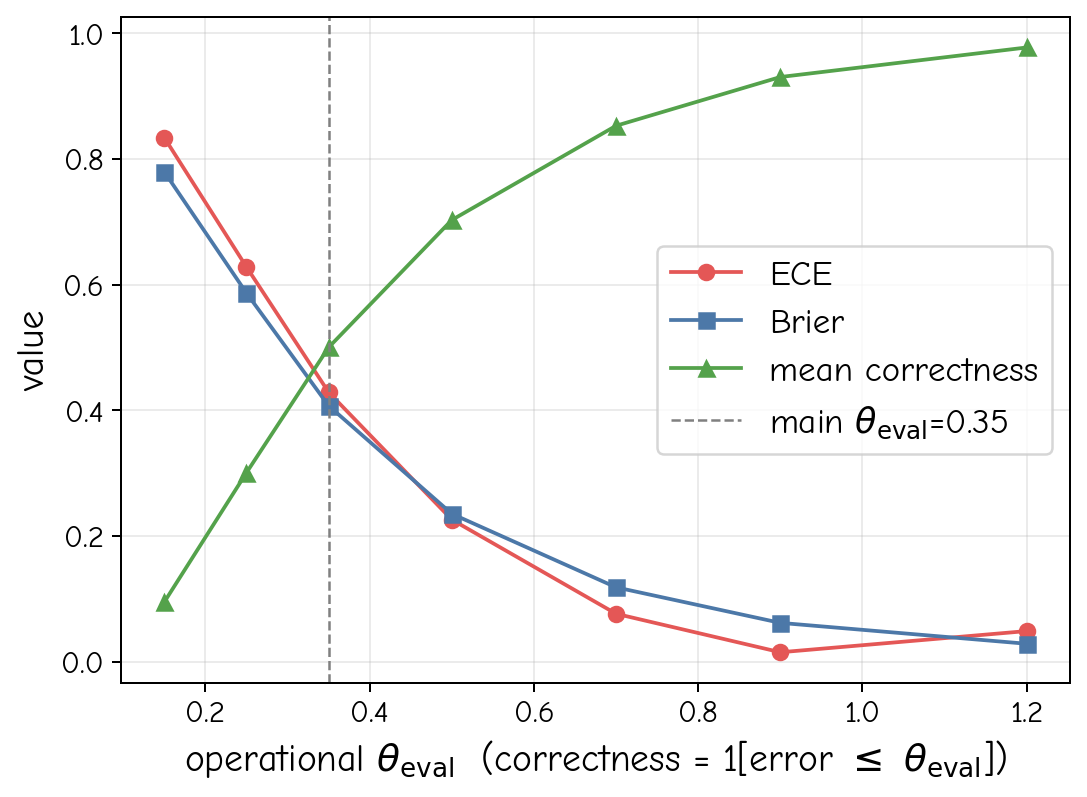}
        \vspace{-0.15cm}
        \centerline{\small (a) Latent-space calibration sweep}
    \end{minipage}
    \qquad
    \begin{minipage}[h]{0.4\textwidth}
        \centering
        \includegraphics[width=\linewidth]{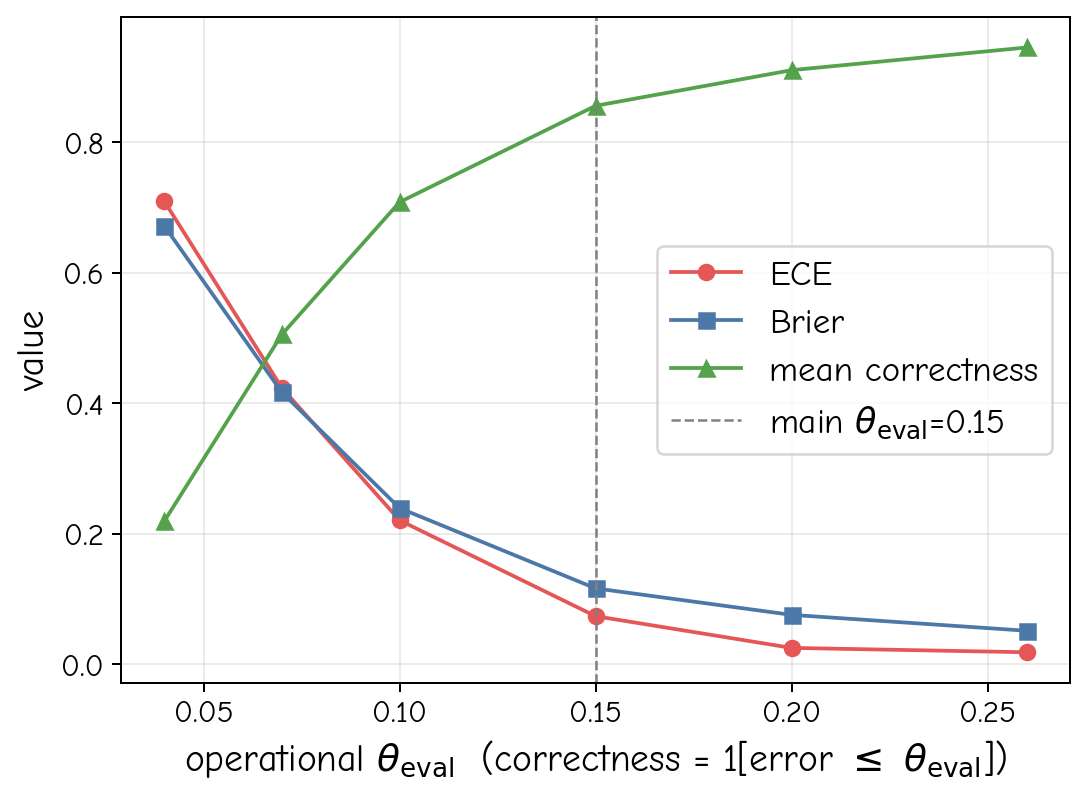}
        \vspace{-0.15cm}
        \centerline{\small (b) Pixel-space calibration sweep}
    \end{minipage}
    \vspace{0.cm}
    \caption{
    \textbf{ECE and Brier scores under different operational thresholds.}
    Calibration is evaluated by sweeping the correctness threshold $\theta_{\mathrm{eval}}$ in
    (a) latent space and (b) pixel space. The confidence maps are fixed throughout; only the binary correctness definition changes.
    }
    \label{fig:app_confidence_calibration_curve}
    \vspace{-0.1cm}
\end{figure*}

\begin{figure*}[h]
    \centering
    \vspace{-0.1cm}
    \begin{minipage}[h]{0.88\textwidth}
        \centering
        \includegraphics[width=\linewidth]{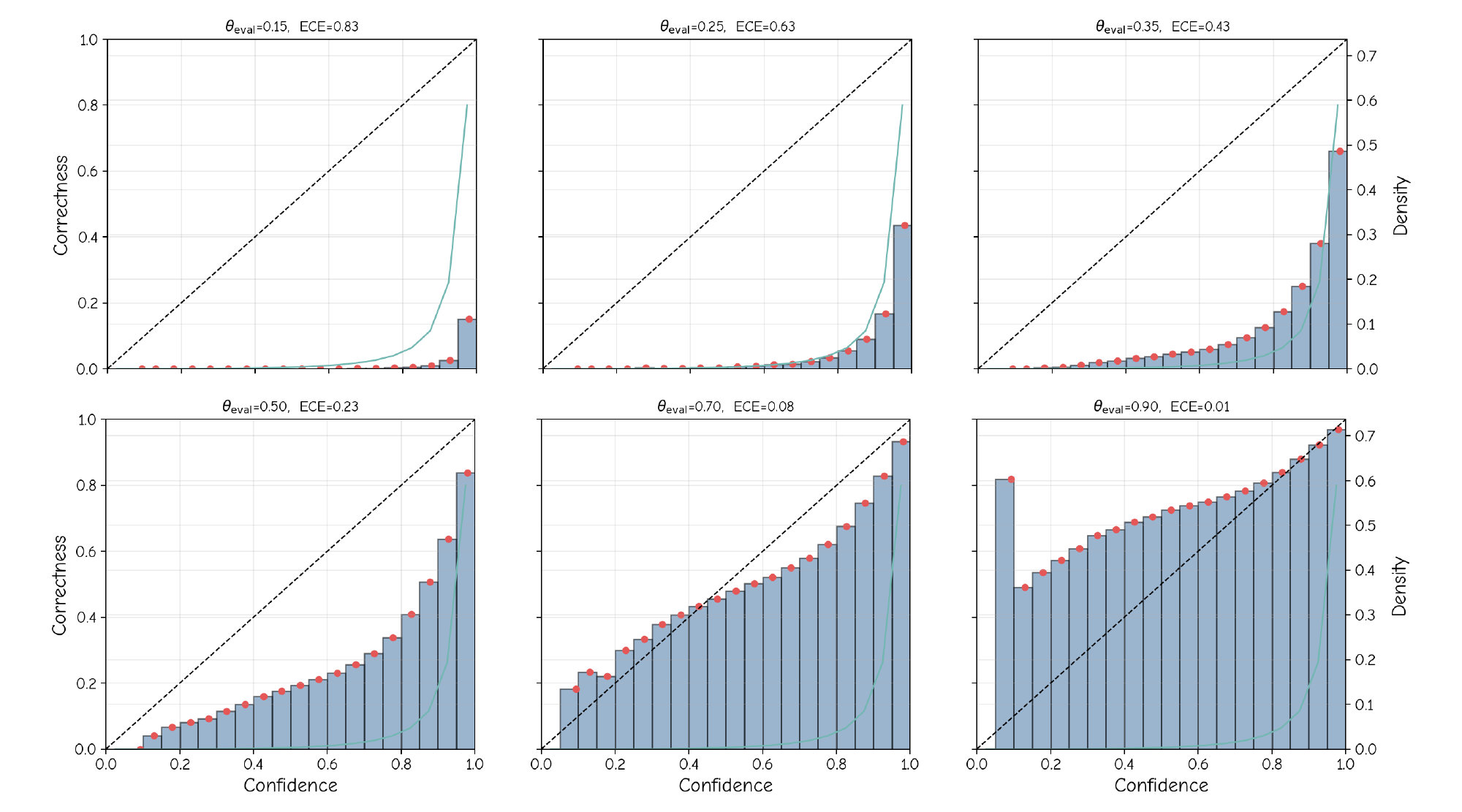}
        \vspace{-0.3cm}
        \centerline{\small (a) Latent-space reliability diagrams}
    \end{minipage}

    \vspace{0.4cm}

    \begin{minipage}[h]{0.88\textwidth}
        \centering
        \includegraphics[width=\linewidth]{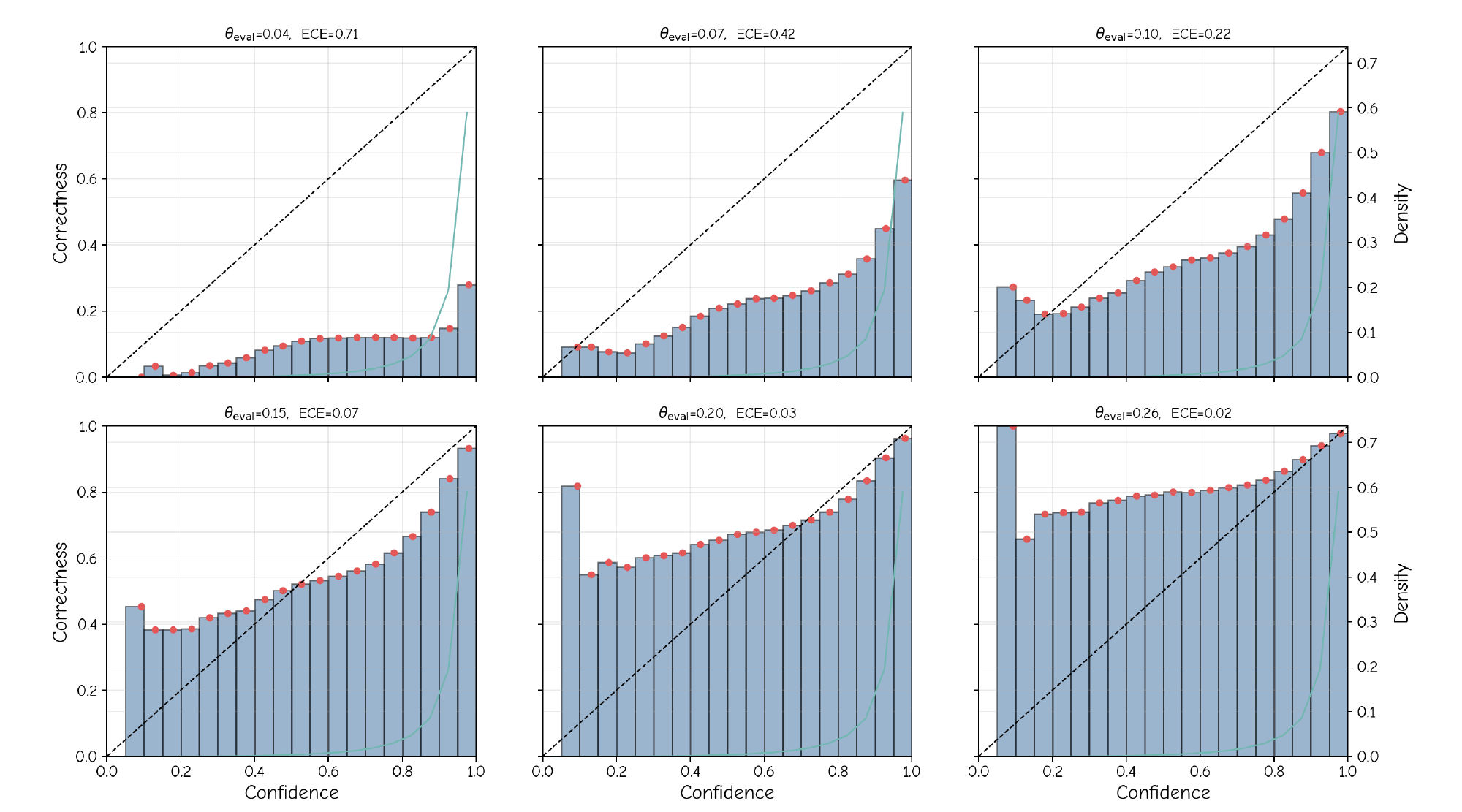}
        \vspace{-0.2cm}
        \centerline{\small (b) Pixel-space reliability diagrams}
    \end{minipage}
    \vspace{0cm}
    \caption{
    \textbf{Reliability diagrams across operational thresholds.}
    Each subfigure contains reliability diagrams under different choices of $\theta_{\mathrm{eval}}$:
    (a) latent space and (b) pixel space. As the threshold becomes looser, the observed correctness generally moves closer to the predicted confidence, although the latent-space probe remains more overconfident near the main operating point.
    }
    \label{fig:app_confidence_reliability}
    \vspace{-2cm}
\end{figure*}

\newpage
\paragraph{Operational calibration.}
Operational calibration evaluates fixed confidence maps against generated-rollout errors. We denote the operational threshold by $\theta_{\mathrm{eval}}$ and define correctness as $\mathbb{I}\{e^{\mathrm{roll}}_{t,(i,j)} \leq \theta_{\mathrm{eval}}\}$. Here, $e^{\mathrm{roll}}_{t,(i,j)}$ denotes channel-averaged absolute rollout error, spatially pooled to the probe grid, in latent or pixel space as indicated. Varying $\theta_{\mathrm{eval}}$ changes the post-hoc correctness labels while keeping the predicted confidence maps fixed.

Figure~\ref{fig:app_confidence_calibration_curve} summarizes calibration quality under different choices of $\theta_{\mathrm{eval}}$. In latent space, ECE and Brier improve as the threshold becomes looser, but the probe remains noticeably overconfident around the main operating point. Pixel-space calibration appears better under its own selected threshold, although the latent and pixel settings are not directly comparable because they rely on different error scales. Figure~\ref{fig:app_confidence_reliability} further visualizes the corresponding reliability diagrams. In both spaces, the curves become closer to the diagonal as $\theta_{\mathrm{eval}}$ increases, confirming that the apparent calibration quality depends strongly on the operational correctness definition. These 50-episode operational sweeps use rollout MAE; Appendix~\ref{app:c3_controlled} evaluates matched thresholds on 250 episodes using single-step MAE.

\begin{figure*}[h]
    \centering
    \begin{minipage}[h]{0.4\textwidth}
        \centering
        \includegraphics[width=\linewidth]
        {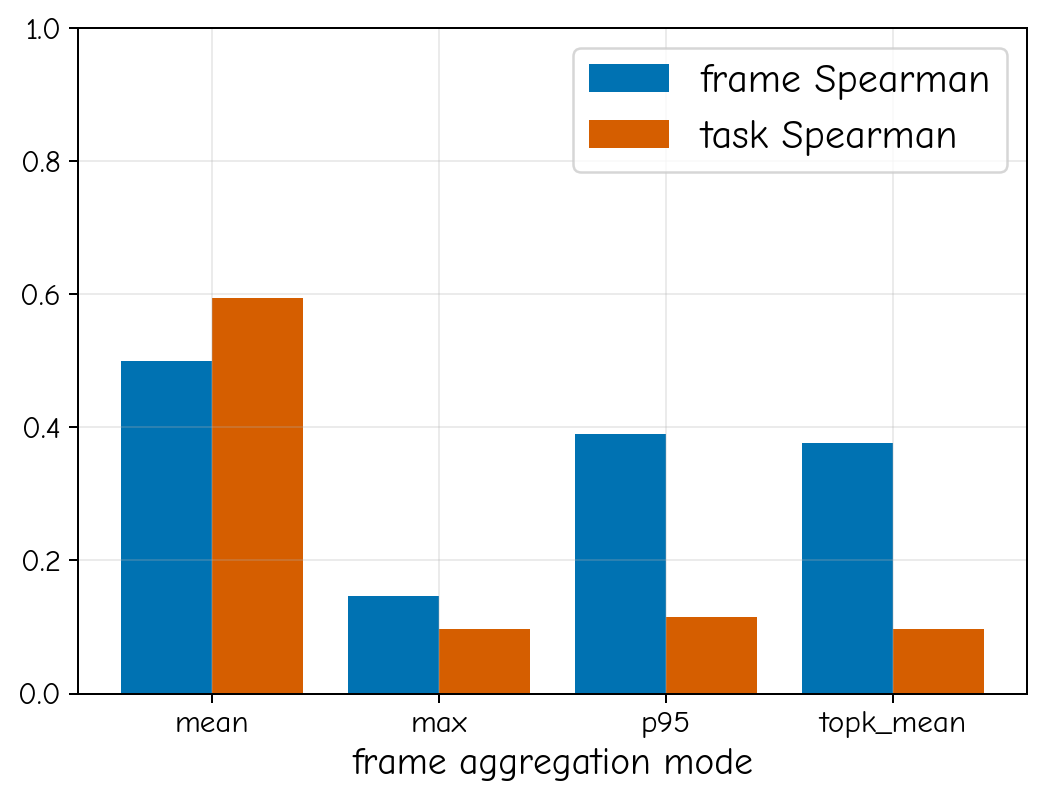}
        \vspace{-0.15cm}
        \centerline{\small (a) Frame aggregation in latent space}
    \end{minipage}
    \qquad
    \begin{minipage}[h]{0.4\textwidth}
        \centering
        \includegraphics[width=\linewidth]
        {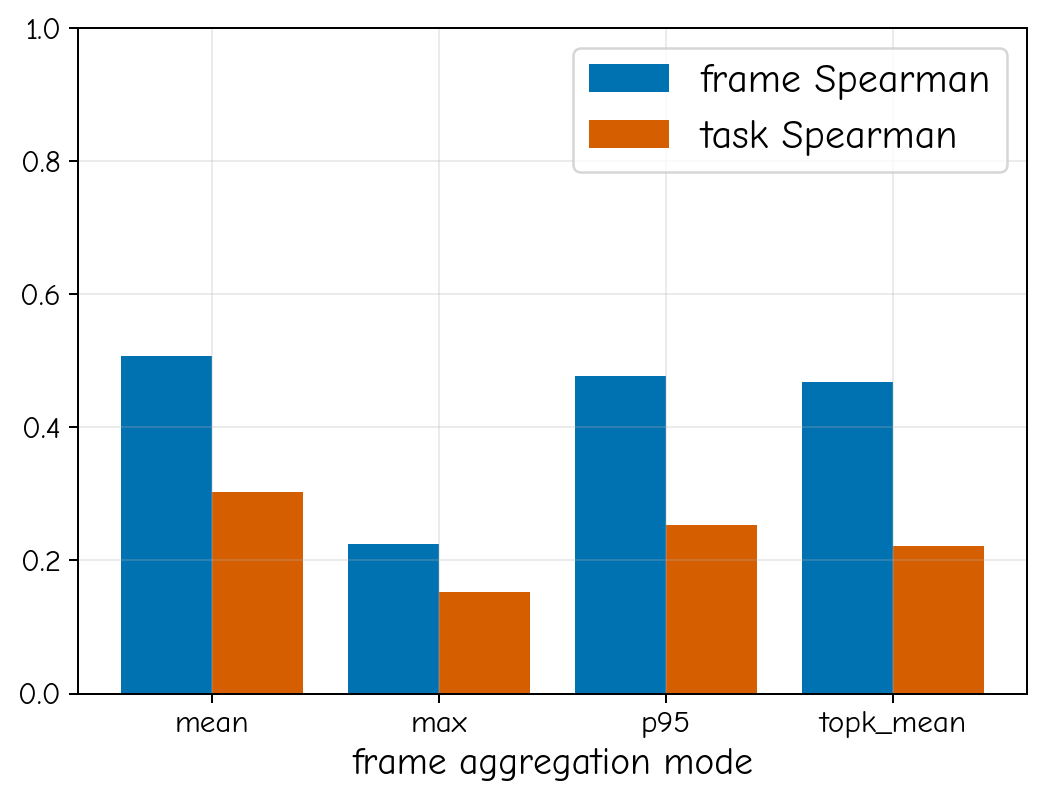}
        \vspace{-0.15cm}
        \centerline{\small (b) Frame aggregation in pixel space}
    \end{minipage}

    \vspace{0.3cm}

    \begin{minipage}[h]{0.62\textwidth}
        \centering
        \includegraphics[width=\linewidth]
        {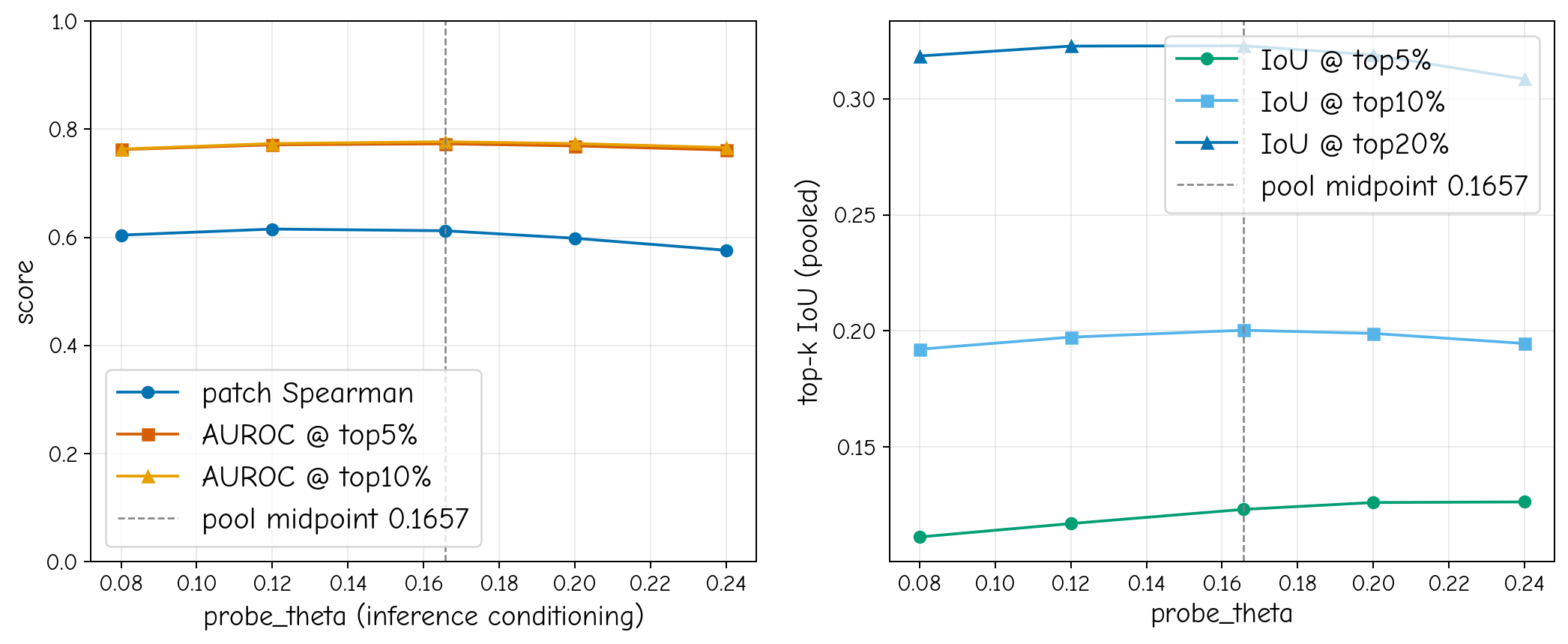}
        \vspace{-0.15cm}
        \centerline{\small (c) Localization sensitivity}
    \end{minipage}
    \hfill
    \begin{minipage}[h]{0.35\textwidth}
        \centering
        \includegraphics[width=\linewidth]
        {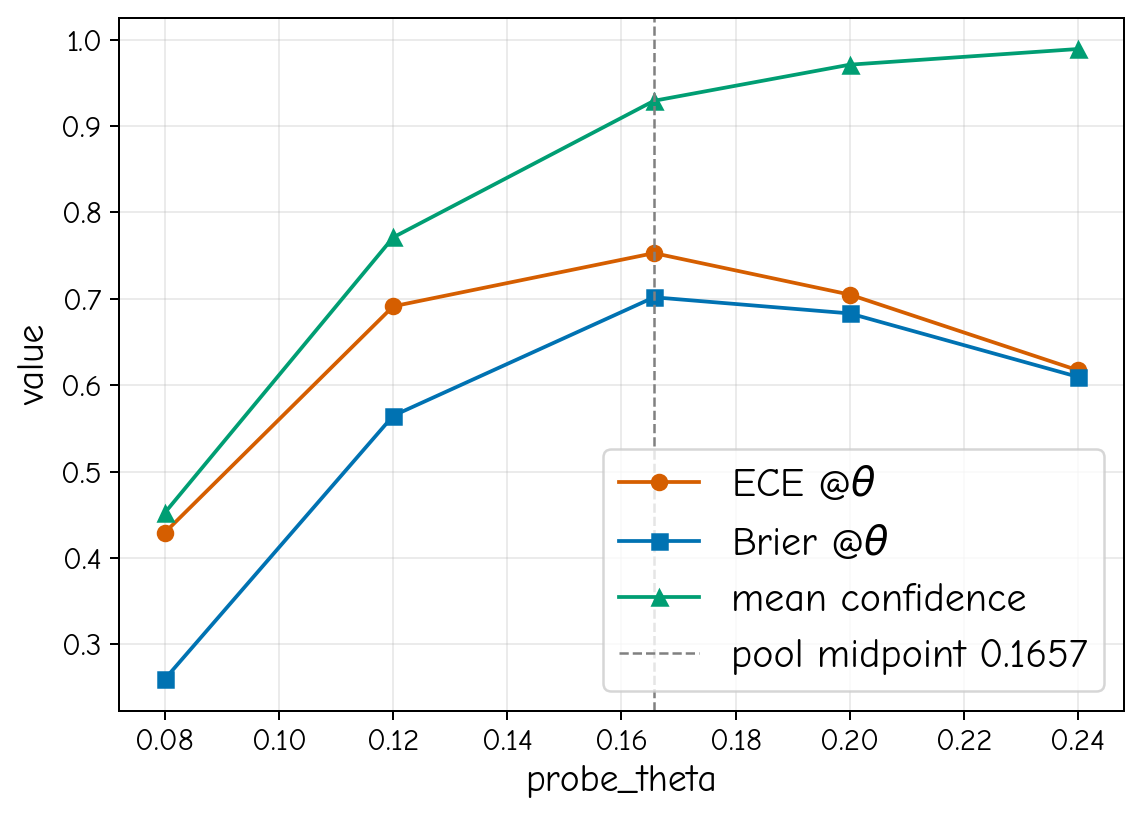}
        \vspace{-0.15cm}
        \centerline{\small (d) Calibration sensitivity}
    \end{minipage}

    \caption{
    \textbf{Sensitivity to confidence aggregation and probe conditioning.}
    (a--b) Frame- and task-level Spearman correlations under mean, maximum,
    percentile, and top-$k$ aggregation in latent and pixel spaces.
    (c) High-error detection and spatial localization under different probe
    conditioning thresholds $\theta_{\mathrm{cond}}$.
    (d) ECE, Brier score, and mean confidence under the same thresholds.
    Unlike the operational threshold $\theta_{\mathrm{eval}}$,
    changing $\theta_{\mathrm{cond}}$ regenerates the confidence maps.
    }
    \label{fig:app_confidence_sensitivity}
    \vspace{-0.2cm}
\end{figure*}

\paragraph{Parameter sensitivity.}
We examine two implementation choices that are not reported in the main text.
First, Figures~\ref{fig:app_confidence_sensitivity}(a--b) compare mean, maximum,
percentile, and top-$k$ frame aggregation. Mean aggregation provides the strongest
frame- and task-level Spearman correlations in both latent and pixel spaces,
whereas extreme-value aggregation amplifies isolated noisy patches. This motivates
the mean aggregation adopted in our main evaluation.

Second, Figures~\ref{fig:app_confidence_sensitivity}(c--d) vary the probe
conditioning threshold $\theta_{\mathrm{cond}}$ and regenerate the confidence maps.
Low-to-moderate thresholds provide the strongest spatial discrimination, while
large thresholds make most patches highly confident and reduce localization
contrast. The default threshold lies near the best-performing range, and the
overall variation remains moderate, indicating that the confidence-ranking
conclusion is not sensitive to a narrowly tuned threshold.

\newpage
\subsection{Additional Numerical Results for Active Learning}
\label{app:al_full_results}

This subsection provides the complete numerical results underlying the active-learning comparison in Section~\ref{subsec:why_active_learning}. We first report the four aggregated dimensions used in the main comparison, followed by paired-bootstrap improvements relative to EVAC-v1. We then provide the complete seed-wise component metrics for seeds 42, 3407, and 123.

\paragraph{Aggregated results and paired-bootstrap evidence.}
Table~\ref{tab:al_aggregated_three_seed} reports the aggregated active-learning results. Reconstruction, Scene, Motion, and Semantics follow the same definitions as in Section~\ref{subsec:why_active_learning}. For learned scoring methods, the reported values are averaged over seeds 42, 3407, and 123. Table~\ref{tab:al_bootstrap_three_seed} further reports paired-bootstrap improvements over EVAC-v1. For each learned scoring method, we concatenate the episode-level paired differences from all three seeds and perform 10,000 percentile-bootstrap resamples at 95\% confidence. The pairing with EVAC-v1 is preserved within each seed before pooling. Pooled intervals quantify episode-level variation; the seed-wise tables report the results of each individual run.

\begin{table*}[h]
\centering
\vspace{-0.3cm}
\caption{
Aggregated active-learning results.
Each entry reports the aggregated mean followed by its relative change from EVAC-v1.
Results for learned scoring methods are averaged over three seeds.
}
\label{tab:al_aggregated_three_seed}
\scriptsize
\setlength{\tabcolsep}{4.0pt}
\renewcommand{\arraystretch}{1.05}
\resizebox{0.9\textwidth}{!}{
\begin{tabular}{l|l||cccc}
\toprule
\rowcolor{gray!12}
Scoring & Weighting
& Reconstruction ($\uparrow$)
& Scene ($\uparrow$)
& Motion ($\uparrow$)
& Semantics ($\uparrow$) \\
\midrule
\multicolumn{2}{l||}{Base EVAC} & 0.5655\,\textcolor{red!70!black}{(-17.8\%)} & 0.8757\,\textcolor{red!70!black}{(-3.2\%)} & 0.0014\,\textcolor{red!70!black}{(-99.6\%)} & 0.4873\,\textcolor{red!70!black}{(-14.3\%)} \\
\multicolumn{2}{l||}{Base EVAC (Warmup v1)} & 0.6878\,\textcolor{gray}{(0.0\%)} & 0.9047\,\textcolor{gray}{(0.0\%)} & 0.3205\,\textcolor{gray}{(0.0\%)} & 0.5690\,\textcolor{gray}{(0.0\%)} \\
\midrule
\rowcolor{gray!6}
\multicolumn{6}{c}{\textbf{Selection-only retraining: no additional weighting after selection}} \\
\midrule
Random & None & 0.6409\,\textcolor{red!70!black}{(-6.8\%)} & 0.8524\,\textcolor{red!70!black}{(-5.8\%)} & 0.3310\,\textcolor{green!45!black}{(+3.3\%)} & 0.4872\,\textcolor{red!70!black}{(-14.4\%)} \\
RoboReward$^{1}$ & None & 0.6712\,\textcolor{red!70!black}{(-2.4\%)} & 0.8470\,\textcolor{red!70!black}{(-6.4\%)} & 0.3360\,\textcolor{green!45!black}{(+4.8\%)} & \cellcolor{green!15}\textbf{0.6022\,\textcolor{green!45!black}{(+5.8\%)}} \\
GVL$^{2}$ & None & \cellcolor{red!12}0.7088\,\textcolor{green!45!black}{(+3.1\%)} & \cellcolor{yellow!20}0.9057\,\textcolor{green!45!black}{(+0.1\%)} & \cellcolor{red!12}0.3393\,\textcolor{green!45!black}{(+5.9\%)} & \cellcolor{red!12}0.5913\,\textcolor{green!45!black}{(+3.9\%)} \\
Robometer-Prog$^{3}$ & None & 0.6317\,\textcolor{red!70!black}{(-8.2\%)} & 0.8532\,\textcolor{red!70!black}{(-5.7\%)} & 0.3307\,\textcolor{green!45!black}{(+3.2\%)} & 0.5759\,\textcolor{green!45!black}{(+1.2\%)} \\
Robometer-Pref$^{3}$ & None & 0.6474\,\textcolor{red!70!black}{(-5.9\%)} & 0.8443\,\textcolor{red!70!black}{(-6.7\%)} & 0.3208\,\textcolor{green!45!black}{(+0.1\%)} & 0.5789\,\textcolor{green!45!black}{(+1.7\%)} \\
PRM-as-Judge$^{4}$ & None & 0.7024\,\textcolor{green!45!black}{(+2.1\%)} & \cellcolor{red!12}0.9033\,\textcolor{red!70!black}{(-0.2\%)} & \cellcolor{yellow!20}0.3505\,\textcolor{green!45!black}{(+9.4\%)} & \cellcolor{yellow!20}0.5922\,\textcolor{green!45!black}{(+4.1\%)} \\
LRMs$^{5}$ & None & \cellcolor{yellow!20}0.7110\,\textcolor{green!45!black}{(+3.4\%)} & 0.8966\,\textcolor{red!70!black}{(-0.9\%)} & 0.2885\,\textcolor{red!70!black}{(-10.0\%)} & 0.5776\,\textcolor{green!45!black}{(+1.5\%)} \\
\textbf{Confidence (Ours)} & \textbf{None} & \cellcolor{green!15}\textbf{0.7219\,\textcolor{green!45!black}{(+5.0\%)}} & \cellcolor{green!15}\textbf{0.9196\,\textcolor{green!45!black}{(+1.6\%)}} & \cellcolor{green!15}\textbf{0.3643\,\textcolor{green!45!black}{(+13.7\%)}} & 0.5897\,\textcolor{green!45!black}{(+3.6\%)} \\
\midrule
\rowcolor{gray!6}
\multicolumn{6}{c}{\textbf{Selection + additional weighting: extra scoring after selection}} \\
\midrule
Robometer-Prog$^{3}$ & Frame & 0.6944\,\textcolor{green!45!black}{(+1.0\%)} & 0.8882\,\textcolor{red!70!black}{(-1.8\%)} & \cellcolor{red!12}0.3357\,\textcolor{green!45!black}{(+4.7\%)} & 0.5884\,\textcolor{green!45!black}{(+3.4\%)} \\
Robometer-Pref$^{3}$ & Frame & 0.6803\,\textcolor{red!70!black}{(-1.1\%)} & \cellcolor{red!12}0.8945\,\textcolor{red!70!black}{(-1.1\%)} & 0.2969\,\textcolor{red!70!black}{(-7.4\%)} & 0.5601\,\textcolor{red!70!black}{(-1.6\%)} \\
PRM-as-Judge$^{4}$ & Frame & \cellcolor{yellow!20}0.7123\,\textcolor{green!45!black}{(+3.6\%)} & \cellcolor{yellow!20}0.9053\,\textcolor{green!45!black}{(+0.1\%)} & 0.3347\,\textcolor{green!45!black}{(+4.4\%)} & \cellcolor{yellow!20}0.5963\,\textcolor{green!45!black}{(+4.8\%)} \\
LRMs$^{5}$ & Frame & 0.6955\,\textcolor{green!45!black}{(+1.1\%)} & 0.8882\,\textcolor{red!70!black}{(-1.8\%)} & 0.3171\,\textcolor{red!70!black}{(-1.1\%)} & 0.5815\,\textcolor{green!45!black}{(+2.2\%)} \\
\textbf{Confidence (Ours)} & \textbf{Frame} & \cellcolor{red!12}0.7047\,\textcolor{green!45!black}{(+2.5\%)} & \cellcolor{green!15}\textbf{0.9143\,\textcolor{green!45!black}{(+1.1\%)}} & \cellcolor{yellow!20}0.3398\,\textcolor{green!45!black}{(+6.0\%)} & \cellcolor{red!12}0.5944\,\textcolor{green!45!black}{(+4.5\%)} \\
\textbf{Confidence (Ours)} & \textbf{Fr.+Patch} & \cellcolor{green!15}\textbf{0.7265\,\textcolor{green!45!black}{(+5.6\%)}} & 0.8942\,\textcolor{red!70!black}{(-1.2\%)} & \cellcolor{green!15}\textbf{0.3510\,\textcolor{green!45!black}{(+9.5\%)}} & \cellcolor{green!15}\textbf{0.6034\,\textcolor{green!45!black}{(+6.1\%)}} \\
\bottomrule
\end{tabular}}
\end{table*}

\begin{table*}[h]
\vspace{-0.3cm}
\centering
\caption{
Paired-bootstrap improvements over EVAC-v1.
For each learned scoring method, episode-level paired differences from seeds 42, 3407, and 123 are pooled before bootstrap resampling.
Each entry gives the 95\% confidence interval followed by the paired mean in parentheses.
}
\label{tab:al_bootstrap_three_seed}
\scriptsize
\setlength{\tabcolsep}{3.4pt}
\renewcommand{\arraystretch}{1.05}
\resizebox{\textwidth}{!}{
\begin{tabular}{l|l||cccc}
\toprule
\rowcolor{gray!12}
Scoring & Weighting
& $\Delta$ Reconstruction ($\uparrow$)
& $\Delta$ Scene ($\uparrow$)
& $\Delta$ Motion ($\uparrow$)
& $\Delta$ Semantics ($\uparrow$) \\
\midrule
\multicolumn{2}{l||}{Base EVAC} & [-0.133,\,-0.112]\,\textcolor{red!70!black}{(-0.122)} & [-0.038,\,-0.020]\,\textcolor{red!70!black}{(-0.029)} & -- & [-0.113,\,-0.049]\,\textcolor{red!70!black}{(-0.081)} \\
\multicolumn{2}{l||}{Base EVAC (Warmup v1)} & -- & -- & -- & -- \\
\midrule
\rowcolor{gray!6}
\multicolumn{6}{c}{\textbf{Selection-only retraining: no additional weighting after selection}} \\
\midrule
Random & None & [-0.055,\,-0.039]\,\textcolor{red!70!black}{(-0.047)} & [-0.060,\,-0.045]\,\textcolor{red!70!black}{(-0.052)} & [-0.035,\,+0.056]\,\textcolor{green!45!black}{(+0.011)} & [-0.112,\,-0.046]\,\textcolor{red!70!black}{(-0.079)} \\
RoboReward$^{1}$ & None & [-0.020,\,-0.013]\,\textcolor{red!70!black}{(-0.017)} & [-0.063,\,-0.052]\,\textcolor{red!70!black}{(-0.058)} & [-0.006,\,+0.038]\,\textcolor{green!45!black}{(+0.015)} & \cellcolor{green!15}\textbf{[+0.016,\,+0.053]\,\textcolor{green!45!black}{(+0.035)}} \\
GVL$^{2}$ & None & \cellcolor{red!12}[+0.018,\,+0.024]\,\textcolor{green!45!black}{(+0.021)} & \cellcolor{yellow!20}[-0.001,\,+0.003]\,\textcolor{green!45!black}{(+0.001)} & \cellcolor{red!12}[-0.008,\,+0.046]\,\textcolor{green!45!black}{(+0.019)} & \cellcolor{yellow!20}[+0.003,\,+0.042]\,\textcolor{green!45!black}{(+0.023)} \\
Robometer-Prog$^{3}$ & None & [-0.061,\,-0.051]\,\textcolor{red!70!black}{(-0.056)} & [-0.056,\,-0.047]\,\textcolor{red!70!black}{(-0.051)} & [-0.017,\,+0.038]\,\textcolor{green!45!black}{(+0.010)} & [-0.014,\,+0.024]\,\textcolor{green!45!black}{(+0.006)} \\
Robometer-Pref$^{3}$ & None & [-0.046,\,-0.035]\,\textcolor{red!70!black}{(-0.040)} & [-0.067,\,-0.054]\,\textcolor{red!70!black}{(-0.060)} & [-0.024,\,+0.025]\,\textcolor{green!45!black}{(+0.000)} & [-0.012,\,+0.028]\,\textcolor{green!45!black}{(+0.008)} \\
PRM-as-Judge$^{4}$ & None & [+0.011,\,+0.018]\,\textcolor{green!45!black}{(+0.015)} & \cellcolor{red!12}[-0.004,\,+0.001]\,\textcolor{red!70!black}{(-0.001)} & \cellcolor{yellow!20}[+0.005,\,+0.056]\,\textcolor{green!45!black}{(+0.030)} & \cellcolor{red!12}[+0.001,\,+0.040]\,\textcolor{green!45!black}{(+0.021)} \\
LRMs$^{5}$ & None & \cellcolor{yellow!20}[+0.020,\,+0.027]\,\textcolor{green!45!black}{(+0.023)} & [-0.011,\,-0.006]\,\textcolor{red!70!black}{(-0.008)} & [-0.057,\,-0.008]\,\textcolor{red!70!black}{(-0.032)} & [-0.010,\,+0.029]\,\textcolor{green!45!black}{(+0.010)} \\
\textbf{Confidence (Ours)} & \textbf{None} & \cellcolor{green!15}\textbf{[+0.031,\,+0.037]\,\textcolor{green!45!black}{(+0.034)}} & \cellcolor{green!15}\textbf{[+0.012,\,+0.018]\,\textcolor{green!45!black}{(+0.015)}} & \cellcolor{green!15}\textbf{[+0.018,\,+0.070]\,\textcolor{green!45!black}{(+0.044)}} & [-0.002,\,+0.038]\,\textcolor{green!45!black}{(+0.018)} \\
\midrule
\rowcolor{gray!6}
\multicolumn{6}{c}{\textbf{Selection + additional weighting: extra scoring after selection}} \\
\midrule
Robometer-Prog$^{3}$ & Frame & [+0.003,\,+0.011]\,\textcolor{green!45!black}{(+0.007)} & [-0.021,\,-0.012]\,\textcolor{red!70!black}{(-0.017)} & \cellcolor{red!12}[-0.010,\,+0.041]\,\textcolor{green!45!black}{(+0.015)} & [-0.000,\,+0.038]\,\textcolor{green!45!black}{(+0.019)} \\
Robometer-Pref$^{3}$ & Frame & [-0.011,\,-0.004]\,\textcolor{red!70!black}{(-0.007)} & \cellcolor{red!12}[-0.013,\,-0.007]\,\textcolor{red!70!black}{(-0.010)} & [-0.049,\,+0.002]\,\textcolor{red!70!black}{(-0.024)} & [-0.031,\,+0.008]\,\textcolor{red!70!black}{(-0.011)} \\
PRM-as-Judge$^{4}$ & Frame & \cellcolor{yellow!20}[+0.021,\,+0.028]\,\textcolor{green!45!black}{(+0.024)} & \cellcolor{yellow!20}[-0.003,\,+0.004]\,\textcolor{green!45!black}{(+0.001)} & [-0.010,\,+0.039]\,\textcolor{green!45!black}{(+0.014)} & \cellcolor{yellow!20}[+0.008,\,+0.046]\,\textcolor{green!45!black}{(+0.027)} \\
LRMs$^{5}$ & Frame & [+0.004,\,+0.011]\,\textcolor{green!45!black}{(+0.008)} & [-0.020,\,-0.013]\,\textcolor{red!70!black}{(-0.016)} & [-0.027,\,+0.020]\,\textcolor{red!70!black}{(-0.003)} & [-0.005,\,+0.033]\,\textcolor{green!45!black}{(+0.014)} \\
\textbf{Confidence (Ours)} & \textbf{Frame} & \cellcolor{red!12}[+0.014,\,+0.020]\,\textcolor{green!45!black}{(+0.017)} & \cellcolor{green!15}\textbf{[+0.007,\,+0.012]\,\textcolor{green!45!black}{(+0.010)}} & \cellcolor{yellow!20}[-0.007,\,+0.045]\,\textcolor{green!45!black}{(+0.019)} & \cellcolor{red!12}[+0.007,\,+0.044]\,\textcolor{green!45!black}{(+0.026)} \\
\textbf{Confidence (Ours)} & \textbf{Fr.+Patch} & \cellcolor{green!15}\textbf{[+0.036,\,+0.042]\,\textcolor{green!45!black}{(+0.039)}} & [-0.013,\,-0.008]\,\textcolor{red!70!black}{(-0.010)} & \cellcolor{green!15}\textbf{[+0.008,\,+0.054]\,\textcolor{green!45!black}{(+0.031)}} & \cellcolor{green!15}\textbf{[+0.011,\,+0.049]\,\textcolor{green!45!black}{(+0.030)}} \\
\bottomrule
\end{tabular}}
\vspace{-0.3cm}
\end{table*}

\paragraph{Seed-wise detailed results.}
Tables~\ref{tab:al_seed42}--\ref{tab:al_seed123} report the complete normalized component metrics for seeds 42, 3407, and 123, respectively. Seed 42 is the default seed. Base EVAC, EVAC-v1, and Random are shared reference results and are repeated in each table for ease of comparison.

\begin{table*}[h]
\centering
\vspace{-0.3cm}
\caption{Detailed normalized active-learning results for seed 42.}
\label{tab:al_seed42}
\scriptsize
\setlength{\tabcolsep}{2.8pt}
\renewcommand{\arraystretch}{1.04}
\resizebox{\textwidth}{!}{
\begin{tabular}{l|l||cc|c|ccc|ccc}
\toprule
\rowcolor{gray!12}
\multirow{2}{*}{Scoring} & \multirow{2}{*}{Weighting}
& \multicolumn{2}{c}{Reconstruction ($\uparrow$)}
& \multicolumn{1}{c}{Scene ($\uparrow$)}
& \multicolumn{3}{c}{Semantics ($\uparrow$)}
& \multicolumn{3}{c}{Motion ($\uparrow$)} \\
\cmidrule{3-11}
\rowcolor{gray!12}
& & PSNR ($\uparrow$) & SSIM ($\uparrow$) & Scene Cons. ($\uparrow$)
& Logics ($\uparrow$) & Sem.-CLIP ($\uparrow$) & Sem.-BLEU ($\uparrow$)
& Traj-HSD ($\uparrow$) & Traj-Dyn ($\uparrow$) & Traj-nDTW ($\uparrow$) \\
\midrule
\multicolumn{2}{l||}{Base EVAC} & 0.5532 & 0.5778 & 0.8757 & 0.4298 & 0.8523 & 0.1799 & 0.0007 & 0.0001 & 0.0006 \\
\multicolumn{2}{l||}{Base EVAC (Warmup v1)} & 0.6446 & 0.7309 & 0.9047 & 0.5537 & 0.8824 & 0.2708 & 0.1045 & 0.0721 & 0.1439 \\
\midrule
\rowcolor{gray!6}
\multicolumn{11}{c}{\textbf{Selection-only retraining: no additional weighting after selection}} \\
\midrule
Random & None & 0.5968 & 0.6849 & 0.8524 & 0.3554 & 0.8689 & 0.2372 & 0.1067 & \cellcolor{red!12}0.0725 & \cellcolor{red!12}0.1518 \\
RoboReward$^{1}$ & None & 0.6276 & 0.7067 & 0.8712 & \cellcolor{yellow!20}0.6322 & \cellcolor{yellow!20}0.8895 & \cellcolor{yellow!20}0.2832 & 0.1076 & 0.0662 & 0.1501 \\
GVL$^{2}$ & None & \cellcolor{red!12}0.6550 & 0.7449 & \cellcolor{yellow!20}0.9136 & 0.5826 & 0.8885 & 0.2733 & 0.1033 & 0.0607 & 0.1381 \\
Robometer-Prog$^{3}$ & None & 0.5547 & 0.6711 & 0.8208 & 0.5496 & 0.8800 & 0.2680 & \cellcolor{yellow!20}0.1166 & \cellcolor{yellow!20}0.0774 & \cellcolor{green!15}\textbf{0.1805} \\
Robometer-Pref$^{3}$ & None & 0.6047 & 0.7005 & 0.9005 & 0.6157 & 0.8866 & 0.2729 & 0.0912 & 0.0492 & 0.1242 \\
PRM-as-Judge$^{4}$ & None & 0.6523 & \cellcolor{yellow!20}0.7641 & \cellcolor{red!12}0.9102 & \cellcolor{green!15}\textbf{0.6405} & \cellcolor{green!15}\textbf{0.8926} & \cellcolor{green!15}\textbf{0.3023} & \cellcolor{red!12}0.1095 & 0.0696 & 0.1495 \\
LRMs$^{5}$ & None & \cellcolor{yellow!20}0.6577 & \cellcolor{red!12}0.7528 & 0.8869 & 0.5785 & 0.8803 & 0.2550 & 0.0960 & 0.0595 & 0.1422 \\
\textbf{Confidence (Ours)} & \textbf{None} & \cellcolor{green!15}\textbf{0.6838} & \cellcolor{green!15}\textbf{0.7797} & \cellcolor{green!15}\textbf{0.9322} & \cellcolor{red!12}0.6198 & \cellcolor{red!12}0.8894 & \cellcolor{red!12}0.2754 & \cellcolor{green!15}\textbf{0.1313} & \cellcolor{green!15}\textbf{0.0921} & \cellcolor{yellow!20}0.1780 \\
\midrule
\rowcolor{gray!6}
\multicolumn{11}{c}{\textbf{Selection + additional weighting: extra scoring after selection}} \\
\midrule
Robometer-Prog$^{3}$ & Frame & 0.5932 & 0.6981 & 0.8399 & 0.6074 & \cellcolor{red!12}0.8883 & \cellcolor{green!15}\textbf{0.2976} & \cellcolor{yellow!20}0.1151 & \cellcolor{red!12}0.0703 & \cellcolor{red!12}0.1655 \\
Robometer-Pref$^{3}$ & Frame & 0.6412 & 0.7324 & 0.8948 & 0.6033 & 0.8868 & 0.2805 & 0.0981 & 0.0678 & 0.1291 \\
PRM-as-Judge$^{4}$ & Frame & 0.6695 & \cellcolor{yellow!20}0.7733 & \cellcolor{yellow!20}0.9193 & \cellcolor{yellow!20}0.6281 & 0.8880 & \cellcolor{yellow!20}0.2937 & 0.1000 & 0.0592 & 0.1443 \\
LRMs$^{5}$ & Frame & \cellcolor{red!12}0.6716 & 0.7616 & 0.8958 & 0.5785 & \cellcolor{yellow!20}0.8896 & 0.2733 & 0.0985 & 0.0623 & 0.1497 \\
\textbf{Confidence (Ours)} & \textbf{Frame} & \cellcolor{yellow!20}0.6737 & \cellcolor{red!12}0.7650 & \cellcolor{green!15}\textbf{0.9295} & \cellcolor{green!15}\textbf{0.6405} & 0.8867 & 0.2843 & \cellcolor{green!15}\textbf{0.1179} & \cellcolor{yellow!20}0.0728 & \cellcolor{green!15}\textbf{0.1706} \\
\textbf{Confidence (Ours)} & \textbf{Fr.+Patch} & \cellcolor{green!15}\textbf{0.6795} & \cellcolor{green!15}\textbf{0.7746} & \cellcolor{red!12}0.9014 & \cellcolor{red!12}0.6240 & \cellcolor{green!15}\textbf{0.8945} & \cellcolor{red!12}0.2859 & \cellcolor{red!12}0.1066 & \cellcolor{green!15}\textbf{0.0761} & \cellcolor{yellow!20}0.1665 \\
\bottomrule
\end{tabular}}
\end{table*}

\begin{table*}[h]
\centering
\vspace{-0.3cm}
\caption{Detailed normalized active-learning results for seed 3407.}
\label{tab:al_seed3407}
\scriptsize
\setlength{\tabcolsep}{2.8pt}
\renewcommand{\arraystretch}{1.04}
\resizebox{\textwidth}{!}{
\begin{tabular}{l|l||cc|c|ccc|ccc}
\toprule
\rowcolor{gray!12}
\multirow{2}{*}{Scoring} & \multirow{2}{*}{Weighting}
& \multicolumn{2}{c}{Reconstruction ($\uparrow$)}
& \multicolumn{1}{c}{Scene ($\uparrow$)}
& \multicolumn{3}{c}{Semantics ($\uparrow$)}
& \multicolumn{3}{c}{Motion ($\uparrow$)} \\
\cmidrule{3-11}
\rowcolor{gray!12}
& & PSNR ($\uparrow$) & SSIM ($\uparrow$) & Scene Cons. ($\uparrow$)
& Logics ($\uparrow$) & Sem.-CLIP ($\uparrow$) & Sem.-BLEU ($\uparrow$)
& Traj-HSD ($\uparrow$) & Traj-Dyn ($\uparrow$) & Traj-nDTW ($\uparrow$) \\
\midrule
\multicolumn{2}{l||}{Base EVAC} & 0.5532 & 0.5778 & 0.8757 & 0.4298 & 0.8523 & 0.1799 & 0.0007 & 0.0001 & 0.0006 \\
\multicolumn{2}{l||}{Base EVAC (Warmup v1)} & 0.6446 & 0.7309 & 0.9047 & 0.5537 & 0.8824 & 0.2708 & 0.1045 & 0.0721 & 0.1439 \\
\midrule
\rowcolor{gray!6}
\multicolumn{11}{c}{\textbf{Selection-only retraining: no additional weighting after selection}} \\
\midrule
Random & None & 0.5968 & 0.6849 & 0.8524 & 0.3554 & 0.8689 & 0.2372 & 0.1067 & 0.0725 & 0.1518 \\
RoboReward$^{1}$ & None & 0.6262 & 0.7290 & 0.7985 & \cellcolor{green!15}\textbf{0.6488} & 0.8863 & 0.2775 & 0.1089 & \cellcolor{red!12}0.0732 & 0.1554 \\
GVL$^{2}$ & None & \cellcolor{red!12}0.6546 & 0.7503 & 0.8959 & \cellcolor{yellow!20}0.6157 & \cellcolor{red!12}0.8892 & 0.2868 & \cellcolor{green!15}\textbf{0.1200} & \cellcolor{yellow!20}0.0759 & \cellcolor{yellow!20}0.1681 \\
Robometer-Prog$^{3}$ & None & 0.6492 & 0.7490 & 0.8857 & \cellcolor{red!12}0.6033 & 0.8814 & 0.2580 & \cellcolor{red!12}0.1148 & 0.0715 & 0.1491 \\
Robometer-Pref$^{3}$ & None & 0.6442 & \cellcolor{red!12}0.7535 & 0.8779 & \cellcolor{red!12}0.6033 & \cellcolor{yellow!20}0.8895 & \cellcolor{yellow!20}0.2877 & 0.1066 & 0.0671 & \cellcolor{red!12}0.1565 \\
PRM-as-Judge$^{4}$ & None & \cellcolor{green!15}\textbf{0.6714} & \cellcolor{green!15}\textbf{0.7728} & \cellcolor{yellow!20}0.9020 & 0.5950 & \cellcolor{green!15}\textbf{0.8896} & \cellcolor{red!12}0.2873 & \cellcolor{yellow!20}0.1197 & \cellcolor{green!15}\textbf{0.0784} & \cellcolor{green!15}\textbf{0.1757} \\
LRMs$^{5}$ & None & \cellcolor{yellow!20}0.6619 & \cellcolor{yellow!20}0.7658 & \cellcolor{red!12}0.8991 & 0.5331 & 0.8817 & 0.2672 & 0.0925 & 0.0564 & 0.1299 \\
\textbf{Confidence (Ours)} & \textbf{None} & 0.6521 & 0.7443 & \cellcolor{green!15}\textbf{0.9029} & 0.5537 & 0.8891 & \cellcolor{green!15}\textbf{0.2953} & 0.0998 & 0.0618 & 0.1469 \\
\midrule
\rowcolor{gray!6}
\multicolumn{11}{c}{\textbf{Selection + additional weighting: extra scoring after selection}} \\
\midrule
Robometer-Prog$^{3}$ & Frame & \cellcolor{yellow!20}0.6785 & \cellcolor{yellow!20}0.7739 & \cellcolor{green!15}\textbf{0.9148} & 0.5661 & \cellcolor{yellow!20}0.8896 & \cellcolor{yellow!20}0.2925 & 0.1042 & \cellcolor{red!12}0.0683 & 0.1454 \\
Robometer-Pref$^{3}$ & Frame & 0.6329 & 0.7343 & \cellcolor{red!12}0.9000 & 0.5372 & 0.8875 & 0.2796 & 0.0867 & 0.0542 & 0.1241 \\
PRM-as-Judge$^{4}$ & Frame & 0.6544 & \cellcolor{red!12}0.7606 & 0.8816 & \cellcolor{yellow!20}0.6364 & 0.8861 & 0.2681 & \cellcolor{green!15}\textbf{0.1276} & \cellcolor{green!15}\textbf{0.0949} & \cellcolor{green!15}\textbf{0.1827} \\
LRMs$^{5}$ & Frame & 0.6391 & 0.7323 & 0.8984 & 0.5868 & 0.8872 & \cellcolor{red!12}0.2884 & 0.1047 & 0.0582 & 0.1508 \\
\textbf{Confidence (Ours)} & \textbf{Frame} & \cellcolor{red!12}0.6644 & 0.7518 & \cellcolor{yellow!20}0.9127 & \cellcolor{red!12}0.6074 & \cellcolor{red!12}0.8887 & 0.2734 & \cellcolor{red!12}0.1121 & \cellcolor{yellow!20}0.0704 & \cellcolor{red!12}0.1663 \\
\textbf{Confidence (Ours)} & \textbf{Fr.+Patch} & \cellcolor{green!15}\textbf{0.6798} & \cellcolor{green!15}\textbf{0.7822} & 0.8948 & \cellcolor{green!15}\textbf{0.6612} & \cellcolor{green!15}\textbf{0.8926} & \cellcolor{green!15}\textbf{0.2979} & \cellcolor{yellow!20}0.1161 & 0.0679 & \cellcolor{yellow!20}0.1665 \\
\bottomrule
\end{tabular}}
\end{table*}

\begin{table*}[h]
\centering
\vspace{-0.3cm}
\caption{Detailed normalized active-learning results for seed 123.}
\label{tab:al_seed123}
\scriptsize
\setlength{\tabcolsep}{2.8pt}
\renewcommand{\arraystretch}{1.04}
\resizebox{\textwidth}{!}{
\begin{tabular}{l|l||cc|c|ccc|ccc}
\toprule
\rowcolor{gray!12}
\multirow{2}{*}{Scoring} & \multirow{2}{*}{Weighting}
& \multicolumn{2}{c}{Reconstruction ($\uparrow$)}
& \multicolumn{1}{c}{Scene ($\uparrow$)}
& \multicolumn{3}{c}{Semantics ($\uparrow$)}
& \multicolumn{3}{c}{Motion ($\uparrow$)} \\
\cmidrule{3-11}
\rowcolor{gray!12}
& & PSNR ($\uparrow$) & SSIM ($\uparrow$) & Scene Cons. ($\uparrow$)
& Logics ($\uparrow$) & Sem.-CLIP ($\uparrow$) & Sem.-BLEU ($\uparrow$)
& Traj-HSD ($\uparrow$) & Traj-Dyn ($\uparrow$) & Traj-nDTW ($\uparrow$) \\
\midrule
\multicolumn{2}{l||}{Base EVAC} & 0.5532 & 0.5778 & 0.8757 & 0.4298 & 0.8523 & 0.1799 & 0.0007 & 0.0001 & 0.0006 \\
\multicolumn{2}{l||}{Base EVAC (Warmup v1)} & 0.6446 & 0.7309 & 0.9047 & 0.5537 & 0.8824 & 0.2708 & 0.1045 & 0.0721 & 0.1439 \\
\midrule
\rowcolor{gray!6}
\multicolumn{11}{c}{\textbf{Selection-only retraining: no additional weighting after selection}} \\
\midrule
Random & None & 0.5968 & 0.6849 & 0.8524 & 0.3554 & 0.8689 & 0.2372 & 0.1067 & 0.0725 & 0.1518 \\
RoboReward$^{1}$ & None & 0.6260 & 0.7117 & 0.8712 & \cellcolor{green!15}\textbf{0.6364} & 0.8825 & 0.2838 & \cellcolor{yellow!20}0.1174 & 0.0685 & 0.1607 \\
GVL$^{2}$ & None & \cellcolor{yellow!20}0.6795 & \cellcolor{yellow!20}0.7686 & \cellcolor{yellow!20}0.9075 & 0.5992 & \cellcolor{yellow!20}0.8918 & \cellcolor{yellow!20}0.2946 & 0.1109 & 0.0698 & \cellcolor{yellow!20}0.1713 \\
Robometer-Prog$^{3}$ & None & 0.5388 & 0.6272 & 0.8532 & 0.5992 & 0.8815 & 0.2620 & 0.0955 & 0.0582 & 0.1282 \\
Robometer-Pref$^{3}$ & None & 0.5267 & 0.6546 & 0.7544 & 0.5248 & 0.8816 & 0.2479 & 0.1160 & \cellcolor{red!12}0.0752 & \cellcolor{green!15}\textbf{0.1764} \\
PRM-as-Judge$^{4}$ & None & 0.6164 & 0.7372 & 0.8977 & 0.5744 & 0.8817 & 0.2659 & \cellcolor{red!12}0.1168 & \cellcolor{yellow!20}0.0758 & 0.1565 \\
LRMs$^{5}$ & None & \cellcolor{red!12}0.6655 & \cellcolor{red!12}0.7624 & \cellcolor{red!12}0.9037 & \cellcolor{yellow!20}0.6116 & \cellcolor{red!12}0.8862 & \cellcolor{green!15}\textbf{0.3046} & 0.0952 & 0.0603 & 0.1335 \\
\textbf{Confidence (Ours)} & \textbf{None} & \cellcolor{green!15}\textbf{0.6878} & \cellcolor{green!15}\textbf{0.7836} & \cellcolor{green!15}\textbf{0.9237} & \cellcolor{red!12}0.6033 & \cellcolor{green!15}\textbf{0.8924} & \cellcolor{red!12}0.2889 & \cellcolor{green!15}\textbf{0.1232} & \cellcolor{green!15}\textbf{0.0896} & \cellcolor{red!12}0.1702 \\
\midrule
\rowcolor{gray!6}
\multicolumn{11}{c}{\textbf{Selection + additional weighting: extra scoring after selection}} \\
\midrule
Robometer-Prog$^{3}$ & Frame & \cellcolor{yellow!20}0.6600 & \cellcolor{red!12}0.7626 & \cellcolor{yellow!20}0.9099 & \cellcolor{yellow!20}0.5909 & \cellcolor{yellow!20}0.8899 & 0.2734 & \cellcolor{red!12}0.1097 & \cellcolor{red!12}0.0706 & \cellcolor{green!15}\textbf{0.1582} \\
Robometer-Pref$^{3}$ & Frame & 0.6210 & 0.7199 & 0.8887 & 0.4463 & 0.8804 & 0.2395 & 0.1073 & \cellcolor{yellow!20}0.0741 & \cellcolor{red!12}0.1493 \\
PRM-as-Judge$^{4}$ & Frame & \cellcolor{red!12}0.6514 & \cellcolor{yellow!20}0.7645 & \cellcolor{green!15}\textbf{0.9150} & \cellcolor{green!15}\textbf{0.6033} & 0.8862 & \cellcolor{red!12}0.2766 & 0.0971 & 0.0643 & 0.1339 \\
LRMs$^{5}$ & Frame & 0.6282 & 0.7404 & 0.8706 & \cellcolor{red!12}0.5826 & 0.8839 & 0.2634 & \cellcolor{green!15}\textbf{0.1128} & 0.0700 & 0.1443 \\
\textbf{Confidence (Ours)} & \textbf{Frame} & 0.6403 & 0.7332 & \cellcolor{red!12}0.9008 & \cellcolor{green!15}\textbf{0.6033} & \cellcolor{red!12}0.8867 & \cellcolor{yellow!20}0.2790 & 0.1019 & 0.0620 & 0.1455 \\
\textbf{Confidence (Ours)} & \textbf{Fr.+Patch} & \cellcolor{green!15}\textbf{0.6723} & \cellcolor{green!15}\textbf{0.7705} & 0.8864 & \cellcolor{red!12}0.5826 & \cellcolor{green!15}\textbf{0.8903} & \cellcolor{green!15}\textbf{0.3017} & \cellcolor{yellow!20}0.1126 & \cellcolor{green!15}\textbf{0.0837} & \cellcolor{yellow!20}0.1570 \\
\bottomrule
\end{tabular}}
\vspace{-0.3cm}
\end{table*}

\subsection{Controlled Ablations of Features and Threshold Supervision}
\label{app:c3_controlled}

\paragraph{Experimental design.}
We evaluate two design choices for dense confidence estimation in EVAC: the feature layer used by the probe and the supervision threshold range. Following the \cprobe{} formulation~\citep{mei2025c3}, the probe predicts dense confidence from binary targets constructed by thresholding local latent errors. We implement this formulation on a frozen EVAC backbone and compare bottleneck and decoder features under fixed-range and EMA supervision. Throughout this subsection, $q$ denotes the binary confidence target and $\hat q$ denotes the predicted confidence, following Section~\ref{subsec:confidence_probe}.

\paragraph{Backbone, data, and probe training.}
All variants use the same EVAC-v1 checkpoint after 2,000 warmup updates. We partition the original 6,248-episode probe/warmup subset into 5,992 probe-training episodes and 256 calibration episodes using deterministic task-stratified selection. The calibration episodes participated in the earlier EVAC-v1 warmup and are held out during probe optimization. Evaluation uses a separate set of 250 held-out episodes spanning 43 task labels represented in the dataset. Threshold-range selection uses the calibration set, and all variants are evaluated at the final 6,000-step checkpoint.

For each episode, we cache one 20-frame clip consisting of four conditioning frames and $T=16$ future frames, together with one noise realization and the frozen EVAC features. The integer diffusion timestep is sampled from $\tau\in\{150,\ldots,599\}$ for training and matched-noise evaluation. All variants share the cached examples. We use the local latent error $m^{(\tau)}_{t,(i,j)}$ defined in Section~\ref{subsec:confidence_probe} and Appendix~\ref{app:patch_error}: the channel-averaged absolute error between the single-step recovered clean latent $\hat z^{(\tau)}$ and the ground-truth clean latent $z^{(0)}$, spatially pooled to probe patch $(i,j)$. Thresholding this error gives the binary confidence target $q_{t,(i,j)}$.

We compare bottleneck features $h_{\mathrm{mid}}$ at spatial resolution $5\times8$ with decoder features $h_{\mathrm{dec}}^{(\ell)}$ at $\ell=5$ and resolution $20\times32$; both have 1,280 channels. Bottleneck features are bilinearly resized to $20\times32$ before entering the shared probe architecture $f_\phi$. Both variants therefore use the same output grid, with $H_p=20$ and $W_p=32$, the same confidence targets, and 8,157,185 trainable parameters. The probe consists of a 256-dimensional projection, four attention heads, three spatial and three temporal Transformer layers, and diffusion-timestep and threshold conditioning through $e_\tau$ and $e_\theta$. Each variant is trained with seed 42, batch size four, and 6,000 AdamW updates, using a learning rate of $10^{-3}$, weight decay of $0.01$, 300 warmup updates, cosine decay, gradient clipping at 1.0, and bf16 computation. Initialization and training random streams are matched across variants.

\paragraph{Threshold supervision.}
Let $Q_\rho$ denote the $\rho$-quantile of pooled local latent errors in the training cache. We consider four fixed supervision ranges: $[Q_{.10},Q_{.90}]=[0.064462,0.240203]$, $[Q_{.05},Q_{.95}]=[0.053167,0.294761]$, $[Q_{.20},Q_{.80}]=[0.080663,0.186931]$, and the manually specified range $[0.20,0.70]$. Within each fixed range, a threshold $\theta_t$ is sampled uniformly and independently for each sample and future frame, and is shared by all spatial patches within that frame.

EMA supervision follows the update rule in Section~\ref{subsec:confidence_probe} and Appendix~\ref{app:ema_threshold}. The bounds are initialized at $l^{(0)}=0.20$ and $h^{(0)}=0.70$ and updated using the batch-wise 10th and 90th error percentiles. The update rate $\gamma_s$ is 0.20 during the first $S_{\mathrm{warm}}=300$ updates and 0.002 thereafter. At each training step $s$, thresholds are sampled as $\theta_t^{(s)}\sim\mathcal{U}(l^{(s)},h^{(s)})$. An additional variant, EMA-Qinit, initializes the bounds at $[Q_{.10},Q_{.90}]$ to compare fixed and adaptive supervision from the same starting range. Combining two feature choices with six supervision settings yields 12 probe fits.

We select the fixed range by minimizing calibration-set Brier averaged across the two feature choices. This procedure selects $[Q_{.05},Q_{.95}]$, which is also selected when each feature choice is calibrated separately.

\paragraph{Common-threshold evaluation.}
All probes are evaluated at the same nine absolute thresholds:
\begin{equation}
  \theta_k=Q_{.05}+\frac{k-1}{8}(Q_{.95}-Q_{.05}),
  \qquad k=1,\ldots,9.
  \label{eq:c3_shared_thresholds}
\end{equation}
Here, $k$ indexes evaluation thresholds, while $t$ continues to index future frames. We add an episode index $e$ to the notation of Section~\ref{subsec:confidence_probe}. At each $\theta_k$, the probe produces the predicted confidence $\hat q_{e,t,(i,j)}(\theta_k)$ using the corresponding threshold embedding $e_{\theta_k}$. Its binary evaluation target is
\[
  q_{e,t,(i,j)}(\theta_k)
  =\mathbb{I}\bigl\{m^{(\tau)}_{e,t,(i,j)}<\theta_k\bigr\}.
\]
The diffusion timestep $\tau$ is the one sampled for the evaluated example; its dependence is implicit in $q$ and $\hat q$, as in the main text. Using the terminology of Appendix~\ref{app:Additional Confidence Evaluation Details}, this evaluation sets $\theta_{\mathrm{cond}}=\theta_{\mathrm{eval}}=\theta_k$ at each query, so the predicted confidence and binary target refer to the same error threshold. The primary metric is the episode-averaged Brier score:
\begin{equation}
  \mathrm{Brier}
  =\frac{1}{N_{\mathrm{ep}}}
  \sum_{e=1}^{N_{\mathrm{ep}}}
  \frac{1}{9TH_pW_p}
  \sum_{k=1}^{9}\sum_{t=1}^{T}\sum_{i=1}^{H_p}\sum_{j=1}^{W_p}
  \bigl(\hat q_{e,t,(i,j)}(\theta_k)-q_{e,t,(i,j)}(\theta_k)\bigr)^2,
  \label{eq:c3_brier}
\end{equation}
where $N_{\mathrm{ep}}$ is the number of evaluated episodes and $TH_pW_p=16\cdot20\cdot32$.

We also report BCE, ECE, and AUROC. ECE uses 15 equal-width probability bins, pools frame/patch positions across episodes at each threshold, and averages the resulting values over thresholds. AUROC is computed for each episode/threshold pair and averaged over pairs containing both classes. The shared grid provides common conditioning values and evaluation labels for all variants. It spans the training-error interval $[Q_{.05},Q_{.95}]$, including queries outside the supervision range of variants trained on narrower or shifted intervals.

\begin{table}[t]
\vspace{-0.3cm}
\caption{\textbf{Controlled feature and supervision comparison on EVAC-v1.} All rows use seed 42 and identical probe capacity. Tuned fixed denotes the calibration-selected $[Q_{.05},Q_{.95}]$ range. Scores are computed on 250 test episodes with $\tau\in\{150,\ldots,599\}$. Lower Brier, BCE, and ECE and higher AUROC indicate better performance. Boldface marks the best values among the four configurations.}
\label{tab:c3_core}
\centering
\small
\setlength{\tabcolsep}{4pt}
\renewcommand{\arraystretch}{1.08}
\resizebox{0.7\linewidth}{!}{%
\begin{tabular}{l|l|cccc}
\toprule
\rowcolor{gray!12}
Feature & Supervision & Brier $\downarrow$ & BCE $\downarrow$ & ECE $\downarrow$ & AUROC $\uparrow$ \\
\midrule
Bottleneck & Tuned fixed & 0.089487 & 0.288859 & 0.00567 & 0.78379 \\
Bottleneck & EMA & 0.089551 & 0.289220 & 0.00639 & 0.78487 \\
\midrule
Decoder & Tuned fixed & \textbf{0.084181} & \textbf{0.271915} & \textbf{0.00402} & \textbf{0.82237} \\
Decoder & EMA & 0.084243 & 0.272314 & 0.00643 & 0.82184 \\
\bottomrule
\end{tabular}}
\end{table}

\begin{figure}[t]
\centering
\includegraphics[width=0.9\linewidth]{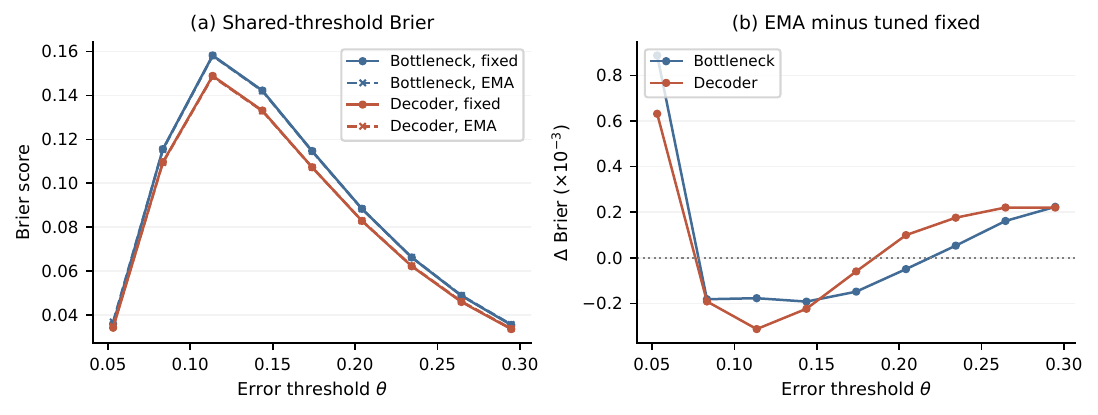}
\vspace{-0.4cm}
\caption{\textbf{Probe performance at shared thresholds.} (a) Brier scores for the four core configurations. (b) EMA-minus-tuned-fixed Brier at each threshold; positive values favor fixed supervision. Fixed and EMA supervision yield closely aligned curves, with a larger separation between decoder and bottleneck features. All configurations use the same binary targets $q_{e,t,(i,j)}(\theta_k)$, constructed from held-out local latent errors.}
\label{fig:c3_shared_thresholds}
\end{figure}

\paragraph{Effect of feature choice.}
Decoder features improve probe quality under both supervision rules (Table~\ref{tab:c3_core}). With tuned fixed supervision, replacing bottleneck features with decoder features reduces Brier from 0.089487 to 0.084181, a relative reduction of 5.93\%, and increases AUROC from 0.78379 to 0.82237. With EMA supervision, Brier decreases from 0.089551 to 0.084243, also a 5.93\% relative reduction. These improvements are measured directly on probe predictions before acquisition and weighted retraining.

\begin{table}[t]
\vspace{-0.3cm}
\caption{\textbf{Complete single-seed range ablation.} All 12 fits use seed 42. Quantiles $Q_\rho$ are computed from training errors. The dagger marks the fixed range selected by calibration Brier, both when averaging across feature choices and when selecting separately for each feature. EMA-Qinit starts from $[Q_{.10},Q_{.90}]$. Test columns use $\tau\in\{150,\ldots,599\}$; the final column evaluates the same episodes with $\tau\in\{50,\ldots,199\}$.}
\label{tab:c3_all_ranges}
\centering
\scriptsize
\setlength{\tabcolsep}{3pt}
\renewcommand{\arraystretch}{1.06}
\resizebox{0.95\linewidth}{!}{%
\begin{tabular}{l|l|ccccc}
\toprule
\rowcolor{gray!12}
Feature & Supervision & Cal. Brier & Test Brier & Test ECE & Test AUROC & Low-noise Brier \\
\midrule
Bottleneck & Fixed $[0.20,0.70]$ & 0.154102 & 0.153483 & 0.12854 & 0.77596 & 0.090326 \\
Bottleneck & Fixed $[Q_{.10},Q_{.90}]$ & 0.092173 & 0.089468 & 0.00670 & 0.78338 & 0.057385 \\
Bottleneck & Fixed $[Q_{.05},Q_{.95}]^{\dagger}$ & 0.092149 & 0.089487 & 0.00567 & 0.78379 & 0.057210 \\
Bottleneck & Fixed $[Q_{.20},Q_{.80}]$ & 0.092776 & 0.090206 & 0.01182 & 0.78185 & 0.058627 \\
Bottleneck & EMA & 0.092239 & 0.089551 & 0.00639 & 0.78487 & 0.057544 \\
Bottleneck & EMA-Qinit & 0.092101 & 0.089477 & 0.00694 & 0.78517 & 0.057503 \\
\midrule
Decoder & Fixed $[0.20,0.70]$ & 0.133136 & 0.133257 & 0.10630 & 0.81088 & 0.085256 \\
Decoder & Fixed $[Q_{.10},Q_{.90}]$ & 0.086885 & 0.084235 & 0.00617 & 0.82313 & 0.054462 \\
Decoder & Fixed $[Q_{.05},Q_{.95}]^{\dagger}$ & 0.086872 & 0.084181 & 0.00402 & 0.82237 & 0.054344 \\
Decoder & Fixed $[Q_{.20},Q_{.80}]$ & 0.087667 & 0.085248 & 0.01142 & 0.82040 & 0.056637 \\
Decoder & EMA & 0.086867 & 0.084243 & 0.00643 & 0.82184 & 0.054400 \\
Decoder & EMA-Qinit & 0.086840 & 0.084266 & 0.00548 & 0.82235 & 0.054522 \\
\bottomrule
\end{tabular}}
\vspace{-0.3cm}
\end{table}

\paragraph{Effect of threshold supervision.}
EMA and tuned fixed supervision yield similar Brier scores across the shared threshold grid (Figure~\ref{fig:c3_shared_thresholds}). For decoder features, Brier is 0.084243 with EMA and 0.084181 with the calibration-selected fixed range, an episode-mean difference of $+6.22\times10^{-5}$. The fixed range also gives lower ECE, at 0.00402 compared with 0.00643 for EMA.

The full range ablation shows a larger effect of supervision-range choice (Table~\ref{tab:c3_all_ranges}). The manual range $[0.20,0.70]$ yields a decoder Brier of 0.133257, compared with 0.084181--0.085248 for the three training-quantile ranges. Initializing EMA at $[Q_{.10},Q_{.90}]$ gives a Brier of 0.084266, close to the 0.084235 obtained by keeping those bounds fixed. Thus, training-error quantiles provide effective supervision ranges in this setting, and EMA tracks the error scale online with performance close to the calibrated fixed alternatives.

\paragraph{Paired task-level comparisons.}
We estimate uncertainty across tasks by averaging episode Brier within each of the 43 test tasks, computing paired method differences, and resampling tasks with replacement 3,000 times. Table~\ref{tab:c3_contrasts} reports task-macro differences and percentile 95\% confidence intervals. Task-macro estimates weight each task equally, while the episode means in Table~\ref{tab:c3_core} weight tasks according to their episode counts. The intervals are conditional on seed 42 and use unadjusted percentile bounds for the exploratory comparisons.

\begin{figure}[t]
\centering
\includegraphics[width=0.9\linewidth]{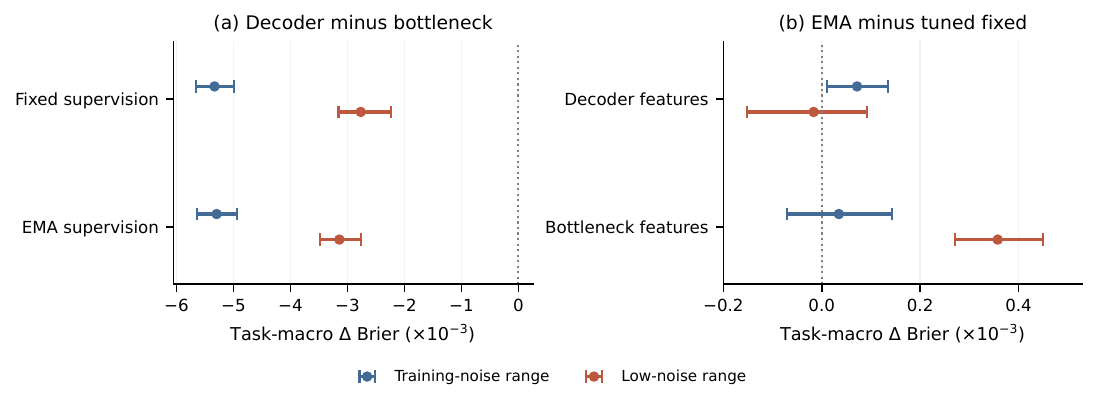}
\vspace{-0.3cm}
\caption{\textbf{Feature and supervision effects across tasks.} (a) Decoder-minus-bottleneck Brier contrasts under fixed and EMA supervision. (b) EMA-minus-tuned-fixed contrasts for each feature choice. The panels use different horizontal scales to display the respective effect sizes. Error bars indicate percentile 95\% task-bootstrap intervals at seed 42.}
\label{fig:c3_factorial_contrasts}
\end{figure}

The decoder-minus-bottleneck intervals lie below zero under both supervision rules. Within the training-noise range, the decoder EMA-minus-fixed difference is small and positive, with an interval above zero, indicating a slight advantage for fixed supervision. The corresponding bottleneck interval includes zero. Figure~\ref{fig:c3_factorial_contrasts} shows that the feature effect is substantially larger than the difference between EMA and tuned fixed supervision.

\begin{table}[t]
\caption{\textbf{Paired task-macro Brier contrasts.} All values, including interval endpoints, are multiplied by $10^3$. Intervals are percentile 95\% task-bootstrap intervals at seed 42. Negative differences favor the first named configuration. Let $B_{a,b}$ denote task-macro Brier for feature choice $a$ and supervision rule $b$, with $\mathrm{dec}$ and $\mathrm{mid}$ denoting decoder and bottleneck features. The interaction is $(B_{\mathrm{dec,EMA}}-B_{\mathrm{mid,EMA}})-(B_{\mathrm{dec,fixed}}-B_{\mathrm{mid,fixed}})$.}
\label{tab:c3_contrasts}
\centering
\scriptsize
\setlength{\tabcolsep}{3pt}
\renewcommand{\arraystretch}{1.07}
\resizebox{0.85\linewidth}{!}{%
\begin{tabular}{l|cc|cc}
\toprule
\rowcolor{gray!12}
& \multicolumn{2}{c|}{Training-noise range} & \multicolumn{2}{c}{Low-noise range} \\
\rowcolor{gray!12}
Contrast & Difference & 95\% CI & Difference & 95\% CI \\
\midrule
Decoder $-$ bottleneck (fixed) & $-5.331$ & $[-5.657, -4.987]$ & $-2.767$ & $[-3.156, -2.243]$ \\
Decoder $-$ bottleneck (EMA) & $-5.294$ & $[-5.637, -4.934]$ & $-3.141$ & $[-3.475, -2.762]$ \\
EMA $-$ fixed (decoder) & $+0.072$ & $[+0.011, +0.134]$ & $-0.016$ & $[-0.151, +0.091]$ \\
EMA $-$ fixed (bottleneck) & $+0.035$ & $[-0.071, +0.143]$ & $+0.358$ & $[+0.270, +0.449]$ \\
Feature $\times$ supervision & $+0.037$ & $[-0.090, +0.170]$ & $-0.374$ & $[-0.562, -0.229]$ \\
\bottomrule
\end{tabular}}
\vspace{-0.3cm}
\end{table}

\paragraph{Evaluation at lower noise levels.}
We further evaluate the same test episodes at $\tau\in\{50,\ldots,199\}$ with fixed probes and threshold grids. This range partially overlaps the training range and is closer to acquisition-time scoring. Decoder features achieve lower Brier under both supervision rules: 0.054344 versus 0.057210 with fixed supervision, and 0.054400 versus 0.057544 with EMA. The decoder EMA-minus-fixed task-bootstrap interval includes zero (Table~\ref{tab:c3_contrasts}).

\begin{figure}[t]
\centering
\includegraphics[width=0.9\linewidth]{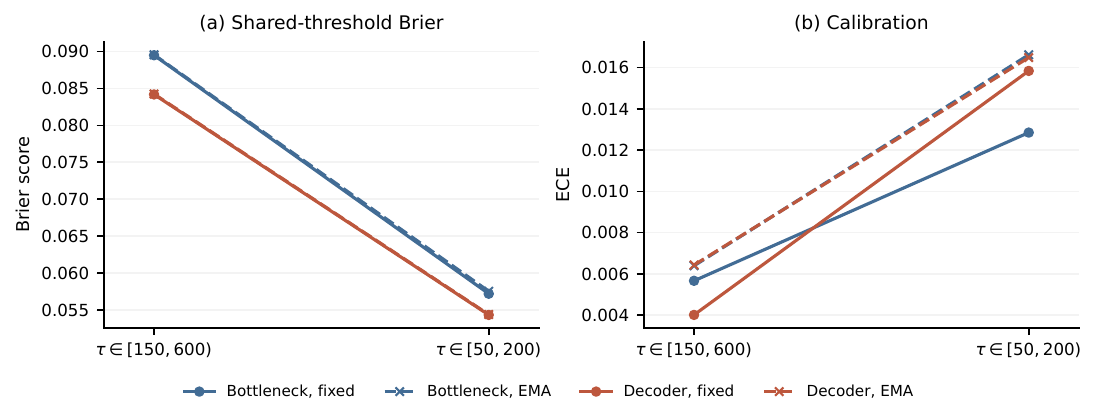}
\vspace{-0.3cm}
\caption{\textbf{Noise-range transfer on held-out episodes.} Brier (left) and ECE (right) are evaluated on the same episodes with fixed probe parameters and threshold ranges. Decoder features achieve lower Brier in both noise ranges. Low-noise evaluation shows lower Brier but higher ECE due to changes in local-error distributions and target balance.}
\vspace{-0.4cm}
\label{fig:c3_noise_transfer}
\end{figure}

At lower noise levels, Brier decreases while ECE increases (Figure~\ref{fig:c3_noise_transfer}). For decoder features, ECE increases from 0.00402 to 0.01584 with fixed supervision and from 0.00643 to 0.01648 with EMA. The diffusion timestep changes the distribution of $m^{(\tau)}_{e,t,(i,j)}$ and the positive-target ratio $q_{e,t,(i,j)}(\theta_k)=1$, so cross-range comparisons depend on both prediction quality and target statistics.

\paragraph{Alignment of confidence-derived weights with local error.}
We examine how probe predictions translate into weighting signals by applying the same transformation to every variant. Following the risk definition in Section~\ref{subsec:confidence_probe}, the risk at threshold $\theta_k$ is $r_{e,t,(i,j)}(\theta_k)=1-\hat q_{e,t,(i,j)}(\theta_k)$. We average this risk over the nine evaluation thresholds:
\[
  \bar r_{e,t,(i,j)}
  =\frac{1}{9}\sum_{k=1}^{9}r_{e,t,(i,j)}(\theta_k)
  =1-\frac{1}{9}\sum_{k=1}^{9}\hat q_{e,t,(i,j)}(\theta_k).
\]
Here, the bar denotes averaging over thresholds. We normalize $\bar r$ using each episode's 5th and 95th percentiles and clip the result to $[0,1]$, denoting the normalized risk by $\widetilde{\bar r}_{e,t,(i,j)}$. The diagnostic weights are formed as $1+\widetilde{\bar r}_{e,t,(i,j)}$, normalized to unit mean within each episode, and clipped to $[0.5,3]$. This transformation uses the patch-level weighting setting $\alpha=1$ and unit weighting strength, with the nine-threshold average supplying the risk input. We evaluate the resulting weights directly on the common probe grid before task-budget allocation and EVAC retraining.

\begin{figure}[t]
\centering
\includegraphics[width=0.85\linewidth]{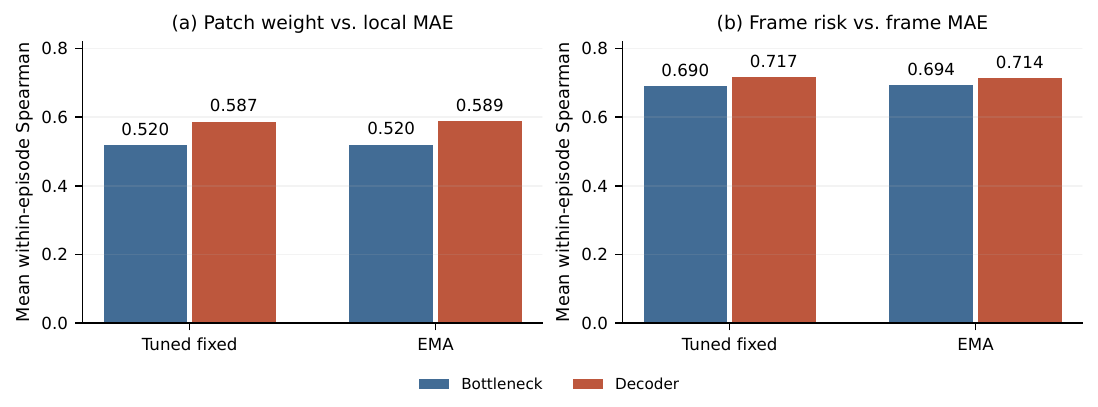}
\vspace{-0.3cm}
\caption{\textbf{Alignment of confidence-derived signals with prediction error.} (a) Mean within-episode Spearman correlation between normalized patch weights and local latent MAE. (b) Mean within-episode Spearman correlation between spatially averaged frame risk and frame MAE. Risk is averaged over the nine evaluation thresholds before constructing these diagnostics. All variants use the same weighting transformation, and values are computed before EVAC retraining at seed 42.}
\vspace{-0.2cm}
\label{fig:c3_weight_diagnostics}
\end{figure}

With fixed supervision, the mean within-episode Spearman correlation between patch weights and local latent error $m^{(\tau)}_{e,t,(i,j)}$ increases from 0.51954 for bottleneck features to 0.58734 for decoder features. With EMA, it increases from 0.52043 to 0.58859. Correlations between spatially averaged frame risk and frame MAE show the same feature trend (Figure~\ref{fig:c3_weight_diagnostics}). The stronger local-error alignment of decoder predictions is therefore retained after the weighting transformation.

\begin{figure}[t]
\centering
\includegraphics[width=0.9\linewidth]{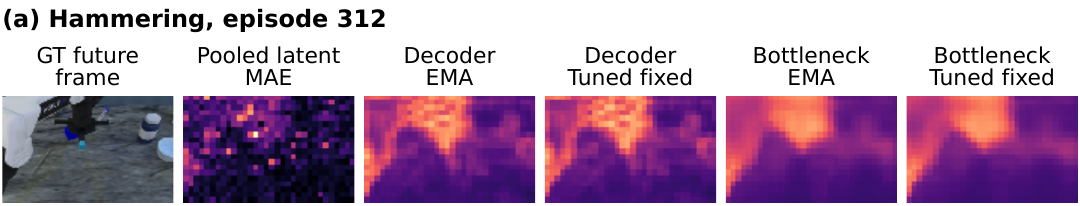}
\par\medskip
\includegraphics[width=0.9\linewidth]{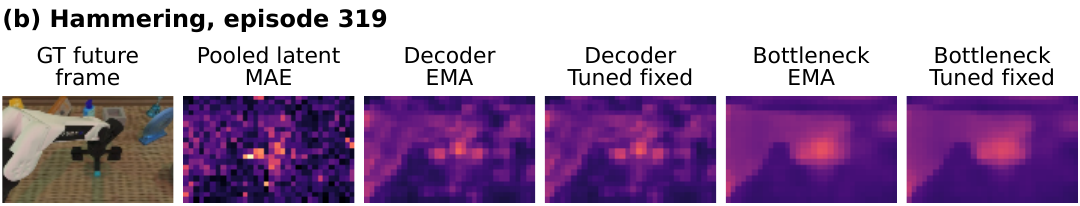}
\vspace{-0.2cm}
\caption{\textbf{Qualitative comparison of feature choices and threshold supervision.} Two held-out episodes from the hammering task. Each row shows the first ground-truth future frame, pooled latent MAE, and probe responses for decoder features with EMA and tuned fixed supervision, followed by bottleneck features with EMA and tuned fixed supervision. Decoder responses contain finer spatial structure, while EMA and tuned fixed supervision yield similar patterns within each feature choice.}
\vspace{-0.2cm}
\label{fig:c3_qualitative}
\end{figure}

\paragraph{Qualitative comparison.}
Figure~\ref{fig:c3_qualitative} shows two episodes from the hammering task, together with pooled latent-error maps and the responses of the four probe configurations. Decoder-based responses exhibit finer spatial variation, while bottleneck-based responses are smoother and cover broader regions. The central response in the second example illustrates this difference. Within each feature choice, EMA and tuned fixed supervision produce similar spatial patterns, consistent with their closely matched quantitative results.

\paragraph{Discussion.}
Decoder features consistently improve confidence prediction and error alignment at matched probe capacity and resolution. This reflects both finer spatial information and decoder feature fusion. Training-quantile fixed ranges and EMA achieve similar Brier scores, with the calibrated fixed range slightly outperforming EMA for decoder features under training noise. The experiments characterize probe behavior on frozen EVAC-v1 with one seed and one cached clip/noise realization per episode. Evaluations with additional seeds, free-running rollouts, and other architectures would further examine generalization. Crossed retraining experiments could quantify interactions among probe design, acquisition, and weighting in EVAC-v2.

%% file: sections/7_appendix_B2.tex
\subsection{Direct-Error Selection from the Complete Candidate Pool}
\label{app:oracle_error}

\paragraph{Direct-error acquisition.}
The candidate pool contains complete, pre-collected demonstrations, making future ground-truth frames available for offline selection. We rank all 18,244 episodes by the mean squared difference between EVAC-v1's generated latents and their ground-truth counterparts, averaged over channels, spatial locations, and the first 16 predicted future frames. The four conditioning frames are excluded. This rollout reconstruction score complements the local denoising error used to supervise the confidence probe in Section~\ref{subsec:confidence_probe}. The 7,298 highest-error episodes are selected for unweighted retraining under the 4,000-update protocol in Section~\ref{subsec:exp_setup}. Confidence uses the task-level allocation described in Appendix~\ref{app:selection_protocol}.

\begin{table*}[h]
\caption{
\textbf{Direct-error and confidence-based selection.}
Both models use optimizer seed 42, 7,298 selected episodes,
4,000 updates, and no loss weighting.
Confidence (Selection) uses the seed-42 result in Table~\ref{tab:al_seed42},
while Table~\ref{tab:main_results} reports the average
over three confidence runs.
Bold values indicate the better result in each column.
All metrics are higher-is-better.
}
\label{tab:supp_oracle}
\centering
\scriptsize
\setlength{\tabcolsep}{3pt}
\renewcommand{\arraystretch}{1.05}
\resizebox{0.95\textwidth}{!}{%
\begin{tabular}{l||cc|c|ccc|ccc}
\toprule
\rowcolor{gray!12}
\multirow{2}{*}{Selection}
& \multicolumn{2}{c}{Reconstruction ($\uparrow$)}
& \multicolumn{1}{c}{Scene ($\uparrow$)}
& \multicolumn{3}{c}{Semantics ($\uparrow$)}
& \multicolumn{3}{c}{Motion ($\uparrow$)} \\
\cmidrule{2-10}
\rowcolor{gray!12}
& PSNR & SSIM & Scene Cons.
& Logics & Sem.-CLIP & Sem.-BLEU
& Traj-HSD & Traj-Dyn & Traj-nDTW \\
\midrule
Oracle error
& 0.6458
& 0.7542
& 0.8988
& 0.5826
& 0.8875
& \textbf{0.2914}
& 0.0985
& 0.0587
& 0.1396 \\
Confidence
& \textbf{0.6838}
& \textbf{0.7797}
& \textbf{0.9322}
& \textbf{0.6198}
& \textbf{0.8894}
& 0.2754
& \textbf{0.1313}
& \textbf{0.0921}
& \textbf{0.1780} \\
\bottomrule
\end{tabular}%
}
\end{table*}

\begin{figure*}[h]
\centering
\includegraphics[width=0.98\linewidth]{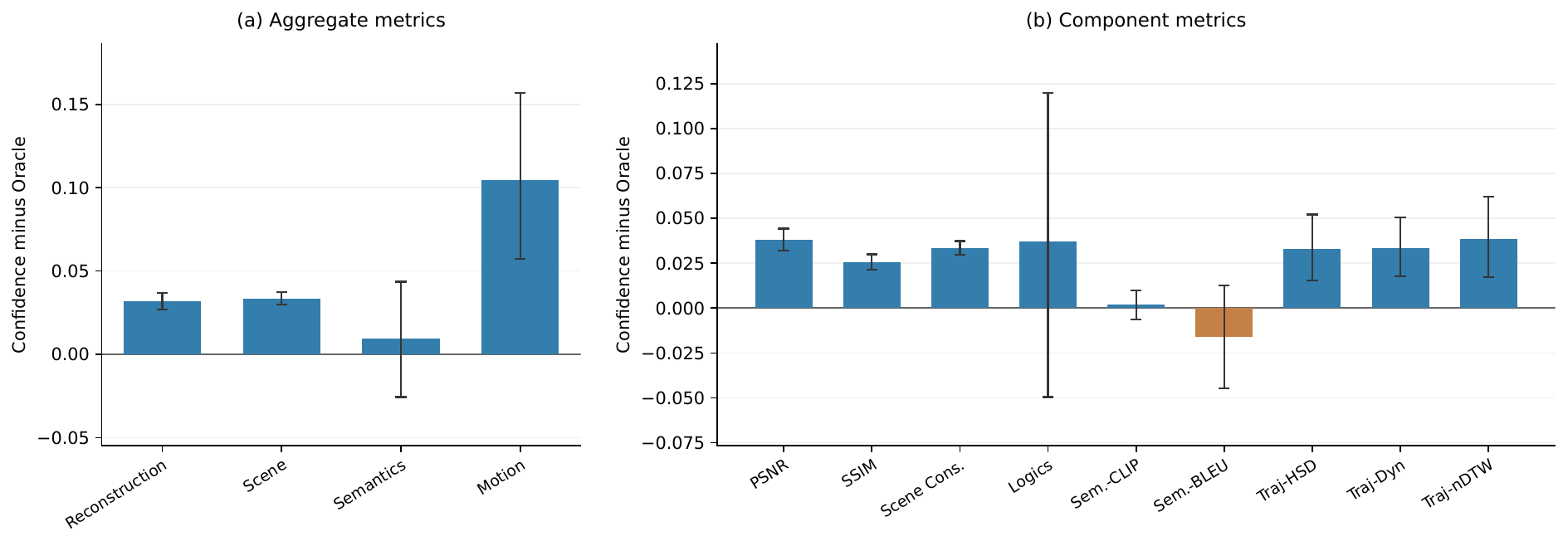}
\caption{\textbf{Confidence versus direct-error selection.}
Bars show Confidence minus Oracle for four aggregate metrics (left) and nine component metrics (right), using the unweighted models in Table~\ref{tab:supp_oracle}. Positive values favor Confidence. Error bars denote paired 95\% episode-bootstrap intervals. Most comparisons use 242 common episodes; Sem.-CLIP, Sem.-BLEU, and aggregate Semantics use 190 common episodes.}
\label{fig:supp_oracle}
\end{figure*}

\paragraph{Results.}
Confidence-guided task allocation exceeds global top-error selection on eight of nine component metrics in Table~\ref{tab:supp_oracle}; Oracle attains the higher Sem.-BLEU. The aggregate gains are $0.0318$ in Reconstruction, $0.0334$ in Scene, and $0.1045$ in Motion. Figure~\ref{fig:supp_oracle} presents paired differences and their uncertainty at both aggregate and component levels. This comparison evaluates two acquisition policies: global ranking by current rollout MSE and confidence-guided task quotas with within-task random sampling. Their differences jointly reflect the scoring signal and the distribution of selected episodes across tasks.

\subsection{Data-Budget Sensitivity under Fixed Short Optimization}
\label{app:budget_sensitivity}

\paragraph{Protocol.}
We compare unweighted confidence and uniform-task random selection at 10\%, 20\%, and 40\% of the candidate pool, together with a 100\% full-pool reference. All models receive 750 updates with effective batch size 16 and optimizer seed 42. Selection also uses seed 42. The two selection methods share the quota allocator and within-task sampling procedure; random selection assigns equal weights to tasks. Evaluation uses a fixed 64-episode subset spanning the 32 tasks represented in the original validation split. This sweep measures data-budget sensitivity under the same short optimization schedule.

\begin{table}[t]
\caption{\textbf{Budget sensitivity under 750-update post-training.}
All models use optimizer seed 42 and no loss weighting. Bold values mark the higher score within each confidence/random pair. Aggregates follow the main-text definitions and use each component's available mean. Reconstruction, Scene, and Motion cover 62 episodes; CLIP/BLEU cover 56 episodes.}
\label{tab:supp_budget}
\centering\scriptsize
\setlength{\tabcolsep}{3pt}\renewcommand{\arraystretch}{1.05}
\resizebox{0.95\linewidth}{!}{%
\begin{tabular}{c|c|c||cccc|c}
\toprule
\rowcolor{gray!12}
Budget & Selection & Episodes & Reconstruction $\uparrow$ & Scene $\uparrow$ & Semantics $\uparrow$ & Motion $\uparrow$ & Train GPU-h\\
\midrule
10\% & Random & 1,824 & 0.6341 & 0.8772 & 0.5007 & 0.1701 & 4.47\\
10\% & Confidence & 1,824 & \textbf{0.6995} & \textbf{0.9088} & \textbf{0.5900} & \textbf{0.2502} & 4.45\\
\midrule
20\% & Random & 3,649 & \textbf{0.6679} & \textbf{0.9003} & 0.5391 & \textbf{0.2360} & 4.45\\
20\% & Confidence & 3,649 & 0.6642 & 0.8812 & \textbf{0.6167} & 0.1970 & 4.44\\
\midrule
40\% & Random & 7,298 & \textbf{0.6874} & 0.8786 & 0.5576 & 0.2004 & 4.44\\
40\% & Confidence & 7,298 & 0.6611 & \textbf{0.8981} & \textbf{0.5987} & \textbf{0.2305} & 4.47\\
\midrule
100\% & Full pool & 18,244 & 0.6902 & 0.9030 & 0.5263 & 0.2023 & 4.46\\
\bottomrule
\end{tabular}}
\end{table}

\begin{figure*}[t]
\centering
\includegraphics[width=0.98\linewidth]{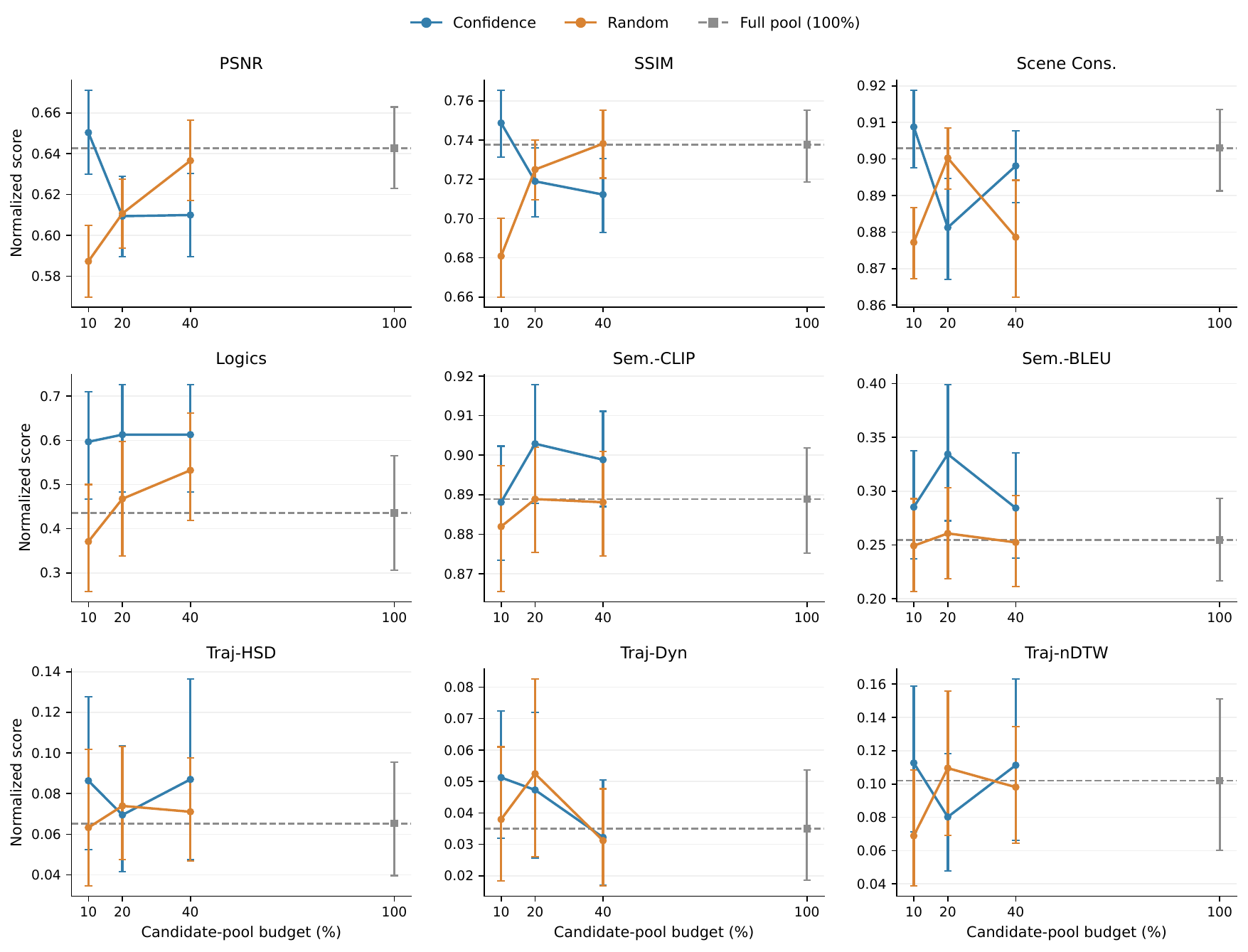}
\caption{\textbf{Budget--performance curves under fixed optimization.}
All metrics are higher-is-better. Confidence and random receive the same 750 updates. In each panel, the gray square marks the 100\% full-pool result, and the gray horizontal dashed line extends its mean across budgets for reference. Error bars show 95\% episode-bootstrap intervals, using 62 common episodes for most metrics and 56 for CLIP/BLEU. Lines between measured budget points serve as visual guides.}
\vspace{-0.4cm}
\label{fig:supp_budget}
\end{figure*}

\paragraph{Results and effect of data budget.}
Under the fixed 750-update schedule, performance depends on both the acquisition strategy and the available data budget. At 10\%, confidence improves all four aggregate scores over random, with positive paired intervals in Table~\ref{tab:supp_budget_ci}, and its point estimates also exceed the full-pool reference. At larger budgets, the gains become metric-dependent: confidence improves Semantics at 20\%, and Scene, Semantics, and Motion at 40\%, while random achieves higher Reconstruction at both budgets. Each run processes 12,000 sample presentations, corresponding to approximately 6.58 passes over the 10\% subset but only 0.66 passes over the full pool of 18,244 eligible episodes. Thus, increasing the pool broadens data coverage while reducing repeated exposure under the same optimization budget, providing one explanation for the observed non-monotonic trends. Overall, the clearest advantage of confidence-guided selection appears at the smallest tested data budget, and the full-pool result serves as a reference under the same fixed-update setting.

\begin{table}[t]
\caption{\textbf{Paired confidence-minus-random differences.}
Each entry gives the paired mean and its 95\% interval from 10,000 episode-bootstrap resamples. Larger differences favor confidence. Reconstruction, Scene, and Motion use 62 common episodes; Semantics uses the 56-episode intersection of its three components. This common-episode calculation can differ slightly from subtracting the component-wise aggregates in Table~\ref{tab:supp_budget}.}
\label{tab:supp_budget_ci}
\centering\scriptsize
\setlength{\tabcolsep}{3pt}\renewcommand{\arraystretch}{1.08}
\resizebox{\linewidth}{!}{%
\begin{tabular}{c||cccc}
\toprule
\rowcolor{gray!12}
Budget & $\Delta$Reconstruction $\uparrow$ & $\Delta$Scene $\uparrow$ & $\Delta$Semantics $\uparrow$ & $\Delta$Motion $\uparrow$\\
\midrule
10\% & $0.0655\ [0.0534,0.0772]$ & $0.0316\ [0.0229,0.0396]$ & $0.0914\ [0.0247,0.1563]$ & $0.0801\ [0.0293,0.1391]$\\
20\% & $-0.0037\ [-0.0186,0.0112]$ & $-0.0190\ [-0.0296,-0.0089]$ & $0.0768\ [0.0119,0.1425]$ & $-0.0389\ [-0.1023,0.0185]$\\
40\% & $-0.0263\ [-0.0411,-0.0120]$ & $0.0195\ [0.0031,0.0362]$ & $0.0440\ [-0.0164,0.1021]$ & $0.0301\ [-0.0570,0.1390]$\\
\bottomrule
\end{tabular}}
\end{table}

\paragraph{Cost accounting.}
All short-budget models require approximately 2.22--2.23 wall-clock hours on two A800 GPUs (4.44--4.47 GPU-hours). Selecting fewer eligible episodes therefore does not reduce training compute under this fixed-update protocol. The 4,000-update Oracle run takes 11.67 hours on two A800 GPUs (23.35 GPU-hours). Given generated validation videos, EWMBench scoring takes 2.09 hours on one A800 for the 250-episode manifest, and 0.51--0.57 hours for the 64-episode subset. Generating these videos takes a further 4.66 and 1.16--1.21 hours, respectively, on two A800 GPUs. These timings separate post-training, rollout generation, and metric evaluation; acquisition and probe-training overhead require additional accounting in an end-to-end comparison.

\subsection{Selection, Weighting, and Evaluation Protocol}
\label{app:selection_protocol}

\paragraph{Task scores and sampling quotas.}
Each task is represented by one randomly chosen episode and scored using $S_{\mathrm{mean}}$ from Section~\ref{subsec:confidence_al}. Let $K$ denote the number of distinct task scores. We rank these distinct scores in ascending order as $k=0,\ldots,K-1$, with tasks sharing the same score assigned the same $k$. The allocation multiplier for a task with rank $k$ is
\begin{equation}
w_{\mathrm{task}}(k)=
0.25
+
1.75\frac{k}{\max(K-1,1)}.
\label{eq:supp_quota_weights}
\end{equation}
Thus, the multipliers range linearly from $0.25$ to $2.0$ when $K>1$, while $w_{\mathrm{task}}=0.25$ for all tasks when $K=1$. Each nonempty task first receives one episode. The remaining budget is allocated proportionally to $w_{\mathrm{task}}$ using floored shares, with residual units assigned by largest fractional remainder. Tasks are capped by their available episode counts, and any unused quota is redistributed among the remaining tasks. A single Python random-number generator with seed 42 shuffles each task's episode list in lexicographic task order. Setting all task scores equal recovers the uniform-task random baseline.

\paragraph{Features and confidence weighting.}
We use the decoder features $h_{\mathrm{dec}}^{(\ell)}$ with $\ell=5$, corresponding to zero-based \texttt{output\_blocks[5]} and a $1280\times20\times32$ feature tensor. The frame-and-patch setting uses $\alpha=0.5$ in $r^{(\alpha)}_{t,(i,j)}$. After interpolation to the training-loss grid, the 5th and 95th percentiles are computed separately for each video's complete time--space risk tensor. Subtracting the lower percentile, dividing by the percentile range plus $10^{-6}$, and clipping to $[0,1]$ gives $\widetilde r^{(\alpha)}_{t,(i,j)}$. We use the main-text schedule $\lambda_{\mathrm{eff}}(s)=\lambda_{\mathrm{conf}}\min(1,s/s_{\mathrm{warm}})$ with $\lambda_{\mathrm{conf}}=1$ and $s_{\mathrm{warm}}=1000$. The resulting $w_{t,(i,j)}$ is normalized to unit mean per video and clipped to $[0.5,3]$. Confidence maps remain detached during world-model training.

\begin{table}[t]
\caption{\textbf{Selection and retraining settings.}
The weighting parameters apply to the weighted experiments in the main text. The new Oracle and budget-sweep models use unweighted losses.}
\label{tab:supp_settings}
\centering\small
\setlength{\tabcolsep}{4pt}\renewcommand{\arraystretch}{1.08}
\begin{tabular}{l|l}
\toprule
\rowcolor{gray!12} Setting & Value\\
\midrule
Task representatives / acquisition score & 1 per task / $S_{\mathrm{mean}}$\\
Rank-based allocation multiplier & $0.25$--$2.0$\\
Decoder index $\ell$ & 5 (zero-based)\\
Frame-and-patch mixing $\alpha$ & 0.5\\
Confidence strength $\lambda_{\mathrm{conf}}$ & 1.0\\
Weighting warmup $s_{\mathrm{warm}}$ & 1,000 updates\\
Risk normalization percentiles & 5th and 95th, per video\\
Weight clipping after mean normalization & $[0.5,3.0]$\\
Probe / EVAC-v1 training & 6,000 / 2,000 updates\\
Selected-data initialization & Original pretrained EVAC\\
Effective training batch size & 16 (8 per GPU, two GPUs)\\
\bottomrule
\end{tabular}
\end{table}

\paragraph{Reward and progress baselines.}
Each baseline scores generated and ground-truth videos. RoboReward parses progress $s_{\rm raw}\in\{1,\ldots,5\}$; we read its normalized \texttt{score}, $s=(s_{\rm raw}-1)/4$, with $s=0$ for parse failures. GVL reads \texttt{task\_completion\_percentage} on $[0,100]$. All 1,400 cached values are zero or above 1, so the saved scores equal $s_{\rm raw}/100$; replaying this percentage conversion leaves scores and selection risks unchanged. Both Robometer variants read \texttt{progress\_pred} in $[0,1]$. Prog and Pref denote the \texttt{progress} and \texttt{preference} adapter modes, respectively; both reported variants use the progress output head and the common GT-relative risk calculation below. LRMs read per-frame \texttt{score} from completion progress in $[0,1]$. PRM-as-Judge parses signed percentages $p\in[-100,100]$ and uses backward progress $s=1+p/100$. We clip these converted scores to $[0,1]$, linearly resample both sequences to the longer length, and broadcast scalars. The aligned scores $s^{(b)}_{\mathrm{GT},t}$ and $s^{(b)}_{\mathrm{pred},t}$ define risk
\begin{equation}
r^{(b)}_t = 1- \frac{s^{(b)}_{\mathrm{pred},t}-s^{(b)}_{\mathrm{GT},t}+1}{2}=\frac{1+s^{(b)}_{\mathrm{GT},t}-s^{(b)}_{\mathrm{pred},t}}{2}.
\label{eq:supp_baseline_risk}
\end{equation}
Larger ground-truth-relative deficits therefore receive higher risk. We average these frame risks and apply the same task-quota allocator used for confidence.

\paragraph{Evaluation and statistical summaries.}
PSNR is divided by 30 and CLIPScore by 100; normalized components are clipped to $[0,1]$. A shared GT-trajectory filter retains 242/250 full-set episodes and 62/64 subset episodes for all methods. The reported semantic results use 190 and 56 CLIP/BLEU records, respectively, because shortened episode IDs in these evaluation outputs collide. Each metric is averaged over its available records; paired comparisons use records shared by the relevant methods and components. For the new direct-error and budget-sweep comparisons, 10,000 episode-bootstrap resamples with seed 42 quantify evaluation-set variability at fixed checkpoints. Main comparisons pool paired episode differences across three runs (Appendix~\ref{app:al_full_results}); probe ablations use 3,000 task-bootstrap resamples (Appendix~\ref{app:c3_controlled}).

Full task/episode/trial IDs were used to verify every retained CLIP/BLEU record against its corresponding GT caption. All 190/56 retained prediction--GT pairs are correctly matched, and all methods use the same fixed subsets. Overwritten records are excluded consistently, with no unresolved ID ambiguity among the retained records.

\paragraph{Random seeds.}
All single-seed experiments use seed 42, including the Confidence reference in Table~\ref{tab:supp_oracle}, the Oracle experiment, and the seven short-budget models. The main learned-method comparison additionally reports averages over three runs with seeds 42, 3407, and 123, following the setting in Section~\ref{subsec:exp_setup}.

%% file: sections/7_appendix_B_Qualitative.tex
\subsection{Qualitative Evolution from Base EVAC to EVAC-v2}
\label{app:training_evolution}

We further visualize how world-model predictions evolve throughout the
post-training pipeline. We select six representative RoboTwin2.0 episodes
covering object placement, cabinet interaction, block stacking, bowl stacking,
and switch manipulation. These examples expose a consistent progression from
severe zero-shot cross-embodiment failure to target-domain adaptation and,
finally, confidence-guided refinement.

\paragraph{Per-episode progression.}
Table~\ref{tab:app_training_evolution_aggregated} summarizes the four aggregated
evaluation dimensions for the six selected episodes. All relative changes are
computed with respect to EVAC-v1, which serves as the common post-warmup baseline.
Across the selected examples, Base EVAC performs substantially worse under
zero-shot cross-embodiment transfer, while EVAC-v1 already recovers much of the
target-domain prediction quality. Confidence-guided EVAC-v2 further improves
most dimensions, with frame-and-patch weighting achieving the strongest
Reconstruction in all six episodes.

For this per-episode analysis, Semantics is aggregated from normalized
Sem.-CLIP and Sem.-BLEU only; the binary Logics score is omitted because a
single $0/1$ decision would disproportionately affect an individual episode.

\begin{table*}[h]
\centering
\caption{
\textbf{Aggregated evolution on six representative RoboTwin2.0 episodes.}
Each entry reports the normalized aggregate followed by its relative change from EVAC-v1.
Green and red percentages indicate improvement and degradation, respectively.
For Motion, relative changes are omitted when the EVAC-v1 score is zero.
Green, yellow, and red cells indicate the best, second-best, and third-best values within each episode.
}
\label{tab:app_training_evolution_aggregated}
\scriptsize
\setlength{\tabcolsep}{3.8pt}
\renewcommand{\arraystretch}{1.05}
\resizebox{0.86\textwidth}{!}{
\begin{tabular}{c|l||cccc}
\toprule
\rowcolor{gray!12}
Episode & Variant & Reconstruction ($\uparrow$) & Scene ($\uparrow$) & Semantics ($\uparrow$) & Motion ($\uparrow$) \\
\midrule

\multirow{4}{*}{\shortstack[c]{place burger fries\\(ep130)}}
& Base EVAC & 0.4592\,\textcolor{red!70!black}{(-27.3\%)} & 0.7890\,\textcolor{red!70!black}{(-10.0\%)} & 0.3591\,\textcolor{red!70!black}{(-28.9\%)} & 0.0000\,\textcolor{red!70!black}{(-100.0\%)} \\
& EVAC-v1 & \cellcolor{red!12}0.6318\,\textcolor{gray}{(0.0\%)} & \cellcolor{red!12}0.8770\,\textcolor{gray}{(0.0\%)} & \cellcolor{red!12}0.5050\,\textcolor{gray}{(0.0\%)} & \cellcolor{red!12}0.4570\,\textcolor{gray}{(0.0\%)} \\
& EVAC-v2 Frame & \cellcolor{yellow!20}0.6618\,\textcolor{green!45!black}{(+4.7\%)} & \cellcolor{green!15}\textbf{0.9430}\,\textcolor{green!45!black}{(+7.5\%)} & \cellcolor{yellow!20}0.5221\,\textcolor{green!45!black}{(+3.4\%)} & \cellcolor{yellow!20}1.0830\,\textcolor{green!45!black}{(+137.0\%)} \\
& EVAC-v2 Fr.+Patch & \cellcolor{green!15}\textbf{0.7020}\,\textcolor{green!45!black}{(+11.1\%)} & \cellcolor{yellow!20}0.9370\,\textcolor{green!45!black}{(+6.8\%)} & \cellcolor{green!15}\textbf{0.5486}\,\textcolor{green!45!black}{(+8.6\%)} & \cellcolor{green!15}\textbf{1.3080}\,\textcolor{green!45!black}{(+186.2\%)} \\

\midrule
\multirow{4}{*}{\shortstack[c]{place empty cup\\(ep294)}}
& Base EVAC & 0.4605\,\textcolor{red!70!black}{(-38.6\%)} & 0.7120\,\textcolor{red!70!black}{(-27.0\%)} & \cellcolor{yellow!20}0.5454\,\textcolor{green!45!black}{(+2.0\%)} & 0.0000\,\textcolor{red!70!black}{(-100.0\%)} \\
& EVAC-v1 & \cellcolor{yellow!20}0.7503\,\textcolor{gray}{(0.0\%)} & \cellcolor{green!15}\textbf{0.9750}\,\textcolor{gray}{(0.0\%)} & \cellcolor{red!12}0.5349\,\textcolor{gray}{(0.0\%)} & \cellcolor{yellow!20}0.5600\,\textcolor{gray}{(0.0\%)} \\
& EVAC-v2 Frame & \cellcolor{red!12}0.7353\,\textcolor{red!70!black}{(-2.0\%)} & \cellcolor{yellow!20}0.9700\,\textcolor{red!70!black}{(-0.5\%)} & 0.2604\,\textcolor{red!70!black}{(-51.3\%)} & \cellcolor{red!12}0.4150\,\textcolor{red!70!black}{(-25.9\%)} \\
& EVAC-v2 Fr.+Patch & \cellcolor{green!15}\textbf{0.7558}\,\textcolor{green!45!black}{(+0.7\%)} & \cellcolor{red!12}0.9610\,\textcolor{red!70!black}{(-1.4\%)} & \cellcolor{green!15}\textbf{0.6015}\,\textcolor{green!45!black}{(+12.5\%)} & \cellcolor{green!15}\textbf{0.5900}\,\textcolor{green!45!black}{(+5.4\%)} \\

\midrule
\multirow{4}{*}{\shortstack[c]{put object cabinet\\(ep282)}}
& Base EVAC & 0.4372\,\textcolor{red!70!black}{(-23.2\%)} & 0.7770\,\textcolor{red!70!black}{(-10.4\%)} & 0.4856\,\textcolor{red!70!black}{(-0.5\%)} & 0.0000\,\textcolor{gray}{(--)} \\
& EVAC-v1 & \cellcolor{red!12}0.5693\,\textcolor{gray}{(0.0\%)} & \cellcolor{red!12}0.8670\,\textcolor{gray}{(0.0\%)} & \cellcolor{red!12}0.4879\,\textcolor{gray}{(0.0\%)} & 0.0000\,\textcolor{gray}{(0.0\%)} \\
& EVAC-v2 Frame & \cellcolor{yellow!20}0.6178\,\textcolor{green!45!black}{(+8.5\%)} & \cellcolor{green!15}\textbf{0.9260}\,\textcolor{green!45!black}{(+6.8\%)} & \cellcolor{yellow!20}0.5264\,\textcolor{green!45!black}{(+7.9\%)} & 0.0000\,\textcolor{gray}{(--)} \\
& EVAC-v2 Fr.+Patch & \cellcolor{green!15}\textbf{0.6207}\,\textcolor{green!45!black}{(+9.0\%)} & \cellcolor{yellow!20}0.8950\,\textcolor{green!45!black}{(+3.2\%)} & \cellcolor{green!15}\textbf{0.5651}\,\textcolor{green!45!black}{(+15.8\%)} & 0.0000\,\textcolor{gray}{(--)} \\

\midrule
\multirow{4}{*}{\shortstack[c]{stack blocks two\\(ep325)}}
& Base EVAC & 0.3793\,\textcolor{red!70!black}{(-38.4\%)} & 0.7970\,\textcolor{red!70!black}{(-10.4\%)} & 0.4391\,\textcolor{red!70!black}{(-4.6\%)} & 0.0000\,\textcolor{red!70!black}{(-100.0\%)} \\
& EVAC-v1 & \cellcolor{red!12}0.6160\,\textcolor{gray}{(0.0\%)} & \cellcolor{red!12}0.8900\,\textcolor{gray}{(0.0\%)} & \cellcolor{red!15}0.4603\,\textcolor{gray}{(0.0\%)} & \cellcolor{red!12}1.1470\,\textcolor{gray}{(0.0\%)} \\
& EVAC-v2 Frame & \cellcolor{yellow!20}0.6937\,\textcolor{green!45!black}{(+12.6\%)} & \cellcolor{yellow!20}0.9220\,\textcolor{green!45!black}{(+3.6\%)} & \cellcolor{yellow!12}0.4606\,\textcolor{green!45!black}{(+0.1\%)} & \cellcolor{yellow!20}1.1770\,\textcolor{green!45!black}{(+2.6\%)} \\
& EVAC-v2 Fr.+Patch & \cellcolor{green!15}\textbf{0.7053}\,\textcolor{green!45!black}{(+14.5\%)} & \cellcolor{green!15}\textbf{0.9240}\,\textcolor{green!45!black}{(+3.8\%)} & \cellcolor{green!20}\textbf{0.4622}\,\textcolor{green!45!black}{(+0.4\%)} & \cellcolor{green!15}\textbf{1.2470}\,\textcolor{green!45!black}{(+8.7\%)} \\

\midrule
\multirow{4}{*}{\shortstack[c]{stack bowls three\\(ep087)}}
& Base EVAC & 0.5532\,\textcolor{red!70!black}{(-15.5\%)} & \cellcolor{red!12}0.8000\,\textcolor{green!45!black}{(+12.5\%)} & \cellcolor{red!12}0.4340\,\textcolor{green!45!black}{(+4.8\%)} & \cellcolor{red!12}0.0000\,\textcolor{red!70!black}{(-100.0\%)} \\
& EVAC-v1 & \cellcolor{red!12}0.6545\,\textcolor{gray}{(0.0\%)} & 0.7110\,\textcolor{gray}{(0.0\%)} & 0.4143\,\textcolor{gray}{(0.0\%)} & \cellcolor{yellow!20}0.3390\,\textcolor{gray}{(0.0\%)} \\
& EVAC-v2 Frame & \cellcolor{yellow!20}0.7407\,\textcolor{green!45!black}{(+13.2\%)} & \cellcolor{green!15}\textbf{0.9060}\,\textcolor{green!45!black}{(+27.4\%)} & \cellcolor{yellow!20}0.4403\,\textcolor{green!45!black}{(+6.3\%)} & \cellcolor{red!12}0.0000\,\textcolor{red!70!black}{(-100.0\%)} \\
& EVAC-v2 Fr.+Patch & \cellcolor{green!15}\textbf{0.7555}\,\textcolor{green!45!black}{(+15.4\%)} & \cellcolor{yellow!20}0.8880\,\textcolor{green!45!black}{(+24.9\%)} & \cellcolor{green!15}\textbf{0.4687}\,\textcolor{green!45!black}{(+13.1\%)} & \cellcolor{green!15}\textbf{0.4550}\,\textcolor{green!45!black}{(+34.2\%)} \\

\midrule
\multirow{4}{*}{\shortstack[c]{turn switch\\(ep310)}}
& Base EVAC & 0.5578\,\textcolor{red!70!black}{(-19.4\%)} & 0.8260\,\textcolor{red!70!black}{(-11.2\%)} & 0.4431\,\textcolor{red!70!black}{(-31.5\%)} & 0.0000\,\textcolor{gray}{(--)} \\
& EVAC-v1 & \cellcolor{red!12}0.6923\,\textcolor{gray}{(0.0\%)} & \cellcolor{red!12}0.9300\,\textcolor{gray}{(0.0\%)} & \cellcolor{red!12}0.6464\,\textcolor{gray}{(0.0\%)} & 0.0000\,\textcolor{gray}{(0.0\%)} \\
& EVAC-v2 Frame & \cellcolor{yellow!20}0.7028\,\textcolor{green!45!black}{(+1.5\%)} & \cellcolor{green!15}\textbf{0.9560}\,\textcolor{green!45!black}{(+2.8\%)} & \cellcolor{yellow!20}0.6869\,\textcolor{green!45!black}{(+6.3\%)} & 0.0000\,\textcolor{gray}{(--)} \\
& EVAC-v2 Fr.+Patch & \cellcolor{green!15}\textbf{0.7350}\,\textcolor{green!45!black}{(+6.2\%)} & \cellcolor{yellow!20}0.9410\,\textcolor{green!45!black}{(+1.2\%)} & \cellcolor{green!15}\textbf{0.8269}\,\textcolor{green!45!black}{(+27.9\%)} & 0.0000\,\textcolor{gray}{(--)} \\

\bottomrule
\end{tabular}}
\vspace{-0.15cm}
\end{table*}

\begin{figure*}[h]
    \centering
    \includegraphics[width=\textwidth]
    {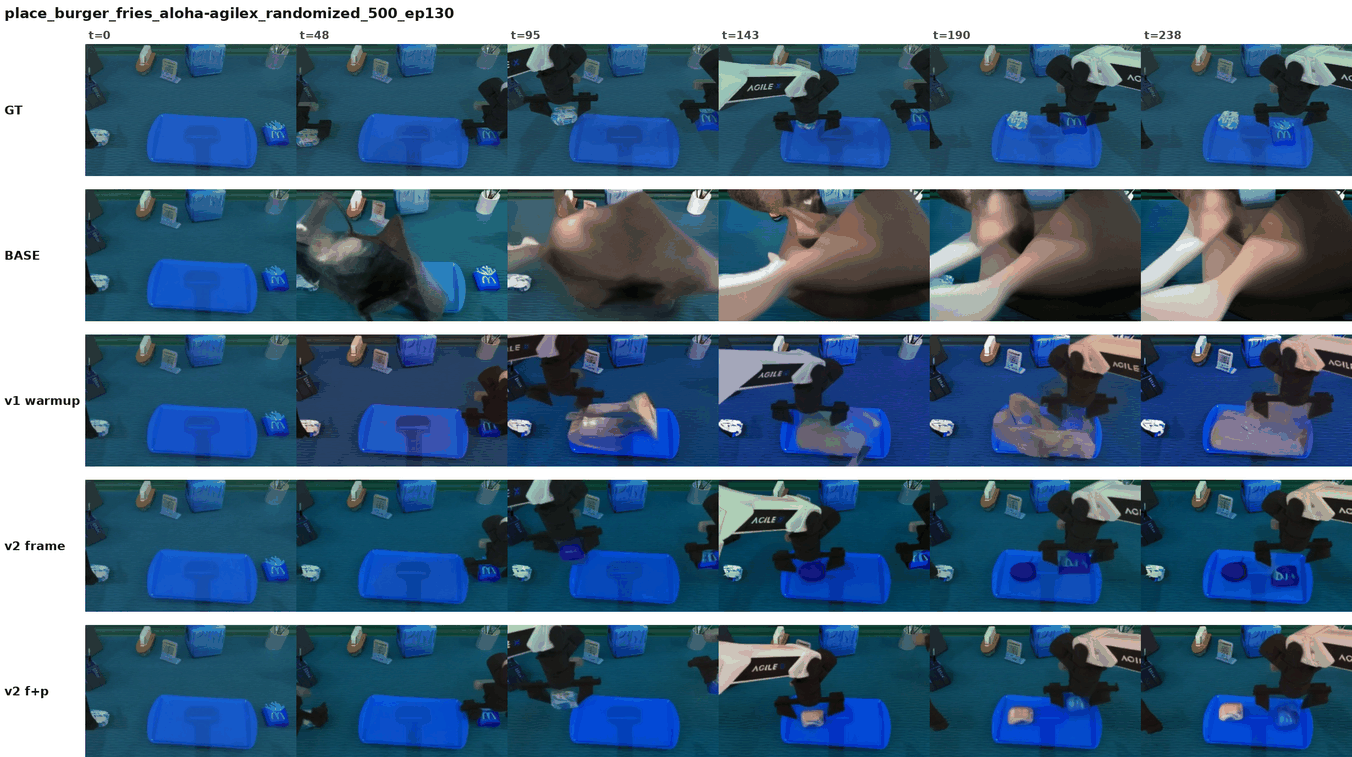}
    \vspace{-0.2cm}
    \caption{
    \textbf{Training evolution on \texttt{place\_burger\_fries} (ep130).}
    EVAC-v1 develops a strong blue color shift and the tray contents collapse
    from the middle of the rollout. EVAC-v2 Frame removes most tray corruption,
    but the left arm hallucinates an object instead of grasping the burger and
    the object held by the right arm is visibly deformed. EVAC-v2 Fr.+Patch
    correctly grasps and places the burger while preserving the fries most faithfully.
    }
    \label{fig:app_evolution_burger_fries}
\end{figure*}

\begin{figure*}[h]
    \centering
    \includegraphics[width=\textwidth]
    {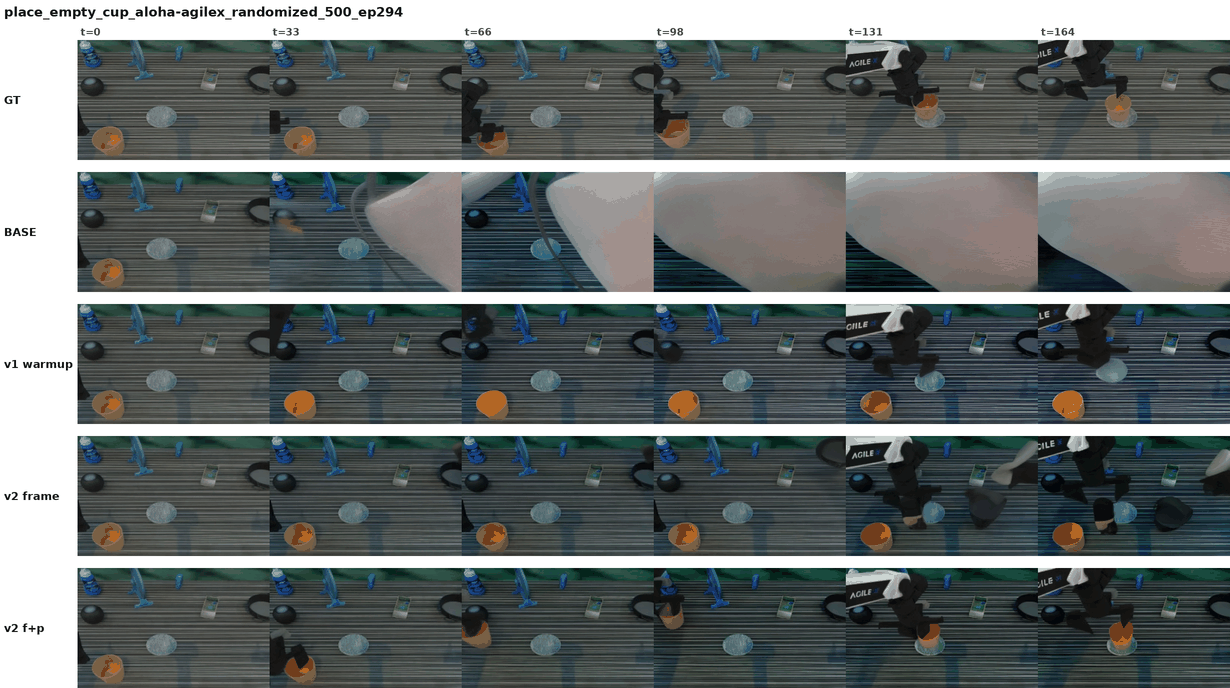}
    \vspace{-0.2cm}
    \caption{
    \textbf{Training evolution on \texttt{place\_empty\_cup} (ep294).}
    In both EVAC-v1 and EVAC-v2 Frame, the arm only partially appears near the
    image boundary and fails to grasp the cup at the beginning; EVAC-v2 Frame
    additionally deteriorates toward the end of the rollout. EVAC-v2 Fr.+Patch
    correctly grasps the cup from the boundary, places it on the plate, and
    maintains a coherent scene throughout the prediction.
    }
    \label{fig:app_evolution_empty_cup}
    \vspace{-1cm}
\end{figure*}

\begin{figure*}[h]
    \centering
    \includegraphics[width=\textwidth]
    {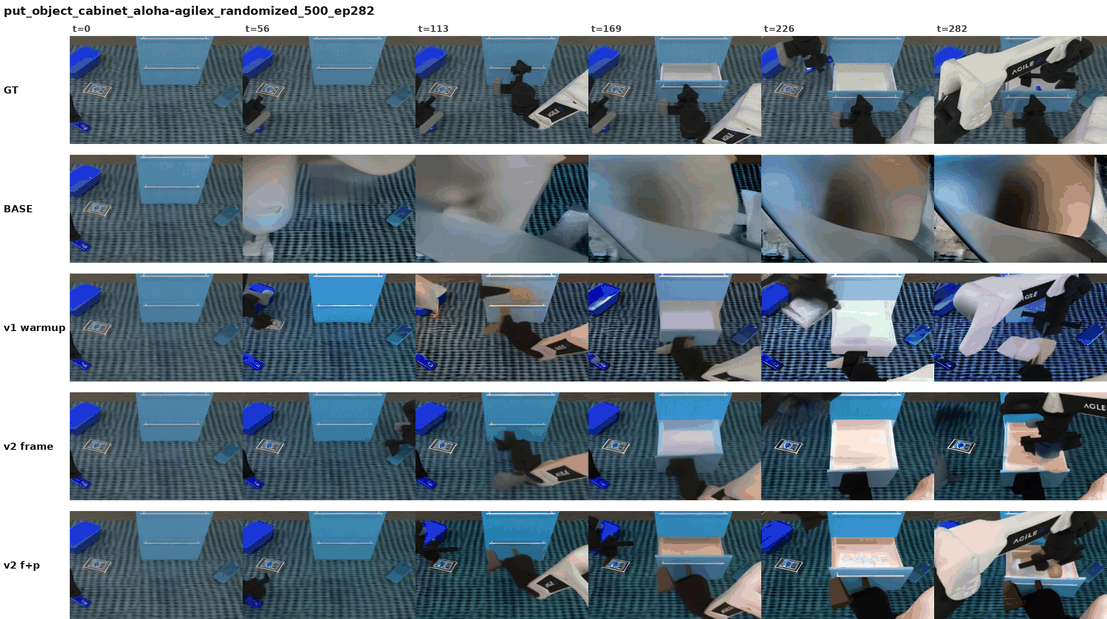}
    \vspace{-0.2cm}
    \caption{
    \textbf{Training evolution on \texttt{put\_object\_cabinet} (ep282).}
    EVAC-v1 shows moderate color bias and eventually produces a floating arm
    together with a corrupted cabinet. EVAC-v2 Frame does not resolve these
    failures: the blue box in the upper-left region disappears and the arm is
    generated at an incorrect location. EVAC-v2 Fr.+Patch preserves the objects,
    cabinet structure, and robot geometry most consistently.
    }
    \label{fig:app_evolution_put_cabinet}
\end{figure*}

\begin{figure*}[h]
    \centering
    \includegraphics[width=\textwidth]
    {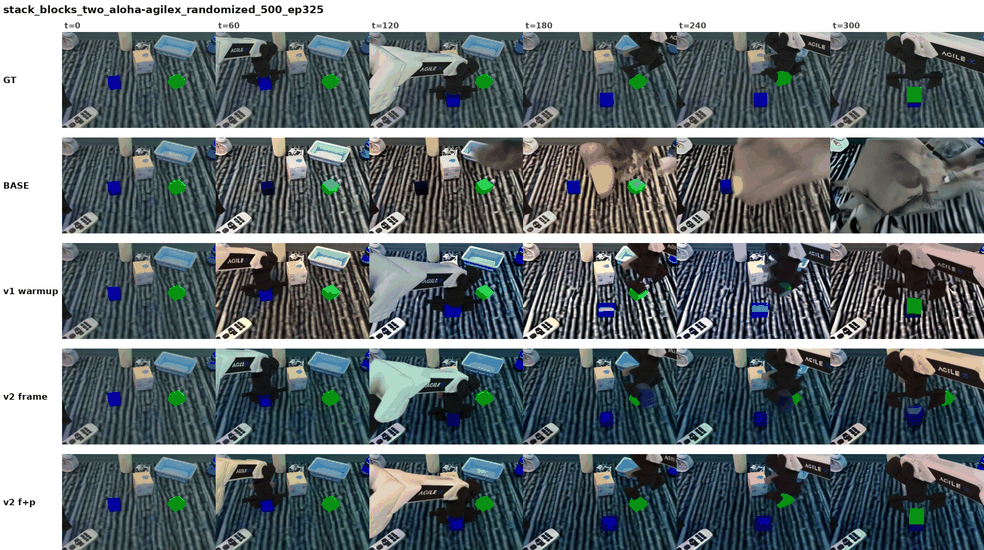}
    \vspace{-0.2cm}
    \caption{
    \textbf{Training evolution on \texttt{stack\_blocks\_two} (ep325).}
    EVAC-v1 exhibits strong color flickering, and the lower blue block disappears
    when the green block is lifted and stacked. EVAC-v2 Frame largely removes the
    color shift but fails to keep the green block attached to the gripper and
    consequently misses the stacking interaction. EVAC-v2 Fr.+Patch resolves
    both the object-persistence and grasping failures.
    }
    \label{fig:app_evolution_stack_blocks}
    \vspace{-1cm}
\end{figure*}

\begin{figure*}[h]
    \centering
    \includegraphics[width=\textwidth]
    {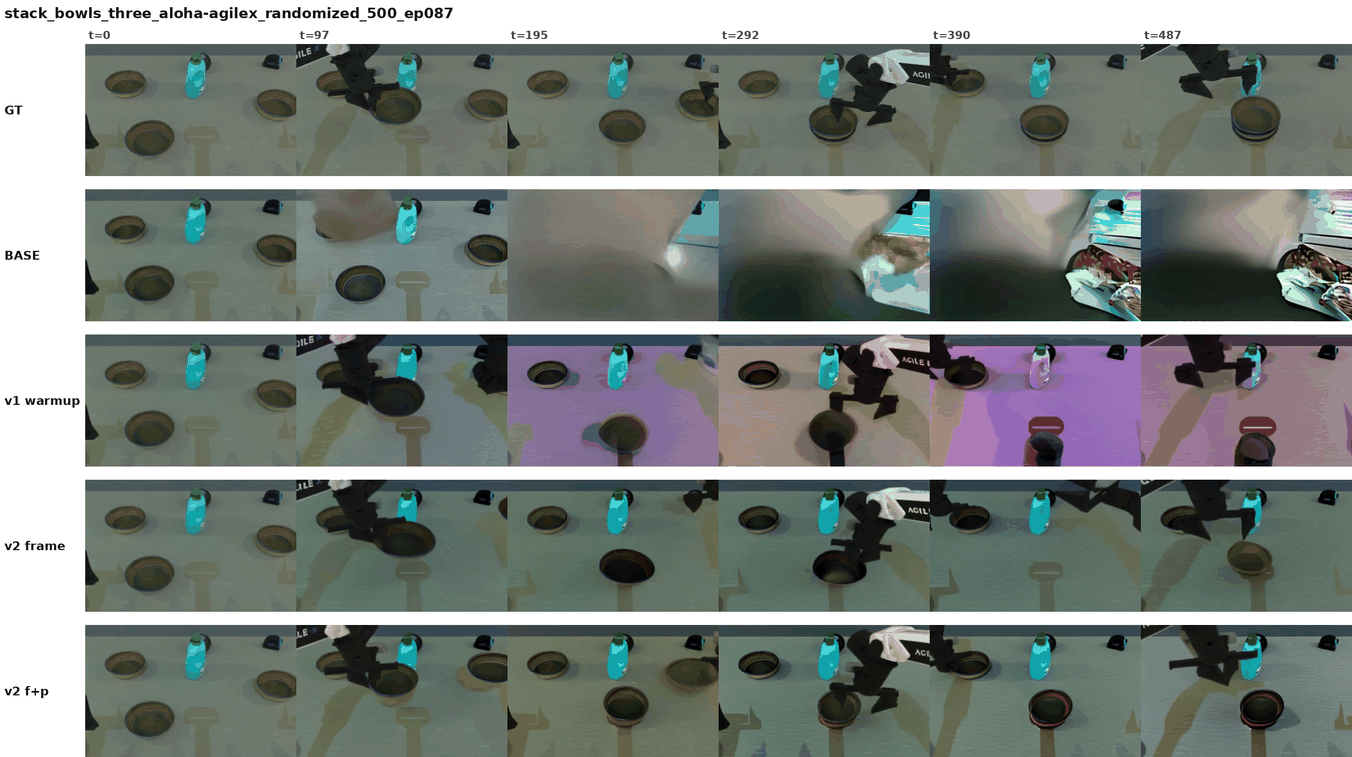}
    \vspace{-0.2cm}
    \caption{
    \textbf{Training evolution on \texttt{stack\_bowls\_three} (ep087).}
    EVAC-v1 suffers from pronounced color drift and repeated red flickering,
    followed by severe object corruption late in the rollout. EVAC-v2 Frame
    removes the color shift but still produces substantial bowl deformation,
    disappearance, and reappearance. EVAC-v2 Fr.+Patch better preserves the bowl
    geometry and maintains object persistence through the later frames.
    }
    \label{fig:app_evolution_stack_bowls}
\end{figure*}

\begin{figure*}[h]
    \centering
    \includegraphics[width=\textwidth]
    {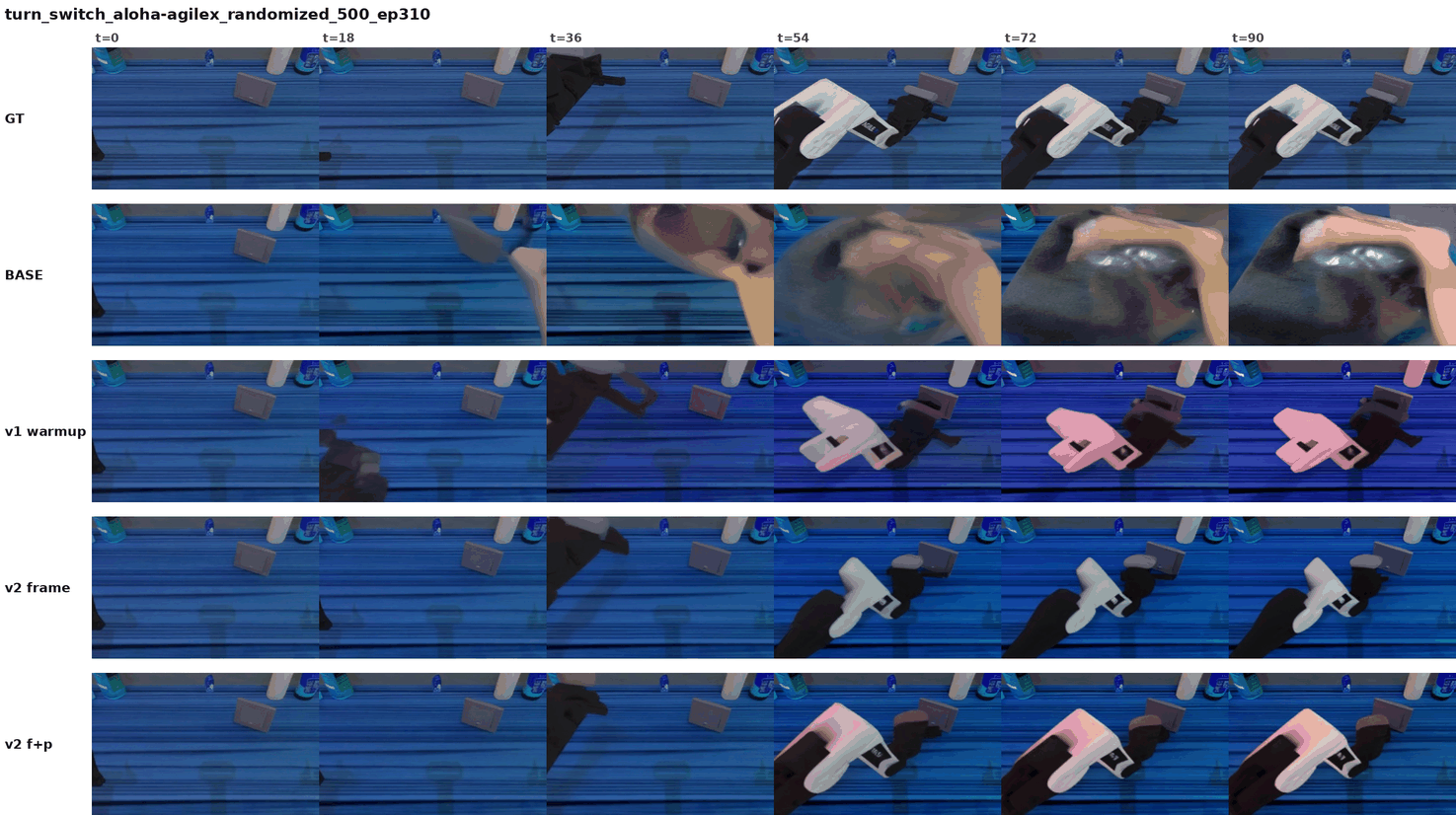}
    \vspace{-0.2cm}
    \caption{
    \textbf{Training evolution on \texttt{turn\_switch} (ep310).}
    EVAC-v1 produces a reddish robot arm, a bluish background, and implausible
    floating-arm geometry. EVAC-v2 Frame substantially reduces the color bias,
    but the arm shape remains distorted. EVAC-v2 Fr.+Patch further improves the
    local arm geometry and produces an embodiment substantially closer to the
    ground truth.
    }
    \label{fig:app_evolution_turn_switch}
    \vspace{-1cm}
\end{figure*}

\paragraph{Representative visual evolution.}
Figures~\ref{fig:app_evolution_burger_fries}--\ref{fig:app_evolution_turn_switch}
provide matched temporal comparisons for the six representative episodes. Base
EVAC is strongly out-of-distribution under the RoboTwin2.0 embodiment and often
fails to preserve a recognizable robot arm or coherent interaction trajectory.
After warmup, EVAC-v1 recovers the robot embodiment and basic scene structure,
but noticeable color drift, object corruption, temporal flickering, and
interaction errors remain. EVAC-v2 with frame weighting substantially improves
the overall rollout, although local object geometry and gripper--object
interactions can still fail. Frame-and-patch weighting further concentrates
learning on these difficult local regions and most clearly improves manipulated
objects, gripper geometry, and fine interaction details in the selected cases.

\newpage

\newpage

\newpage

\newpage
\paragraph{Detailed EWMBench measurements.}
Table~\ref{tab:app_training_evolution_detailed} reports the normalized component metrics underlying these representative examples. Overall, the transition from Base EVAC to EVAC-v1 substantially improves target-domain prediction quality, while confidence-guided EVAC-v2 provides further gains across reconstruction, scene, semantic, and trajectory dimensions. Among the selected episodes, frame-and-patch weighting gives the strongest overall progression, particularly for semantic quality and Motion when valid trajectory signals are available. These quantitative trends are consistent with the qualitative improvements shown above.

\begin{table*}[h]
\vspace{-0.25cm}
\caption{
\textbf{Detailed normalized EWMBench results for the qualitative evolution study.}
All metrics are higher-is-better.
Green, yellow, and red indicate the best, second-best, and third-best distinct values within each episode, respectively.
}
\label{tab:app_training_evolution_detailed}
\centering
\scriptsize
\setlength{\tabcolsep}{2.8pt}
\renewcommand{\arraystretch}{1.04}
\resizebox{\textwidth}{!}{
\begin{tabular}{c|l||cc|c|cc|ccc}
\toprule
\rowcolor{gray!12}
\multirow{2}{*}{Episode} & \multirow{2}{*}{Variant}
& \multicolumn{2}{c}{Reconstruction ($\uparrow$)}
& \multicolumn{1}{c}{Scene ($\uparrow$)}
& \multicolumn{2}{c}{Semantics ($\uparrow$)}
& \multicolumn{3}{c}{Motion ($\uparrow$)} \\
\cmidrule{3-10}
\rowcolor{gray!12}
& & PSNR ($\uparrow$) & SSIM ($\uparrow$) & Scene Cons. ($\uparrow$)
& Sem.-CLIP ($\uparrow$) & Sem.-BLEU ($\uparrow$)
& Traj-HSD ($\uparrow$) & Traj-Dyn ($\uparrow$) & Traj-nDTW ($\uparrow$) \\
\midrule

\multirow{4}{*}{\shortstack[c]{place burger fries\\(ep130)}}
& Base EVAC & 0.4413 & 0.477 & 0.789 & 0.7183 & 0.000 & 0.000 & 0.000 & 0.000 \\
& EVAC-v1 & \cellcolor{red!12}0.5957 & \cellcolor{yellow!20}0.668 & \cellcolor{red!12}0.877 & \cellcolor{red!12}0.7879 & \cellcolor{red!12}0.222 & \cellcolor{red!12}0.095 & \cellcolor{red!12}0.001 & \cellcolor{red!12}0.361 \\
& EVAC-v2 Frame & \cellcolor{yellow!20}0.6777 & \cellcolor{red!12}0.646 & \cellcolor{green!15}\textbf{0.943} & \cellcolor{yellow!20}0.8121 & \cellcolor{yellow!20}0.232 & \cellcolor{yellow!20}0.545 & \cellcolor{green!15}\textbf{0.024} & \cellcolor{yellow!20}0.514 \\
& EVAC-v2 Fr.+Patch & \cellcolor{green!15}\textbf{0.6980} & \cellcolor{green!15}\textbf{0.706} & \cellcolor{yellow!20}0.937 & \cellcolor{green!15}\textbf{0.8182} & \cellcolor{green!15}\textbf{0.279} & \cellcolor{green!15}\textbf{0.761} & \cellcolor{yellow!20}0.002 & \cellcolor{green!15}\textbf{0.545} \\

\midrule
\multirow{4}{*}{\shortstack[c]{place empty cup\\(ep294)}}
& Base EVAC & 0.5100 & 0.411 & 0.712 & \cellcolor{yellow!20}0.8448 & \cellcolor{red!12}0.246 & 0.000 & 0.000 & 0.000 \\
& EVAC-v1 & \cellcolor{green!15}\textbf{0.7217} & \cellcolor{yellow!20}0.779 & \cellcolor{green!15}\textbf{0.975} & \cellcolor{red!12}0.8038 & \cellcolor{yellow!20}0.266 & \cellcolor{yellow!20}0.108 & \cellcolor{yellow!20}0.365 & \cellcolor{yellow!20}0.087 \\
& EVAC-v2 Frame & \cellcolor{red!12}0.7117 & \cellcolor{red!12}0.759 & \cellcolor{yellow!20}0.970 & 0.3568 & 0.164 & \cellcolor{red!12}0.097 & \cellcolor{red!12}0.238 & \cellcolor{red!12}0.080 \\
& EVAC-v2 Fr.+Patch & \cellcolor{yellow!20}0.7197 & \cellcolor{green!15}\textbf{0.792} & \cellcolor{red!12}0.961 & \cellcolor{green!15}\textbf{0.8760} & \cellcolor{green!15}\textbf{0.327} & \cellcolor{green!15}\textbf{0.117} & \cellcolor{green!15}\textbf{0.379} & \cellcolor{green!15}\textbf{0.094} \\

\midrule
\multirow{4}{*}{\shortstack[c]{put object cabinet\\(ep282)}}
& Base EVAC & 0.4593 & 0.415 & 0.777 & \cellcolor{red!12}0.8081 & 0.163 & 0.000 & 0.000 & 0.000 \\
& EVAC-v1 & \cellcolor{red!12}0.5377 & \cellcolor{red!12}0.601 & \cellcolor{red!12}0.867 & 0.7927 & \cellcolor{yellow!20}0.183 & 0.000 & 0.000 & 0.000 \\
& EVAC-v2 Frame & \cellcolor{yellow!20}0.5727 & \cellcolor{green!15}\textbf{0.663} & \cellcolor{green!15}\textbf{0.926} & \cellcolor{yellow!20}0.8708 & \cellcolor{red!12}0.182 & 0.000 & 0.000 & 0.000 \\
& EVAC-v2 Fr.+Patch & \cellcolor{green!15}\textbf{0.5903} & \cellcolor{yellow!20}0.651 & \cellcolor{yellow!20}0.895 & \cellcolor{green!15}\textbf{0.9262} & \cellcolor{green!15}\textbf{0.204} & 0.000 & 0.000 & 0.000 \\

\midrule
\multirow{4}{*}{\shortstack[c]{stack blocks two\\(ep325)}}
& Base EVAC & 0.4537 & 0.305 & 0.797 & 0.7823 & \cellcolor{red!12}0.096 & 0.000 & 0.000 & 0.000 \\
& EVAC-v1 & \cellcolor{red!12}0.5650 & \cellcolor{red!12}0.667 & \cellcolor{red!12}0.890 & \cellcolor{green!15}\textbf{0.8246} & \cellcolor{red!12}0.096 & \cellcolor{red!12}0.303 & \cellcolor{red!12}0.205 & \cellcolor{red!12}0.639 \\
& EVAC-v2 Frame & \cellcolor{yellow!20}0.6313 & \cellcolor{yellow!20}0.756 & \cellcolor{yellow!20}0.922 & \cellcolor{red!12}0.8052 & \cellcolor{green!15}\textbf{0.116} & \cellcolor{yellow!20}0.304 & \cellcolor{yellow!20}0.218 & \cellcolor{yellow!20}0.655 \\
& EVAC-v2 Fr.+Patch & \cellcolor{green!15}\textbf{0.6447} & \cellcolor{green!15}\textbf{0.766} & \cellcolor{green!15}\textbf{0.924} & \cellcolor{yellow!20}0.8244 & \cellcolor{yellow!20}0.100 & \cellcolor{green!15}\textbf{0.319} & \cellcolor{green!15}\textbf{0.259} & \cellcolor{green!15}\textbf{0.669} \\

\midrule
\multirow{4}{*}{\shortstack[c]{stack bowls three\\(ep087)}}
& Base EVAC & 0.4953 & 0.611 & \cellcolor{red!12}0.800 & 0.7860 & \cellcolor{yellow!20}0.082 & \cellcolor{red!12}0.000 & \cellcolor{red!12}0.000 & \cellcolor{red!12}0.000 \\
& EVAC-v1 & \cellcolor{red!12}0.5630 & \cellcolor{red!12}0.746 & 0.711 & \cellcolor{green!15}\textbf{0.8287} & 0.000 & \cellcolor{green!15}\textbf{0.063} & \cellcolor{yellow!20}0.161 & \cellcolor{yellow!20}0.115 \\
& EVAC-v2 Frame & \cellcolor{yellow!20}0.6803 & \cellcolor{green!15}\textbf{0.801} & \cellcolor{green!15}\textbf{0.906} & \cellcolor{red!12}0.8116 & \cellcolor{red!12}0.069 & \cellcolor{red!12}0.000 & \cellcolor{red!12}0.000 & \cellcolor{red!12}0.000 \\
& EVAC-v2 Fr.+Patch & \cellcolor{green!15}\textbf{0.7160} & \cellcolor{yellow!20}0.795 & \cellcolor{yellow!20}0.888 & \cellcolor{yellow!20}0.8204 & \cellcolor{green!15}\textbf{0.117} & \cellcolor{yellow!20}0.062 & \cellcolor{green!15}\textbf{0.225} & \cellcolor{green!15}\textbf{0.168} \\

\midrule
\multirow{4}{*}{\shortstack[c]{turn switch\\(ep310)}}
& Base EVAC & 0.5507 & 0.565 & 0.826 & 0.7632 & 0.123 & 0.000 & 0.000 & 0.000 \\
& EVAC-v1 & \cellcolor{red!12}0.6587 & \cellcolor{yellow!20}0.726 & \cellcolor{red!12}0.930 & \cellcolor{red!12}0.9108 & \cellcolor{red!12}0.382 & 0.000 & 0.000 & 0.000 \\
& EVAC-v2 Frame & \cellcolor{yellow!20}0.6807 & \cellcolor{red!12}0.725 & \cellcolor{green!15}\textbf{0.956} & \cellcolor{yellow!20}0.9308 & \cellcolor{yellow!20}0.443 & 0.000 & 0.000 & 0.000 \\
& EVAC-v2 Fr.+Patch & \cellcolor{green!15}\textbf{0.7170} & \cellcolor{green!15}\textbf{0.753} & \cellcolor{yellow!20}0.941 & \cellcolor{green!15}\textbf{0.9318} & \cellcolor{green!15}\textbf{0.722} & 0.000 & 0.000 & 0.000 \\

\bottomrule
\end{tabular}}
\end{table*}

\subsection{More Results of Qualitative Confidence Visualization}
\label{app: more results of qualitative confidence visualization}

\begin{figure}[h]
    \centering
    \vspace{-0.2cm}
    \includegraphics[width=\linewidth]{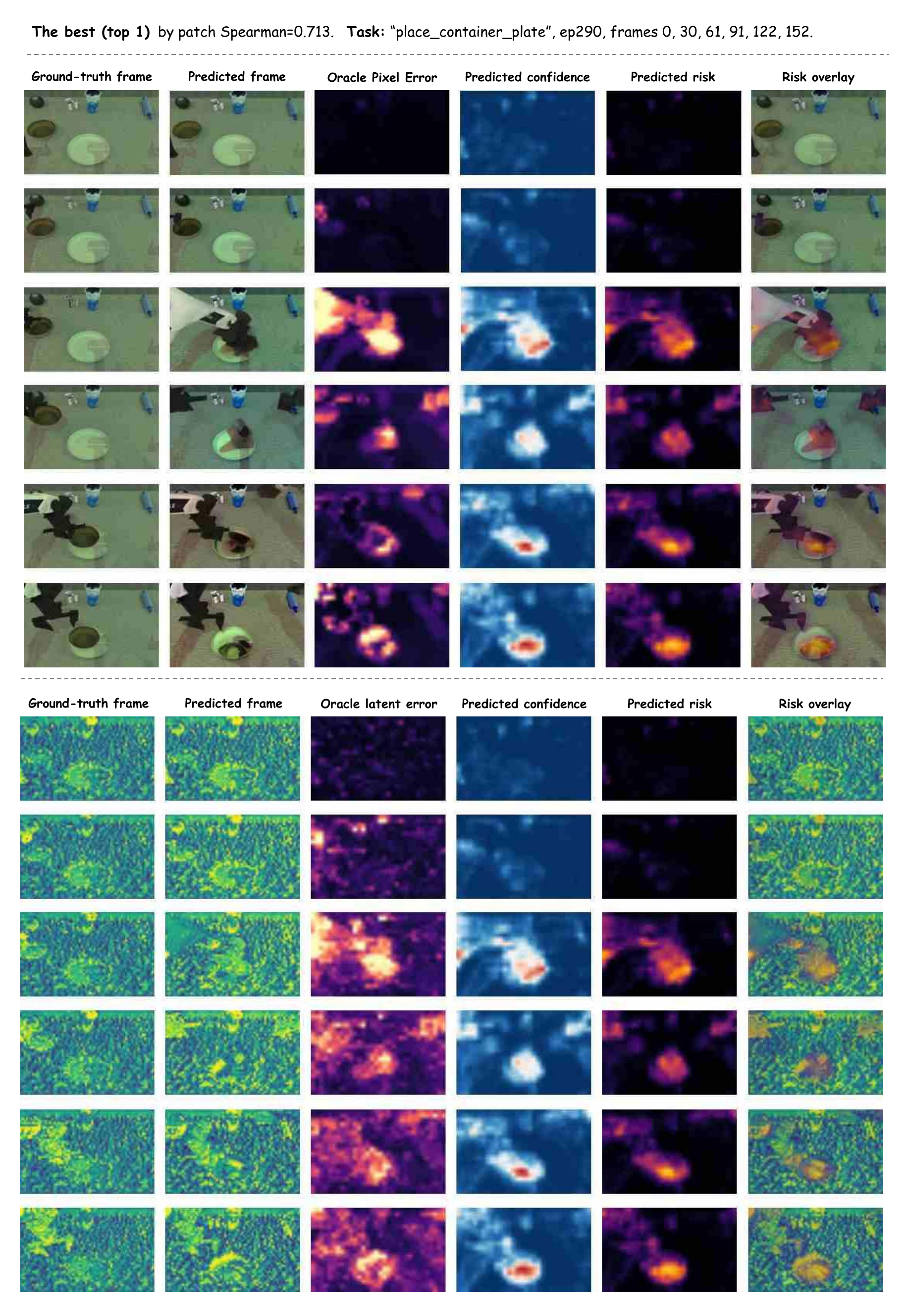}
    \vspace{-0.7cm}
    \caption{
    \textbf{More results of qualitative confidence visualization.}
    Best-ranked example (top 1) by patch-level Spearman correlation.
    }
    \label{fig:app_qual_conf_best1}
\end{figure}

\begin{figure}[h]
    \centering
    \vspace{-0.2cm}
    \includegraphics[width=\linewidth]{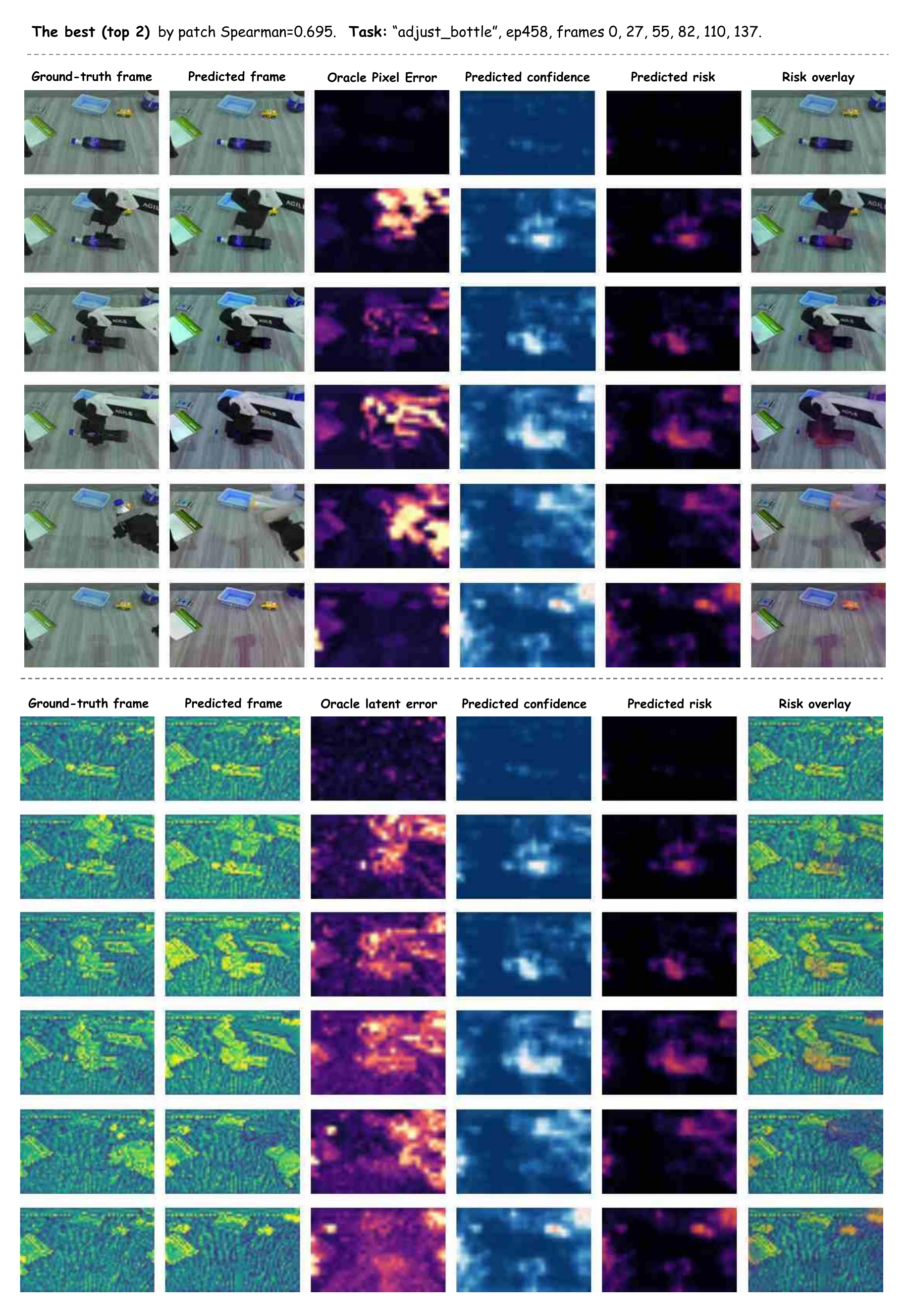}
    \vspace{-0.7cm}
    \caption{
    \textbf{More results of qualitative confidence visualization.}
    Second-highest-ranked example by patch-level Spearman correlation.
    }
    \label{fig:app_qual_conf_best2}
\end{figure}

We provide additional qualitative examples to complement the confidence analysis in the main text. Figures~\ref{fig:app_qual_conf_best1} and~\ref{fig:app_qual_conf_best2} show the two episodes with the highest patch-level Spearman correlations, reaching $0.713$ and $0.695$, respectively. Across different future frames, the predicted risk maps consistently concentrate around manipulators, operated objects, and interaction regions, and largely agree with both pixel- and latent-space oracle errors. The corresponding risk overlays further illustrate that confidence provides spatially localized signals rather than merely reflecting global video quality.

\begin{figure}[h]
    \centering
    \vspace{-0.2cm}
    \includegraphics[width=\linewidth]{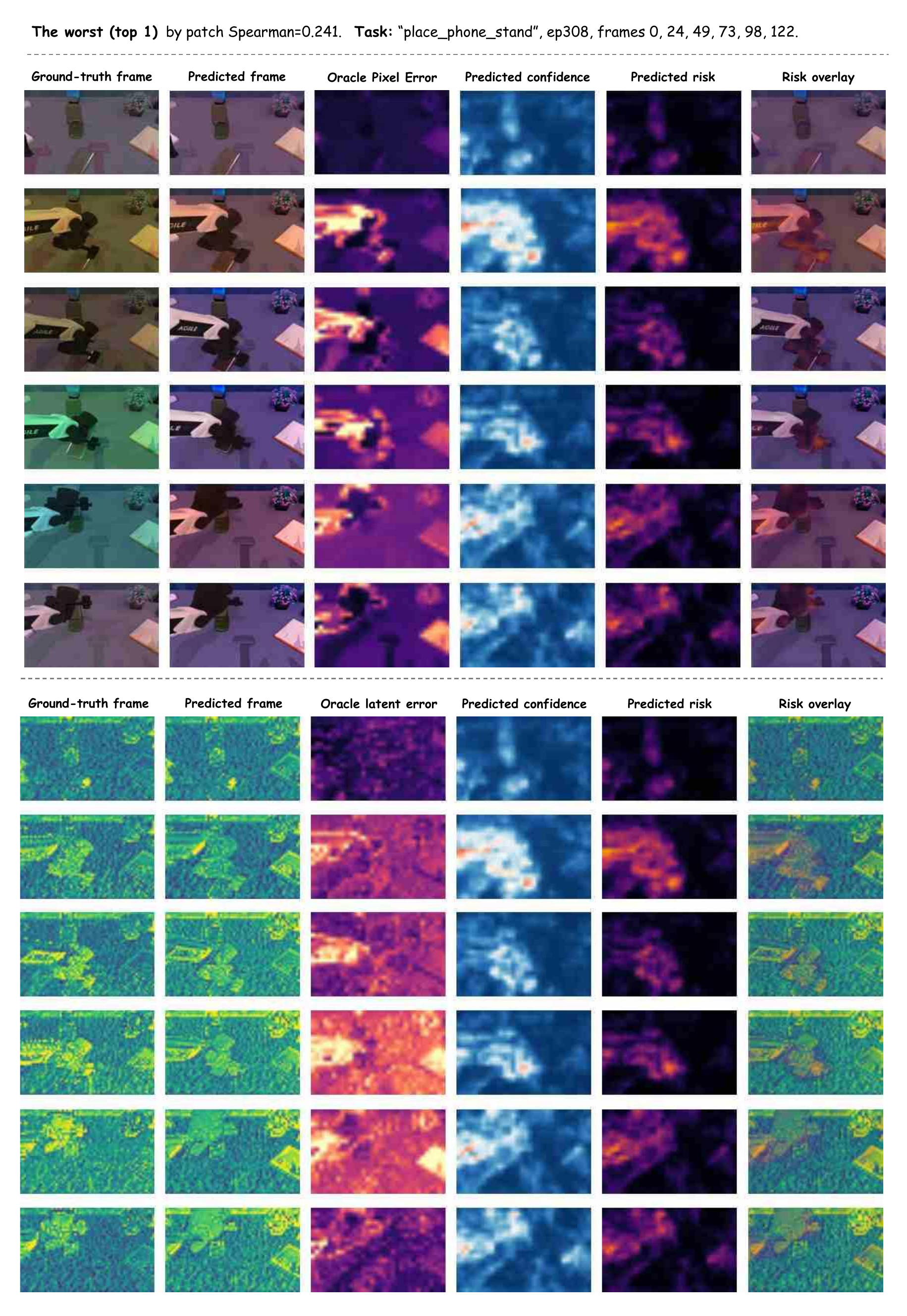}
    \vspace{-0.7cm}
    \caption{
    \textbf{More results of qualitative confidence visualization.}
    Lowest-ranked example (top 1) by patch-level Spearman correlation.
    }
    \label{fig:app_qual_conf_worst1}
\end{figure}

\begin{figure}[h]
    \centering
    \vspace{-0.2cm}
    \includegraphics[width=\linewidth]{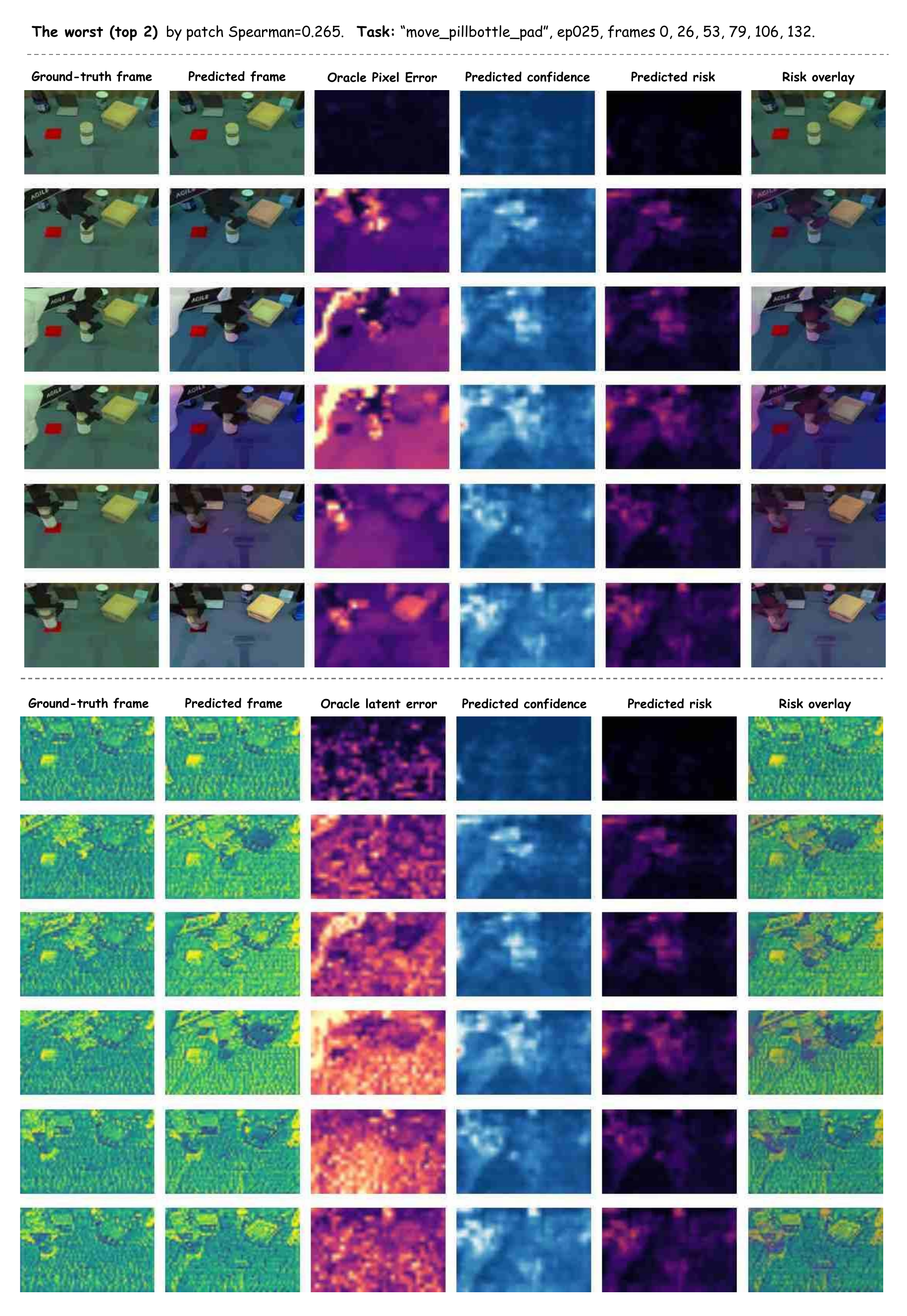}
    \vspace{-0.7cm}
    \caption{
    \textbf{More results of qualitative confidence visualization.}
    Second-lowest-ranked example by patch-level Spearman correlation.
    }
    \label{fig:app_qual_conf_worst2}
\end{figure}

Figures~\ref{fig:app_qual_conf_worst1} and~\ref{fig:app_qual_conf_worst2} show two challenging cases with lower patch-level Spearman correlations of $0.241$ and $0.265$. In these examples, oracle errors extend beyond the regions highlighted by risk, and predicted maps can remain weak despite substantial local errors, revealing missed failures rather than only imprecise spatial boundaries. Together, the best and worst cases illustrate both the localization capability and the remaining limitations of the confidence probe, consistent with our quantitative finding that confidence provides useful rankings but does not reliably recover all high-error regions, particularly when the predicted risk map remains weak despite substantial oracle prediction errors.
 
\newpage

\newpage

\newpage

\newpage

%% file: iclr2027_conference.bib
@misc{jiang2025enerverseac,
  title={EnerVerse-AC: Envisioning Embodied Environments with Action Condition},
  author={Jiang, Yuxin and Chen, Shengcong and Huang, Siyuan and Chen, Liliang and Zhou, Pengfei and Liao, Yue and He, Xindong and Liu, Chiming and Li, Hongsheng and Yao, Maoqing and Ren, Guanghui},
  year={2025},
  eprint={2505.09723},
  archivePrefix={arXiv},
  primaryClass={cs.RO},
  url={https://arxiv.org/abs/2505.09723}
}

@misc{mei2025c3,
  title={World Models That Know When They Don't Know: Controllable Video Generation with Calibrated Uncertainty},
  author={Mei, Zhiting and Yin, Tenny and Baker, Micah and Shorinwa, Ola and Majumdar, Anirudha},
  year={2026},
  eprint={2512.05927},
  archivePrefix={arXiv},
  primaryClass={cs.CV},
  url={https://arxiv.org/abs/2512.05927}
}

@inproceedings{yue2025ewmbench,
    author={Hu, Yue and Huang, Siyuan and Liao, Yue and Chen, Shengcong and Zhou, Pengfei and Chen, Liliang and Ren, Guanghui and Yao, Maoqing},
    title={EWMBench: Evaluating Scene, Motion, and Semantic Quality in Embodied World Models},
    booktitle={36th British Machine Vision Conference 2025, {BMVC} 2025, Sheffield, UK, November 24--27, 2025},
    publisher={BMVA},
    year={2025},
    url={https://bmva-archive.org.uk/bmvc/2025/assets/papers/Paper_736/paper.pdf}
}

@inproceedings{jang2025dreamgen,
  title={DreamGen: Unlocking Generalization in Robot Learning through Video World Models},
  author={Jang, Joel and Ye, Seonghyeon and Lin, Zongyu and Xiang, Jiannan and Bjorck, Johan and Fang, Yu and Hu, Fengyuan and Huang, Spencer and Kundalia, Kaushil and Lin, Yen-Chen and Magne, Loic and Mandlekar, Ajay and Narayan, Avnish and Tan, You Liang and Wang, Guanzhi and Wang, Jing and Wang, Qi and Xu, Yinzhen and Zeng, Xiaohui and Zheng, Kaiyuan and Zheng, Ruijie and Liu, Ming-Yu and Zettlemoyer, Luke and Fox, Dieter and Kautz, Jan and Reed, Scott and Zhu, Yuke and Fan, Linxi},
  booktitle={Proceedings of the 9th Conference on Robot Learning},
  series={Proceedings of Machine Learning Research},
  volume={305},
  pages={5170--5194},
  year={2025},
  publisher={PMLR},
  url={https://proceedings.mlr.press/v305/jang25a.html}
}

@inproceedings{li2025worldmodelbench,
    author={Li, Dacheng and Fang, Yunhao and Chen, Yukang and Yang, Shuo and Cao, Shiyi and Wong, Justin and Luo, Michael and Wang, Xiaolong and Yin, Hongxu and Gonzalez, Joseph and Stoica, Ion and Han, Song and Lu, Yao},
    title={WorldModelBench: Judging Video Generation Models As World Models},
    booktitle={Advances in Neural Information Processing Systems},
    volume={38},
    year={2025},
    publisher={Curran Associates, Inc.},
    doi={10.52202/085713-1834},
    url={https://proceedings.neurips.cc/paper_files/paper/2025/file/4ec03ed08a3fcb59e1c815b5598beff1-Paper-Datasets_and_Benchmarks_Track.pdf}
}

@misc{nvidia2025pbench,
  title={PBench: A Physical AI Benchmark for World Models},
  author={{NVIDIA}},
  year={2025},
  url={https://huggingface.co/datasets/nvidia/PBench}
}

@misc{mei2025squbed,
  title={How Confident are Video Models? Empowering Video Models to Express their Uncertainty},
  author={Mei, Zhiting and Shorinwa, Ola and Majumdar, Anirudha},
  year={2025},
  eprint={2510.02571},
  archivePrefix={arXiv},
  primaryClass={cs.CV},
  url={https://arxiv.org/abs/2510.02571}
}

@misc{ji2026prmjudge,
  title={PRM-as-a-Judge: A Dense Evaluation Paradigm for Fine-Grained Robotic Auditing},
  author={Ji, Yuheng and Liu, Yuyang and Tan, Huajie and Huang, Xuchuan and Huang, Fanding and Xu, Yijie and Chi, Cheng and Zhao, Yuting and Lyu, Huaihai and Co, Peterson and Cao, Mingyu and Zhang, Qiongyu and Li, Zhe and Zhou, Enshen and Wang, Pengwei and Wang, Zhongyuan and Zhang, Shanghang and Zheng, Xiaolong},
  year={2026},
  eprint={2603.21669},
  archivePrefix={arXiv},
  primaryClass={cs.RO},
  url={https://arxiv.org/abs/2603.21669}
}

@misc{lee2026roboreward,
  title={RoboReward: General-Purpose Vision-Language Reward Models for Robotics},
  author={Lee, Tony and Wagenmaker, Andrew and Pertsch, Karl and Liang, Percy and Levine, Sergey and Finn, Chelsea},
  year={2026},
  eprint={2601.00675},
  archivePrefix={arXiv},
  primaryClass={cs.RO},
  url={https://arxiv.org/abs/2601.00675}
}

@misc{liang2026robometer,
  title={Robometer: Scaling General-Purpose Robotic Reward Models via Trajectory Comparisons},
  author={Liang, Anthony and Korkmaz, Yigit and Zhang, Jiahui and Hwang, Minyoung and Anwar, Abrar and Kaushik, Sidhant and Shah, Aditya and Huang, Alex S. and Zettlemoyer, Luke and Fox, Dieter and Xiang, Yu and Li, Anqi and Bobu, Andreea and Gupta, Abhishek and Tu, Stephen and Biyik, Erdem and Zhang, Jesse},
  year={2026},
  eprint={2603.02115},
  archivePrefix={arXiv},
  primaryClass={cs.RO},
  url={https://arxiv.org/abs/2603.02115}
}

@inproceedings{ma2024gvl,
    author={Ma, Yecheng Jason and Hejna, Joey and Fu, Chuyuan and Shah, Dhruv and Liang, Jacky and Xu, Zhuo and Kirmani, Sean and Xu, Peng and Driess, Danny and Xiao, Ted and Bastani, Osbert and Jayaraman, Dinesh and Yu, Wenhao and Zhang, Tingnan and Sadigh, Dorsa and Xia, Fei},
    title={Vision Language Models are In-Context Value Learners},
    booktitle={International Conference on Learning Representations},
    editor={Y. Yue and A. Garg and N. Peng and F. Sha and R. Yu},
    pages={33984--34009},
    volume={2025},
    year={2025},
    url={https://proceedings.iclr.cc/paper_files/paper/2025/file/54854cf15a24fff9f5134a8641136fe4-Paper-Conference.pdf}
}

@misc{wu2026lrm,
  title={Large Reward Models: Generalizable Online Robot Reward Generation with Vision-Language Models},
  author={Wu, Yanru and Yuan, Weiduo and Qi, Ang and Guizilini, Vitor and Mao, Jiageng and Wang, Yue},
  year={2026},
  eprint={2603.16065},
  archivePrefix={arXiv},
  primaryClass={cs.RO},
  url={https://arxiv.org/abs/2603.16065}
}

@inproceedings{eren2024muse1,
    author={Eren, Mehmet Arda and Babič, Jan and Oztop, Erhan},
    title={Sample-Efficient Robot Learning for Supervised Effect Prediction Tasks},
    booktitle={2026 IEEE/SICE International Symposium on System Integration (SII)},
    year={2026},
    pages={1198--1203},
    doi={10.1109/SII64115.2026.11404575}
}

@misc{dasgupta2024actnerf,
  title={Uncertainty-aware Active Learning of NeRF-based Object Models for Robot Manipulators using Visual and Re-orientation Actions},
  author={Dasgupta, Saptarshi and Gupta, Akshat and Tuli, Shreshth and Paul, Rohan},
  year={2024},
  eprint={2404.01812},
  archivePrefix={arXiv},
  primaryClass={cs.RO},
  url={https://arxiv.org/abs/2404.01812}
}

@misc{li2025rwm,
  title={Uncertainty-Aware Robotic World Model Makes Offline Model-Based Reinforcement Learning Work on Real Robots},
  author={Li, Chenhao and Krause, Andreas and Hutter, Marco},
  year={2026},
  eprint={2504.16680},
  archivePrefix={arXiv},
  primaryClass={cs.RO},
  url={https://arxiv.org/abs/2504.16680}
}

@techreport{settles2009active,
  title={Active Learning Literature Survey},
  author={Settles, Burr},
  institution={University of Wisconsin--Madison},
  number={1648},
  type={Computer Sciences Technical Report},
  year={2009},
  url={https://burrsettles.com/pub/settles.activelearning.pdf}
}

@inproceedings{seung1992query,
  title={Query by Committee},
  author={Seung, H. Sebastian and Opper, Manfred and Sompolinsky, Haim},
  booktitle={Proceedings of the Fifth Annual Workshop on Computational Learning Theory},
  pages={287--294},
  year={1992},
  publisher={Association for Computing Machinery},
  series={COLT '92},
  doi={10.1145/130385.130417},
  url={https://doi.org/10.1145/130385.130417}
}

@inproceedings{sener2018coreset,
  title={Active Learning for Convolutional Neural Networks: A Core-Set Approach},
  author={Sener, Ozan and Savarese, Silvio},
  booktitle={International Conference on Learning Representations},
  year={2018},
  url={https://openreview.net/forum?id=H1aIuk-RW}
}

@misc{yuan2026fastwam,
  title={Fast-WAM: Do World Action Models Need Test-time Future Imagination?},
  author={Yuan, Tianyuan and Dong, Zibin and Liu, Yicheng and Zhao, Hang},
  year={2026},
  eprint={2603.16666},
  archivePrefix={arXiv},
  primaryClass={cs.RO},
  url={https://arxiv.org/abs/2603.16666}
}

@misc{lv2026viva,
  title={ViVa: A Video-Generative Value Model for Robot Reinforcement Learning},
  author={Lv, Jindi and Li, Hao and Li, Jie and Kong, Fankun and Wang, Yang and Yi, Pengfei and Nie, Yifei and Wang, Xiaofeng and Zhu, Zheng and Ni, Chaojun and Deng, Qiuping and Li, Hengtao and Lv, Jiancheng and Huang, Guan},
  year={2026},
  eprint={2604.08168},
  archivePrefix={arXiv},
  primaryClass={cs.RO},
  url={https://arxiv.org/abs/2604.08168}
}

@misc{luo2026beingh07,
  title={Being-H0.7: A Latent World-Action Model from Egocentric Videos},
  author={Luo, Hao and Zhang, Wanpeng and Feng, Yicheng and Zheng, Sipeng and Xu, Haiweng and Xu, Chaoyi and Xi, Ziheng and Fu, Yuhui and Lu, Zongqing},
  year={2026},
  eprint={2605.00078},
  archivePrefix={arXiv},
  primaryClass={cs.RO},
  url={https://arxiv.org/abs/2605.00078}
}

@misc{physisforcing2026,
    title={PhysisForcing: Physics Reinforced World Simulator for Robotic Manipulation},
    author={Zhang, Peiwen and Deng, Yufan and Sun, Shangkun and Ma, Juncheng and Wang, Duomin and Du, Jonas and Pan, Zilin and Huang, Ye and Liang, Hao and Huang, Songyan and Zhang, Ruihua and Xie, Enze and Liu, Ming-Yu and Zhou, Daquan},
    year={2026},
    eprint={2606.28128},
    archivePrefix={arXiv},
    primaryClass={cs.CV},
    url={https://arxiv.org/abs/2606.28128}
}

@inproceedings{agibot2025world,
  title={AgiBot World Colosseo: A Large-scale Manipulation Platform for Scalable and Intelligent Embodied Systems},
  author={Bu, Qingwen and Cai, Jisong and Chen, Li and Cui, Xiuqi and Ding, Yan and Feng, Siyuan and Gao, Shenyuan and He, Xindong and Huang, Xu and others},
  booktitle={2025 IEEE/RSJ International Conference on Intelligent Robots and Systems (IROS)},
  pages={3549--3556},
  year={2025},
  organization={IEEE},
  doi={10.1109/IROS60139.2025.11247088}
}

@misc{chen2025robotwin,
    title={RoboTwin 2.0: A Scalable Data Generator and Benchmark with Strong Domain Randomization for Robust Bimanual Robotic Manipulation},
    author={Chen, Tianxing and Chen, Zanxin and Chen, Baijun and Cai, Zijian and Liu, Yibin and Li, Zixuan and Liang, Qiwei and Lin, Xianliang and Ge, Yiheng and Gu, Zhenyu and Deng, Weiliang and Guo, Yubin and Nian, Tian and Xie, Xuanbing and Chen, Qiangyu and Su, Kailun and Xu, Tianling and Liu, Guodong and Hu, Mengkang and Gao, Huan-ang and Wang, Kaixuan and Liang, Zhixuan and Qin, Yusen and Yang, Xiaokang and Luo, Ping and Mu, Yao},
    year={2025},
    eprint={2506.18088},
    archivePrefix={arXiv},
    primaryClass={cs.RO},
    url={https://arxiv.org/abs/2506.18088}
}

@misc{wei2026cdlam,
    title={Causally Debiased Latent Action Model for Embodied Action Conditioned World Models},
    author={Wei, Yufan and Zhou, Kun and Mao, Lingjun and Zhang, Zijun and Xu, Ziming and Xi, Ziqiao and Liang, Shuang and Han, Ruobing and Yan, Yuchen and Wang, Xinyue and Feng, Fan and Huang, Biwei},
    year={2026},
    eprint={2607.09185},
    archivePrefix={arXiv},
    primaryClass={cs.CV},
    url={https://arxiv.org/abs/2607.09185}
}

@inproceedings{zhu2025irasim,
  title={IRASim: A Fine-Grained World Model for Robot Manipulation},
  author={Zhu, Fangqi and Wu, Hongtao and Guo, Song and Liu, Yuxiao and Cheang, Chilam and Kong, Tao},
  booktitle = {Proceedings of the IEEE/CVF International Conference on Computer Vision (ICCV)},
  pages={9834--9844},
  year={2025}
}

@inproceedings{zhou2025dinowm,
  title={{DINO-WM}: World Models on Pre-trained Visual Features Enable Zero-shot Planning},
  author={Zhou, Gaoyue and Pan, Hengkai and LeCun, Yann and Pinto, Lerrel},
  booktitle={Proceedings of the 42nd International Conference on Machine Learning},
  series={Proceedings of Machine Learning Research},
  volume={267},
  pages={79115--79135},
  publisher={PMLR},
  year={2025}
}

@inproceedings{dass2025datamil,
    author={Dass, Shivin and Khaddaj, Alaa and Engstrom, Logan and Madry, Aleksander and Ilyas, Andrew and Mart{\'i}n-Mart{\'i}n, Roberto},
    title={DataMIL: Selecting Data for Robot Imitation Learning with Datamodels},
    booktitle={International Conference on Learning Representations},
    editor={C. Vondrick and B. Hariharan and C. Raffel and L. Pinto and D. Yang and A. Faust},
    pages={1311--1336},
    volume={2026},
    year={2026},
    url={https://proceedings.iclr.cc/paper_files/paper/2026/file/033d9e8dbbbc0ad90e59222cf1db0fc2-Paper-Conference.pdf}
}

@inproceedings{liao2024devil,
    author={Liao, Mingxiang and Lu, Hannan and Zhang, Xinyu and Wan, Fang and Wang, Tianyu and Zhao, Yuzhong and Zuo, Wangmeng and Ye, Qixiang and Wang, Jingdong},
    booktitle={Advances in Neural Information Processing Systems},
    doi={10.52202/079017-3483},
    editor={A. Globerson and L. Mackey and D. Belgrave and A. Fan and U. Paquet and J. Tomczak and C. Zhang},
    pages={109790--109816},
    publisher={Curran Associates, Inc.},
    title={Evaluation of Text-to-Video Generation Models: A Dynamics Perspective},
    url={https://proceedings.neurips.cc/paper_files/paper/2024/file/c6483c8a68083af3383f91ee0dc6db95-Paper-Conference.pdf},
    volume={37},
    year={2024}
}

@misc{dou2026sgc,
    title={Measuring 3D Spatial Geometric Consistency in Dynamic Video Generation},
    author={Dou, Weijia and Zheng, Wenzhao and Chen, Weiliang and Zheng, Yu and Zhou, Jie and Lu, Jiwen},
    year={2026},
    eprint={2603.19048},
    archivePrefix={arXiv},
    primaryClass={cs.CV},
    url={https://arxiv.org/abs/2603.19048}
}

@inproceedings{ye2026shift,
    title={SHIFT: Motion Alignment in Video Diffusion Models with Adversarial Hybrid Fine-Tuning},
    author={Ye, Xi and Yang, Wenjia and Xu, Yangyang and Liu, Xiaoyang and Su, Duo and Xia, Mengfei and Zhu, Jun},
    booktitle={European Conference on Computer Vision (ECCV)},
    year={2026}
}

@misc{romer2026uncertainty,
    title={Uncertainty Quantification for Flow-Based Vision-Language-Action Models},
    author={R{\"o}mer, Ralf and Seeliger, Maximilian and Liu, Saida and Sturgis, Ben and Bagatella, Marco and Marta, Daniel and Krause, Andreas and Schoellig, Angela P.},
    year={2026},
    eprint={2606.18043},
    archivePrefix={arXiv},
    primaryClass={cs.RO},
    url={https://arxiv.org/abs/2606.18043}
}
